\documentclass[journal]{IEEEtran}
\usepackage[pdftex]{graphicx}
\usepackage{cite}
\usepackage{amsmath,amssymb}

\usepackage{amsthm}
\usepackage{mathtools}	%
\usepackage{grffile}	%
\usepackage[tight,footnotesize]{subfigure}
\usepackage{microtype} %
\usepackage{color}
\usepackage{url}
\usepackage[ruled,vlined,linesnumbered]{algorithm2e}
\usepackage{bm} %
\usepackage{comment}
\usepackage{arydshln}
\let\labelindent\relax
\usepackage{enumitem}
	
\usepackage{adjustbox}
\usepackage{tikz}
\usetikzlibrary{shadings, patterns, angles, quotes, arrows.meta, shapes, decorations.pathmorphing, decorations.shapes, decorations.text}
\usetikzlibrary{calc,intersections,arrows.meta}

\usepackage{pgfplots}
\tikzset{
	pil/.style={
		->,
		thick,
		shorten <=2pt,
		shorten >=2pt,}
}

\usepackage{multicol}

\usepackage{hyperref}
\hypersetup{
	colorlinks=true,
	linkcolor=blue,
	citecolor=red,
	urlcolor=black,
}

\theoremstyle{remark}
\newtheorem{remarkx}{Remark}
\newenvironment{remark}
{\pushQED{\qed}\remarkx}
{\popQED\endremarkx}
\newtheorem{examplex}{Example}
\newenvironment{example}
{\pushQED{\qed}\examplex}
{\popQED\endexamplex}

\theoremstyle{definition}

\newtheorem{assump}{Assumption}
\newtheorem{problem}{Problem}
\newtheorem*{problem*}{Problem}

\theoremstyle{plain}
\newtheorem{theorem}{Theorem}

\newtheorem{coroll}{Corollary}

\newcommand{\defeq}{:=} %

\newcommand{\dt}{\frac{{\rm d}}{{\rm d}t}}
\newcommand{\matr}[1]{\begin{bmatrix} #1 \end{bmatrix}}

\newcommand{\transpose}[1]{{#1}^\top}
\newcommand{\diag}[1]{{\rm diag}\{ #1\}}
\newcommand{\cmmnt}[1]{}
\newcommand{\chiup}{\raisebox{2pt}{$\chi$}}
\newcommand{\vf}{\chiup}
\newcommand{\set}[1]{\mathcal{#1}}

\newcommand{\norm}[1]{\left\lVert#1\right\rVert}
\newcommand{\mbr}[1][{}]{\mathbb{R}^{#1}}	%

\newcommand{\hgh}[1]{ \prescript{\mathrm{hgh}}{}{#1} }
\newcommand{\phy}[1]{ \prescript{\mathrm{phy}}{}{#1} }

\newcommand{\vfpf}{\prescript{\rm pf}{}{\vf}}	%
\newcommand{\vfsf}{\prescript{\rm sf}{}{\vf}} 	%
\newcommand{\vfcr}{\prescript{\rm cr}{}{\vf}}
\newcommand{\vfcrone}{\prescript{{\rm cr}}{}{\vf}_1}
\newcommand{\vfcrtwo}{\prescript{{\rm cr}}{}{\vf}_2}
\newcommand{\vfcb}{\mathfrak{X}}%
\newcommand{\vfcbnew}{ {\check{\mathfrak{X}}} }%
\newcommand{\normv}[1]{\overline{#1}}			%

\newcommand{\mvfcb}[2]{ {\vfcb^{[#1]}_{#2}} } %
\newcommand{\mhatvfcb}[2]{ {{\normv{\vfcb^{[#1]}}}_{#2}} }

\newcommand{\mphi}[2]{ {\phi^{[#1]}_{#2}} }	
\newcommand{\mf}[2]{ {f^{[#1]}_{#2}} }
\newcommand{\mfdot}[2]{ {{f^{[#1]}_{#2}}'} }

\newcommand{\mx}[2]{ {x^{[#1]}_{#2}} }
\newcommand{\mk}[2]{ {k^{[#1]}_{#2}} }

\newcommand{\mw}[2]{ {w^{[#1]}_{#2}} }
\newcommand{\mdotw}[2]{ {\dot{w}^{[#1]}_{#2}} }
\newcommand{\mtildew}[2]{ {\tilde{w}^{[#1]}_{#2}} }
\newcommand{\mdottildew}[2]{ {\dot{\tilde{w}}^{[#1]}_{#2}} }
\newcommand{\mc}[2]{ {c^{[#1]}_{#2}} }
\newcommand{\mfdotwone}[2]{ {\partial_{w_{1}} f^{[#1]}_{#2}} }
\newcommand{\mfdotwtwo}[2]{ {\partial_{w_{2}} f^{[#1]}_{#2}} }
\newcommand{\mfddotw}[4]{ {\partial^{w_{#1}}_{w_{#2}} f^{[#3]}_{#4}} }	%
\newcommand{\sat}{ {\mathrm{Sat}_{a}^{b}} }

\newcommand{\scalemath}[2]{\scalebox{#1}{\mbox{\ensuremath{\displaystyle #2}}}}

\begin{document}
\title{Guiding vector fields for the distributed motion coordination of mobile robots}

\author{Weijia Yao, Hector Garcia de Marina, Zhiyong Sun, Ming Cao %
}

\maketitle

\begin{abstract}
We propose coordinating guiding vector fields to achieve two tasks simultaneously with a team of robots: i) the guidance and navigation of multiple robots to possibly different paths or surfaces typically embedded in 2D or 3D; ii) their motion coordination while tracking their prescribed paths or surfaces. The motion coordination is defined by desired parametric displacements between robots on the path or surface. Such a desired displacement is achieved by controlling the virtual coordinates, which correspond to the path or surface's parameters, between guiding vector fields.

Rigorous mathematical guarantees underpinned by dynamical systems theory and Lyapunov theory are provided for the effective distributed motion coordination and navigation of robots on paths or surfaces from all initial positions. As an example for practical robotic applications, we derive a control algorithm from the proposed coordinating guiding vector fields for a Dubins-car-like model with actuation saturation.

Our proposed algorithm is distributed and scalable to an arbitrary number of robots. Furthermore, extensive illustrative simulations and fixed-wing aircraft outdoor experiments validate the effectiveness and robustness of our algorithm.

\end{abstract}


\IEEEpeerreviewmaketitle

\section{Introduction}
The scalable coordination of mobile robots is a key to achieve enhanced synergistic performance in a multi-robot system. For example, the proper coordination of multiple mobile robots minimizes interference and reduces duplicate work in tasks such as multi-robot path planning and coverage, and can boost the performance of cooperative and collaborative multi-robot missions \cite{prorok2021beyond}. However, to reliably and systematically coordinate robots in a large multi-robot system without a central nexus is one of the grand challenges in robotics \cite{yang2018grand}. A fundamental question of this challenge is how to design a mechanism such that multiple robots are able to accurately follow possibly different desired paths and coordinate their motions in a distributed fashion, with the resulting formation satisfying some geometric or parametric constraints. This paper proposes a coordinating guiding vector field to guide an arbitrary number of robots to achieve motion coordination on possibly different desired paths or surfaces in a distributed way through local information exchange. The proposed guiding vector field has many appealing features, such as ensuring rigorous guarantees of motion coordination and path/surface convergence.

\subsection{Related work}
Single-robot path following capability is fundamental in mobile robotics applications, and many algorithms have been proposed and widely studied \cite{Sujit2014}. It is concluded by numerical simulations in \cite{Sujit2014} that those algorithms based on guiding vector fields achieve relatively high path-following accuracy while they require less control effort, compared with several other tested algorithms, such as LQR-based ones \cite{ratnoo2011adaptive} and nonlinear guidance laws (NLGL) \cite{park2007performance}. This conclusion is partly experimentally supported by \cite{caharija2015comparison}. Despite the performance advantages, these vector field based algorithms are mostly designed for a single robot, while the extension for a multi-robot system is relatively less studied. In fact, to the best of our knowledge, there are only a few studies exploiting a guiding vector field for the multi-robot distributed motion coordination. The guiding vector field proposed in \cite{kapitanyuk2017guiding} is exploited in \cite{de2017circular} in combination with a distributed algorithm to dynamically change the radii of circular paths such that multiple fixed-wing aircraft flying at a constant speed can eventually follow the same circular path and keep pre-defined inter-robot distances. A different guiding vector field is derived in \cite{nakai2013vector} based on various potential functions for multiple robots to move on common paths such as a circle or a straight line. Another work \cite{pimenta2013decentralized} presents a distributed control law for a number of robots to circulate along a closed curve described in a specific form in 3D. Recently, the paper \cite{wang2022} employs a guiding vector field to achieve fixed-wing UAV formation control.

Without employing a guiding vector field, various algorithms are proposed in the literature for multi-robot coordinated path-following. A virtual structure is utilized in \cite{ghommam2010formation} for multiple robots to coordinate their motions while they follow predefined paths. As each robot needs to broadcast its states and reference trajectories to other robots, the communication overhead increases as the number of robots grows. A theoretical study in  \cite{zhang2007coordinated} considers planar simple closed curves and unit-speed particles. In \cite{sabattini2015implementation}, the desired motion coordination among robots are generated by a linear exo-system, and an output-regulation controller is implemented on some robots while others are controlled through local interactions. Many other works study multi-robot distributed following of planar simple desired paths (e.g., a straight line), such as \cite{chen2019coordinated,wang2019cooperative,doosthoseini2015coordinated,reyes2014flocking}.
The papers \cite{mong2006pattern,mong2007stabilization} design distributed controllers for robots characterized by single-integrator or double-integrator models  such that they can generate 2D desired geometric patterns  composed of simple closed curves or circulate along a 2D closed curve.
 
Circular formation control and circumnavigation control, which have been extensively studied, can be regarded as specialized cases of multi-robot coordinated path following control. To be more specific, in the circular formation control problem, robots are required to distribute along a circle and maintain some desired arc distances using local interactions \cite{iqbal2019circular,sun2018circular,wang2013forming}. Motivated by some specific applications, such as entrapment of a hostile target, an additional requirement for all robots to move persistently along the circle is imposed in a circumnavigation control problem \cite{Lu2019distributed,yao2019distributed,franchi2016decentralized}. Nevertheless, as clear from their problem formulations, these studies only take into account circular paths mostly in a 2D plane. Although it is possible to extend these studies to consider other closed paths, this might require some continuous transformation between paths, which is non-trivial in general.

\subsection{Contributions} \label{subsec:contribution}
There are two major contributions in this paper. The first focuses on extending the guiding vector field for tracking desired paths (i.e., one-dimensional manifolds) presented in \cite{yao2021singularity} to parametric surfaces. Such an extension allows a richer repertoire of coordinated collective motions. %
The second contribution focuses on the design of a coordinating guiding vector field for a multi-robot system to achieve global convergence to desired paths and surfaces while their motions are coordinated in a distributed way via local communication.

As a practical illustration, we show how from the proposed coordinating guiding vector field, one can design a control algorithm for guiding robots using a non-holonomic Dubins-car-like model with saturated actuators. Specifically, we consider desired paths and surfaces defined as parametric equations such that the motion coordination among robots can be quantified by relative parametric differences. The parameters describing the paths or surfaces are treated as virtual coordinates to construct the singularity-free guiding vector fields. Such virtual coordinates can be coupled between robots via a consensus term in an undirected communication graph. Consequently, these new guiding vector fields effectively control the \emph{relative parametric separation} among robots.

We would like to highlight a non-exhaustive list of four appealing aspects of our approach:
\begin{enumerate}
	\item There are rigorous mathematical guarantees on the global convergence of robots' trajectories converging to the desired paths or surfaces and achieving the motion coordination.
	\item Our approach can deal with complex paths, such as self-intersecting, non-closed, and non-convex ones in an $n$-dimensional configuration space, and it is also applicable for arbitrary surfaces described by parametric equations. The approach can also be naturally extended for higher-dimensional manifolds (e.g., a cube).
	\item Our approach is distributed and scalable. Given a fixed communication frequency, the communication cost is low, since every two neighboring robots only need to receive and transmit the virtual coordinates, of which the number is equal to that of the path or surface parameters. The approach is also computationally cheap and is suitable for real-time applications since online optimization is not required.
	\item We also demonstrate that our algorithm is promising in applications such as area or volume coverage tasks by exploiting 2D or 3D Lissajous curves with \emph{irrational coefficients}.
\end{enumerate}

This paper is a significant extension of our preliminary version in \cite{yao2021multi}. Hereby, we list below the improvements and original contributions with respect to our conference work \cite{yao2021multi}:

\begin{enumerate}
	\item The conference version only considered the case of multiple robots following different desired paths (i.e., one-dimensional manifolds), while, in this paper, we generalize the result for motion coordination on surfaces and higher-dimensional manifolds (Section \ref{sec:surface}).
	\item The conference version only briefly mentioned the use of safety barrier certificate with the proposed coordinating guiding vector field for collision avoidance, but we present more technical details in this paper (Section \ref{sec:collision}).
	\item  We add new simulations and experiments to verify the aforementioned new theoretical results (Section \ref{sec:simexp}).
	\item  We provide in this paper all the technical proofs that were omitted in \cite{yao2021multi} due to the conference page limit. In addition, in-depth interpretation of the theoretical results, complemented with additional results (e.g., Corollary \ref{coroll1}) is provided in this paper.
\end{enumerate}

The remainder of this paper is organized as follows. Section \ref{sec:preliminary} introduces the preliminaries on graph theory and guiding vector fields for path following. Then Section \ref{sec:pathfollowing} elaborates on how to extend the guiding vector field for multi-robot distributed motion coordination and navigation on desired paths.  The theoretical results are further extended to deal with coordinated maneuvering on parametric surfaces in Section \ref{sec:surface}.  We also discuss how the coordinating guiding vector field can be seamlessly integrated with a safety barrier certificate to address the collision issue in Section \ref{sec:collision}. Next, we consider a realistic robot model for fixed-wing aircraft and derive a control law from the coordinating guiding vector field in Section \ref{sec:controlalg}. Moreover, extensive simulation examples and experiments with fixed-wing aircraft are carried out in Section \ref{sec:simexp}. Finally, Section \ref{sec:conclusion} concludes the paper and indicates future work.

\section{Preliminaries} \label{sec:preliminary}
\subsection{Notations}
The set of integers $\{m \in \mathbb{Z} : i \le m \le j \}$ is denoted by $\mathbb{Z}_{i}^{j}$. We use boldface for a vector $\bm{v} \in \mbr[n]$, and its $j$-th entry is denoted by $v_j$ for $j \in \mathbb{Z}_{1}^{n}$. Consider a system consisting of $N$ robots. Any quantity associated with the $i$-th robot is symbolized by the superscript $(\cdot)^{[i]}$. For example, the notation $\bm{u^{[i]}} \in \mbr[n]$ denotes a vector associated with the $i$-th robot for $i \in \mathbb{Z}_{1}^{N}$, and the $j$-th entry of it is denoted by  $u^{[i]}_{j}$ for $j\in \mathbb{Z}_{1}^{n}$ (note that $u^{[i]}_{j}$ is not boldfaced since it is a scalar). The notation ``$\defeq$'' means ``defined to be'', and $\diag{\cdot}$ denotes a (block) diagonal matrix obtained by putting vectors or matrices on the diagonal. If $\set{A}$ is a finite set, then $|\set{A}|$ denotes its cardinality (i.e., the number of elements in the set).

\subsection{Graph theory}
This subsection follows the convention in \cite{mesbahi2010graph}. A finite, \emph{undirected  graph} (or \emph{graph} for short) is a set-theoretic object $\set{G}=(\set{V}, \set{E})$, where the \emph{vertex set} $\set{V} \defeq \{1,\dots,N\}$ contains a finite set of elements, called \emph{vertices}, and the \emph{edge set} $\set{E}$ is a subset of $\set{V} \times \set{V}$, of which the elements are denoted by $(i,j)$, representing the \emph{adjacent} relation between vertices $i$ and $j$ for $i, j \in \set{V}$. The $k$-th edge is denoted by $\mathcal{E}_k = (\mathcal{E}_k^{\text{head}}, \mathcal{E}_k^{\text{tail}})$, where $\mathcal{E}_k^{\text{head}}, \mathcal{E}_k^{\text{tail}} \in \mathcal{V}$ are the \emph{head} and \emph{tail} of the edge, respectively. The set of neighboring robots of robot $i$ is denoted by $\set{N}_i \defeq \{j \in \set{V} : (i, j) \in \set{E}\}$. The graph $\set{G}$ is \emph{connected} if there is a path between any pair of vertices in $\set{V}$. The \emph{adjacency matrix} $A(\set{G})$ of the undirected graph $\set{G}$ is the symmetric $N \times N$ matrix encoding the adjacency relationships of vertices; that is, $[A(\set{G})]_{ij}=1$ if $(i, j) \in \set{E}$ and $[A(\set{G})]_{ij}=0$ otherwise. The \emph{Laplacian matrix} $L(\set{G})$ of $\set{G}$ is the $N \times N$ matrix defined by $[L(\set{G})]_{ij}=-a_{ij}$ for $i \ne j$ and $[L(\set{G})]_{ii}=\sum_{k=1}^N a_{ik}$ for $i\in \mathbb{Z}_{1}^{N}$, where $a_{ij}$ is the $ij$-th entry of the adjacency matrix. For an undirected graph, we define the entries of the incidence matrix $B\in\mathbb{R}^{|\mathcal{V}|\times|\mathcal{E}|}$ by $b_{ik}=+1$ if $i = {\mathcal{E}}^{\text{head}}_k$, $b_{ik}=-1$ if $i = \mathcal{E}^\text{tail}_k$, and $b_{ik}=0$ otherwise.

The $N$-cycle graph $C_N=(\{1,\dots,N\}, \set{E}_C)$ is the graph where the edge set is $\set{E}_C = \{(1,2),(2,3),\dots, (N-2, N-1), (N-1, N), (N,1) \}$. Namely, the $N$ robots are connected in sequence to form a cycle. 

\subsection{Guiding vector field with a virtual coordinate for a single robot} \label{subsec:pre_gvf}
Guiding vector fields with virtual coordinates for robot navigation enjoy many features, such as guaranteeing global convergence to the desired path and enabling self-intersected desired path following \cite{yao2021singularity,yao2020mobile,yaothesis2021}. We present here a brief introduction. Suppose the (physical)  desired path $\phy{\set{P}}$ is parameterized by
\[
x_{1} = f_1(w), \dots, x_{n}=f_n(w),
\]
where $x_j$ is the $j$-th coordinate, $w \in \mbr[]$ is the parameter of the path and $f_j$ is twice continuously differentiable (i.e., $f_j \in C^2$), for $j \in \mathbb{Z}_{1}^{n}$. To derive a corresponding guiding vector field for the desired path, we need to describe the desired path as the intersection of $n$ hyper-surfaces \cite{yao2018cdc,Goncalves2010}. To this end, taking $w$ as an additional argument, we define $n$ \emph{surface functions} $\phi_j: \mbr[n+1] \to \mbr$ as follows: $\phi_1(\bm{\xi})=x_{1}-f_1(w), \dots, \phi_n(\bm{\xi})=x_{n} - f_n(w)$, where $\bm{\xi}=(x_1,\dots,x_n,w) \in \mbr[n+1]$ is the generalized coordinate with an additional \emph{virtual coordinate} $w$. Therefore, the desired path with an additional coordinate is the intersection of the hyper-surfaces described by the zero-level set of these functions. Namely,
\[
\hgh{\set{P}}\defeq \{ \bm{\xi} \in \mbr[n+1]: \phi_j(\bm{\xi})=0, j \in \mathbb{Z}_{1}^{n} \}.
\]
The projection of the higher-dimensional $\hgh{\set{P}}$ onto the hyper-plane spanned by the first $n$ coordinates is the original (physical) desired path $\phy{\set{P}}$, so we can use the higher-dimensional guiding vector field $\vf: \mbr[n+1] \to \mbr[n+1]$ corresponding to $\hgh{\set{P}}$ to follow the original (physical) desired path $\phy{\set{P}}$ by using a projection technique. The vector field $\vf$ is defined by
\begin{equation} \label{eq_gvf}
\vf(\bm{\xi}) = \times \big( \bm{\nabla \phi_1},\dots, \bm{\nabla \phi_n} \big) - \sum_{j=1}^{n} k_j \phi_j \bm{\nabla \phi_j},
\end{equation}
where  $k_j$ is a positive gain, $\bm{\nabla \phi_j} \in \mbr[n+1]$ is the gradient of $\phi_j$ with respect to its generalized coordinate $\bm{\xi}$, and the first term is the wedge product \cite[Eq. (5)]{yao2021singularity} of all the gradients $\bm{\nabla \phi_j} \in \mbr[n+1], j \in \mathbb{Z}_{1}^{n}$ (it degenerates to the cross product if $n=2$). The physical interpretation of the vector field $\vf$ is clear: the second term $ - \sum_{j=1}^{n} k_j \phi_j \bm{\nabla \phi_j}$ is a weighted sum of all the gradients, which guides the trajectory towards the intersection of the hyper-surfaces (i.e., the desired path), while the first term $\times \big( \bm{\nabla \phi_1},\dots, \bm{\nabla \phi_n} \big)$, being orthogonal to all the gradients $\bm{\nabla \phi_j}$ \cite[Proposition 7.2.1]{galbis2012vector}, provides a propagation direction along the desired path.

\begin{remark}
	The virtual coordinate $w$ can be seen as a ``carrot'' tracked by robots. Notwithstanding, the dynamics of the ``carrot'' are in closed loop with the positions of the robots, and thus it may go backwards on the desired path. The detail is illustrated in \cite[Section VII]{yao2021singularity} and its supplementary video. 
	Similarly, another approach to generate vector fields from the parametric equations of desired paths has recently been proposed in \cite{rezende2022ConstructiveTimeVaryingVector}, which is able to deal with time-varying desired paths and enables the isotropic-in-space convergence. However, \cite{rezende2022ConstructiveTimeVaryingVector} only focused on the single-robot case.  We will show in the sequel how the guiding vector field introduced in this subsection can be extended for multi-robot motion coordination on desired paths and surfaces with collision avoidance. Additionally, our approach can deal with self-intersecting desired paths.
\end{remark}

\section{Distributed motion coordination on desired paths using guiding vector fields} \label{sec:pathfollowing}

In our previous work \cite{yao2021singularity,yao2020mobile}, the higher-dimensional guiding vector field in \eqref{eq_gvf} is proved to possess no singular points where the vector field becomes zero, thanks to the additional dimension (i.e., the additional virtual coordinate $w$). However, like many of other current studies (e.g., \cite{Goncalves2010,nelson2007vector}), the vector field is used for the guidance of only \emph{one single} robot. We aim to extend the vector field to include a \emph{coordination component} and achieve motion coordination among multiple robots. Since the additional virtual coordinate $w$ not only helps eliminate singular points, but also acts as the path parameter, one idea is to utilize the virtual coordinate $w$ to coordinate the robots' motions. Thus, the question is:  Suppose there are $N > 1$ robots, and the communication topology is described by an undirected graph $\set{G}=(\set{V},\set{E})$, where $\set{V}=\{1,\dots,N\}$ represent these $N$ robots; if $(i, j) \in \set{E}$, then information can flow between Robot $i$ and Robot $j$. Based on the higher-dimensional vector field in \eqref{eq_gvf}, how to design an extra \emph{coordination mechanism} using the virtual coordinates $w^{[i]}$, $i\in \mathbb{Z}_{1}^{N}$, such that:

\begin{enumerate}
	\item Each robot can follow their desired paths;
	\item All robots can coordinate their motions by controlling the virtual coordinates $w^{[i]},i\in \mathbb{Z}_{1}^{N}$, via \emph{local} information exchange among neighboring robots?
\end{enumerate}

The extra coordination mechanism and the precise meaning of motion coordination will become clear after we present the problem formulation using the dynamical systems theory.

\subsection{Mathematical problem formulation} \label{sec:pfproblemform}

Suppose the $i$-th robot is required to follow a path in $\mbr[n]$, parameterized by $n$ parametric equations:
\begin{equation} \label{eq_param_func}
\mx{i}{1} = \mf{i}{1}(w^{[i]})	\quad \dots  \quad 	 \mx{i}{n} = \mf{i}{n}(w^{[i]}),
\end{equation}
where $\mx{i}{j}$ is the $j$-th coordinate, $\mf{i}{j} \in C^2$ is the $j$-th parametric function for the $i$-th robot, $i\in \mathbb{Z}_{1}^{N}, j\in \mathbb{Z}_{1}^{n}$, and $w^{[i]}$ is the parameter of the desired path. To derive the path-following guiding vector field, we use the parameter $w^{[i]}$ as an additional virtual coordinate, and the higher-dimensional desired path is described by
\[
\set{P}^{[i]} \defeq \{ \bm{\xi^{[i]}} \in \mbr[n+1]: \mphi{i}{1}(\bm{\xi^{[i]}})=0, \dots,  \mphi{i}{n}(\bm{\xi^{[i]}})=0 \},
\]
where $\bm{\xi^{[i]}} \defeq (\mx{i}{1}, \dots, \mx{i}{n}, w^{[i]}) \in \mbr[n+1]$ denotes the \emph{generalized coordinate} of the $i$-th robot. Note that the $(n+1)$-th entry of $\bm{\xi^{[i]}}$ is the additional virtual coordinate $w^{[i]}$. As before, the surface functions are defined by 
\[
	\mphi{i}{j}(\mx{i}{1},\dots,\mx{i}{n}, w^{[i]})=\mx{i}{j}-\mf{i}{j}(w^{[i]})
\]
for $i\in \mathbb{Z}_{1}^{N}$ and $j\in \mathbb{Z}_{1}^{n}$. Let
\[
	\bm{\Phi^{[i]}}(\bm{\xi^{[i]}}) \defeq \transpose{(\mphi{i}{1}(\bm{\xi^{[i]}}), \dots, \mphi{i}{n}(\bm{\xi^{[i]}}))} \in \mbr[n].
\]
Observe that $\bm{\xi^{[i]}} \in \set{P}^{[i]}$ if and only if $\norm{\bm{\Phi^{[i]}}(\bm{\xi^{[i]}})}=0$. Therefore, we can use $\bm{\Phi^{[i]}}(\bm{\xi^{[i]}})$ to quantify the distance to the desired path $\set{P}^{[i]}$. In the context of path following, we call $\bm{\Phi^{[i]}}(\bm{\xi^{[i]}})$ the \emph{path-following error} to $\set{P}^{[i]}$. The aim is to design controllers such that the norm $\norm{\bm{\Phi^{[i]}}(\bm{\xi^{[i]}})}$ converges to zero eventually. By combining \eqref{eq_gvf} and \eqref{eq_param_func}, we obtain the analytic expression of the \emph{path-following guiding vector field} $\bm{\vfpf^{[i]}}: \mbr[n+1] \to \mbr[n+1]$ for the $i$-th robot:
\begin{align} \label{eq_vfpf}
\bm{\vfpf^{[i]}}(\bm{\xi^{[i]}}) = \matr{ (-1)^n \mfdot{i}{1}(w^{[i]}) - \mk{i}{1} \mphi{i}{1}(\bm{\xi^{[i]}}) \\  
	\vdots \\ 
	(-1)^n \mfdot{i}{n}(w^{[i]}) - \mk{i}{n} \mphi{i}{n}(\bm{\xi^{[i]}}) \\ 
	(-1)^n + \sum_{l=1}^{n} \mk{i}{l} \mphi{i}{l}(\bm{\xi^{[i]}}) \mfdot{i}{l}(w^{[i]}) } %
\end{align}
for $i\in \mathbb{Z}_{1}^{N}$, where $\mk{i}{j}>0$ are constant gains, and $\mfdot{i}{j}$ are the derivatives of ${\mf{i}{j}}$ with respect to the argument $w^{[i]} \in \mbr[]$.%

To achieve coordination in $w^{[i]}$, thus indirectly coordinate the positions of robots, we introduce a new concept: a \emph{coordination component}. The coordination component $\bm{\vfcr^{[i]}}: \mbr[] \times \mbr[N] \to \mbr[n+1]$ for the $i$-th robot, $i\in \mathbb{Z}_{1}^{N}$, is:
\begin{align} \label{eq_vfco}
\bm{\vfcr^{[i]}}(t, \bm{w}) = \transpose{\big( 0, \cdots, 0, c^{[i]}(t, \bm{w}) \big)} \in \mbr[n+1], 
\end{align}
where $\bm{w}=\transpose{(w^{[1]}, \dots, w^{[N]})}$, and $c^{[i]}: \mbr[] \times \mbr[N] \to \mbr$ is called the \emph{coordination function} to be designed later\footnote{Although the argument of the coordination function $\mc{i}{}(\cdot, \cdot)$ contains all the virtual coordinates $\mw{j}{}$, $j\in \mathbb{Z}_{1}^{N}$, this is only for a compact and readable notation. The coordination function for each individual robot only requires the knowledge of the neighbors' $\mw{j}{}$ for $j \in \mathcal{N}_{i}$ as shown later.}, which enables coordination among robots through the local interactions via the neighboring virtual coordinates $w^{[j]}$ for $j \in \mathcal{N}_{i}$.  Specifically, we want the virtual coordinates of multiple robots $w^{[i]}(t) - w^{[j]}(t)$ to converge to $\Delta^{[i,j]}(t)$ for $(i, j) \in \set{E}$, where $\Delta^{[i,j]}(t) \in \mbr$ are real-valued continuously differentiable functions, representing the desired differences between $w^{[i]}(t)$ and $w^{[j]}(t)$ at time $t$, satisfying $\Delta^{[i,j]}(t)=-\Delta^{[j,i]}(t)$. It is naturally assumed that $\Delta^{[i,j]}(t)$ are chosen appropriately such that the resulting formation is feasible\footnote{If the communication topology does not contain any cycles, then arbitrary values of the desired parametric differences $\Delta^{[i,j]}$ are feasible as long as $\Delta^{[i,j]}=-\Delta^{[j,i]}$ is satisfied for $i,j \in \mathbb{Z}_{1}^{N}$; otherwise, one needs to examine the (typically geometric) feasibility of the resulting formation.} at any time $t \ge 0$. 

We design the $i$-th \emph{coordinating guiding vector field} $\bm{\vfcb^{[i]}}: \mbr[] \times \mbr[n+1+N] \to \mbr[n+1]$ to be the weighted sum of the path-following vector field $\bm{\vfpf^{[i]}}$ and the coordination component $\bm{\vfcr^{[i]}}$ as below:
\begin{align}	\label{eq_vf_combined}
\bm{\vfcb^{[i]}}(t, \bm{\xi^{[i]}}, \bm{w}) = \bm{\vfpf^{[i]}}(\bm{\xi^{[i]}}) + k_c \bm{\vfcr^{[i]}}(t, \bm{w}),
\end{align}
where $k_c>0$ is a parameter to adjust the contribution of $\bm{\vfpf^{[i]}}$ and  $\bm{\vfcr^{[i]}}$ to $\bm{\vfcb^{[i]}}$. With a larger value of $k_c$, the motion coordination is achieved faster. The coordinating guiding vector field $\bm{\vfcb^{[i]}}$ represents the desired moving direction for Robot $i$, guiding the robot's motion. Thus it is imperative to study the guidance result, or precisely, the convergence results of the integral curves of the vector field $\bm{\vfcb^{[i]}}$ for $i\in \mathbb{Z}_{1}^{N}$. Precisely, we stack all the robot states as a vector $\bm{\xi} \defeq \transpose{(\transpose{\bm{\xi^{[1]}}}, \dots, \transpose{\bm{\xi^{[N]}}})} \in \mbr[(n+1)N]$, stack all the coordinating guiding vector fields as $\bm{\vfcb}(t, \bm{\xi}) \defeq \transpose{( {\bm{\vfcb^{[1]}}}^\top, \dots, {\bm{\vfcb^{[N]}}}^\top )} \in \mbr[(n+1)N]$, and study the integral curves of $\bm{\vfcb}(t, \bm{\xi})$; that is, the solutions (or trajectories) to the differential equation 
\begin{equation} \label{eq_ode}
\dot{\bm{\xi}} = \bm{\vfcb}(t, \bm{\xi}),
\end{equation}
given an initial condition $\bm{\xi_0} \in \mbr[(n+1)N]$ at $t=t_0 \ge 0$. Note that if the coordination function $c^{[i]}(t, \bm{w})$ in \eqref{eq_vfco} is time-invariant, it is not explicitly dependent on time $t$, and we can simply rewrite it to $c^{[i]}(\bm{w})$. In this case, the system \eqref{eq_ode} can be rewritten as $\dot{\bm{\xi}} = \bm{\vfcb}(\bm{\xi})$, which is an autonomous system; otherwise, it is a non-autonomous system \cite[Chapter 1]{khalil2002nonlinear}, which is more difficult to analyze and left for future work. Throughout this paper, we only consider a time-invariant coordination functions $c^{[i]}(\bm{w})$ and the autonomous system \eqref{eq_ode}.  
We can formally formulate the problem as follows:
\begin{problem}[Multi-robot path following] \label{problem1}
	Design the coordinating guiding vector field $\bm{\vfcb^{[i]}}$ in \eqref{eq_vf_combined} for $i\in \mathbb{Z}_{1}^{N}$, such that the trajectories of \eqref{eq_ode}, given an initial condition $\bm{\xi_0} \in \mbr[(n+1)N]$ at $t=t_0 \ge 0$, fulfill the following two control objectives: 
	\begin{enumerate}[leftmargin=*]
		\item (\textbf{Path Following}) Robot $i$'s path following error to its desired path $\set{P}^{[i]}$ converges to zero asymptotically for $i\in \mathbb{Z}_{1}^{N}$. That is, $\norm{\bm{\Phi^{[i]}}(\bm{\xi^{[i]}}(t))} \to 0$ as $t \to \infty$ for $i\in \mathbb{Z}_{1}^{N}$.
		\item (\textbf{Motion Coordination}) Each robot's motion is coordinated distributedly subject to the communication graph $\set{G}$ (i.e., Robot $i$ can communicate with Robot $j$ if and only if $(i, j) \in \set{E}$) such that their virtual coordinates satisfy $w^{[i]}(t) - w^{[j]}(t) - \Delta^{[i,j]} \to 0$  as $t \to \infty$ for $(i, j) \in \set{E}$.
	\end{enumerate}
\end{problem}
Given the path-following vector field in \eqref{eq_vfpf}, we will design the coordination function $c^{[i]}(\cdot)$ later such that it coordinates the robots' motion but does not affect the  path-following performance. We propose the following mild standing assumption:
\begin{assump} \label{assump_graph}
	The communication graph $\set{G}=(\set{V}, \set{E})$ is undirected and connected.%
\end{assump}
Assumption \ref{assump_graph} implies that if $(i, j) \in \set{E}$, then Robot $i$ and Robot $j$ can share information bidirectionally, and no robot is isolated from the multi-robot system (e.g., a cycle graph satisfies this assumption).

\subsection{Time-invariant coordination component and convergence analysis} \label{subsec:coco}
Observe that the coordination of the virtual coordinates directly affects the coordination of the positions of the robots implicitly, since the virtual coordinate corresponds to the parameter of a desired path. Motivated by this observation, we will design a time-invariant coordination component $c^{[i]}(\bm{w})$ in \eqref{eq_vfco}, and analyze the trajectories of \eqref{eq_ode}.

\subsubsection{Time-invariant coordination component}  \label{subsec:inv_coco}
Given the desired path $\set{P}^{[i]}$, we can design the \emph{desired parametric differences} $\Delta^{[i,j]}$ starting from a particular \emph{reference configuration}\footnote{Note that $\bm{w^*}$ is unknown by the robots. Instead, Robot $i$ only knows the desired parametric differences relative to its neighbors, i.e., $\Delta^{[i,j]}$ for $j \in \mathcal{N}_{i}$.} $\bm{w}^* := \transpose{(w^{[1]*}, \cdots, w^{[N]*})}$. Namely, let $\bm{\Delta}^*$ be the stacked vector of $\Delta^{[i,j]}, (i,j)\in\mathcal{E}$, then $\bm{\Delta}^* \defeq D^\top\bm{w}^*$, where $D \in \mbr[N \times |\set{E}|]$ is an incidence matrix obtained by assigning arbitrary orientations to the edges of the undirected graph\footnote{Different assignments of the edges' orientations only affect the signs of each entry in the incidence matrix $D$, while the desired parametric differences are still the same. In addition, the Laplacian matrix $L= D \transpose{D}$ introduced later remains the same independent of the orientation assignment.}. Now we propose to employ the following consensus control algorithm \cite[p. 25]{ren2008distributed}:
\begin{equation} \label{eq_coordination}
c^{[i]} =  - \sum_{j\in\mathcal{N}_{i}} \big( w^{[i]} - w^{[j]} - \Delta^{[i,j]} \big), \; \forall i \in \mathbb{Z}_{1}^{N}.
\end{equation}
Equation \eqref{eq_coordination} can be rewritten in a compact form as
\begin{equation} \label{eq_coordination_stack}
\bm{c}(\bm{w}) = -L (\bm{w} - \bm{w}^*) = -L \bm{\tilde w},
\end{equation}
where $\bm{c}(\bm{w})=\transpose{(c^{[1]}(\bm{w}),\dots,c^{[N]}(\bm{w}))}$, $L=L(\set{G}) =  D \transpose{D}$ is the Laplacian matrix and 
$
	\bm{\tilde{w}} \defeq \bm{w} - \bm{w}^*.
$
Combining \eqref{eq_vfpf}, \eqref{eq_vfco}, \eqref{eq_vf_combined} and \eqref{eq_coordination}, we attain the coordinating guiding vector field $\bm{\vfcb^{[i]}}$ for $i\in \mathbb{Z}_{1}^{N}$. The vector field $\bm{\vfcb^{[i]}}$ only takes as inputs Robot $i$'s own states and its neighbors' virtual coordinates $w^{[j]}$ for $j \in \mathcal{N}_{i}$.
\begin{remark} \label{remark_neighbor}
	From \eqref{eq_vfco}, \eqref{eq_vf_combined} and \eqref{eq_coordination}, one observes that information exchange only happens in the coordination component $c^{[i]}(\cdot)$. Notably, the communication cost is low: Robot $i$ transmits only a scalar $w^{[i]}$ to the neighboring Robot $j \in \mathcal{N}_{i}$.
\end{remark}

\subsubsection{Convergence analysis}  \label{sec_converge_invariant}
The convergence analysis of trajectories of \eqref{eq_ode} is nontrivial given that the right-hand side of \eqref{eq_ode} is \emph{not} a gradient of any potential function, since the path-following vector field in \eqref{eq_gvf} contains a wedge product of all the gradients. In this subsection, we show that the coordinating guiding vector field \eqref{eq_vf_combined} enables multiple robots to follow their desired paths while they are coordinated by the virtual coordinates such that $w^{[i]}(t) - w^{[j]}(t)$ converges to $\Delta^{[i,j]}$ for $(i,j) \in \set{E}$ as $t\to\infty$.%

Define the \emph{composite error vector} $\bm{e}$ to be 
\begin{equation} \label{eq_error_path}
	\bm{e} = \transpose{( \transpose{\bm{\Phi}}, \transpose{(\transpose{D} \bm{\tilde{w}})} )} \in \mbr[nN+|\set{E}|],
\end{equation}
where
\begin{equation} \label{eq_Phi}
	\bm{\Phi} \defeq \transpose{(\transpose{\bm{\Phi^{[1]}}},  \cdots, \transpose{\bm{\Phi^{[N]}}})} \in \mbr[nN]
\end{equation}
and recall that $D$ is the incidence matrix, which satisfies  $L = D D^\top$. One needs to prove that $\bm{e}(t) \to \bm{0}$ as $t \to \infty$ to show the effectiveness of the coordinating guiding vector field $\bm{\vfcb^{[i]}}$. This result is formally stated after the following assumption:
\begin{assump} \label{assump_bounded_f_deri}
	The first and second derivatives of ${\mf{i}{j}}(\cdot)$ in \eqref{eq_param_func} are bounded for $i\in \mathbb{Z}_{1}^{N}, j\in \mathbb{Z}_{1}^{n}$. %
\end{assump}
\begin{theorem}[Motion coordination on paths] \label{thm_vf_invariant}
	Under Assumptions \ref{assump_graph} and \ref{assump_bounded_f_deri}, and given constant desired parametric differences $\Delta^{[i,j]}$ for $(i,j) \in \set{E}$ , the coordinating guiding vector fields $\bm{\vfcb^{[i]}}$ for $i\in \mathbb{Z}_{1}^{N}$ designed by combining \eqref{eq_vfpf}, \eqref{eq_vfco}, \eqref{eq_vf_combined} and \eqref{eq_coordination} solve Problem \ref{problem1} globally in the sense that the initial composite error $\norm{\bm{e}(t=0)}$ in \eqref{eq_error_path} can be arbitrarily large. %
\end{theorem}
\begin{proof}
	See Appendix \ref{app_thm_vf_invariant}.
\end{proof}

\section{Distributed coordinated maneuvering on surfaces} \label{sec:surface}
In this section, we extend the previous results such that $N$ robots can converge to (possibly different) two-dimensional \emph{surfaces}, and maneuver and coordinate their motions according to some parameters of the surfaces. It is clear in the previous sections what is meant by following a (one-dimensional) desired path, but it is perhaps unclear what is meant by coordinated motion on (two-dimensional) surfaces. Therefore, we first clarify the meaning of coordinated motion on surfaces in Section \ref{subsec:surfcoordmotion}, and then mathematically formulate the problem in Section \ref{subsec:surfproblemform}, and derive the coordinating guiding vector field and conduct the convergence analysis in Section \ref{subsec:surf_inv_coco}.
\subsection{Coordinated motion in a desired set} \label{subsec:surfcoordmotion}
\subsubsection{Single robot trajectories in a desired set}
For simplicity, we first study trajectories of \emph{one} robot within a \emph{desired set} of any dimensions. Suppose the desired set is in the following parametric form:
\begin{equation} \label{eq_param}
x_1 = f_1(w_1,\dots, w_k), \, \dots \, ,  x_n =f_n (w_1,\dots, w_k),
\end{equation}
where $w_1, \dots, w_k \in \mathbb{R}$ are $k$ parameters, $n \in\mathbb{N}$ is the dimension of the ambient Euclidean space (mainly $n \in \{2,3\}$ for applications with mobile robots), $f_j$, $j \in \mathbb{Z}_{1}^{n}$, are twice continuously differentiable functions. Precisely, the desired set described by \eqref{eq_param} is $\set{M} \defeq \{(x_1,\dots, x_n) \in \mathbb{R}^n : \eqref{eq_param}, w_j \in \mathbb{R},\, j \in \mathbb{Z}_{1}^{k} \}$. Generically, if $k=1$, then $\set{M}=\set{P}$ is a desired path, which is one-dimensional (or roughly speaking, of one degree-of-freedom), while if $k=2$, then $\set{M}$ is a two-dimensional surface. The $k$ parameters $w_j$, $j \in \mathbb{Z}_{1}^{k}$, can be used as coordinates to localize a point in $\set{M}$. Therefore, one can represent the position of a robot in $\set{M}$ by a $k$-tuple denoted by $(w_1,\dots, w_k)_{\set{M}}$ in the parameter space. This paper mainly focuses on parametric paths and surfaces (i.e., $k=1$ or $k=2$), but the generalization to $k>2$ is obvious after we elaborate on the case of $k=2$ in this section.

Suppose $k = 1$ and $\set{M}$ is a desired path denoted by $\set{P}$. It is clear that if a robot starts from the position $(w_1)_{\set{M}}$ in $\set{P}$, then its trajectory (subject to directions) coincides with the desired path $\set{P}$, and the speed at which the trajectory extends along the desired path is determined by the time derivative $\dot{w}_1 \in\mathbb{R}$ along the trajectory. Suppose $k = 2$ and $\set{M}$ is a surface denoted by $\set{S}$. If a robot starts from the position $(w_1,w_2)_{\set{M}}$ in $\set{S}$, it is the ratio between the different velocities $\dot{w}_1$ and $\dot{w}_2$ that determines the eventual trajectory of the robot in $\set{S}$. %
\begin{example}
	Suppose the desired set $\set{M}$ is the sphere parameterized by
	\begin{equation} \label{eq: Psphere}
	\set{S}_{\text{sph}} :=
	\begin{cases}
	x_1 = \cos(w_1)\cos(w_2) \\
	x_2 = \cos(w_1)\sin(w_2) \\
	x_3 = \sin(w_1)
	\end{cases}.
	\end{equation}
	If $\dot{w}_1 = 0$ and $\dot{w}_2 \ne 0$, then the robot's trajectory is parallel to the sphere's equator starting from an arbitrary point in $\set{S}_{\text{sph}}$. 
\end{example}
We will show later that our proposed coordinating guiding vector field allows one to set the values of $\dot{w}_j$ for $j \in \mathbb{Z}_{1}^{k}$, so we can guarantee that a robot moves as expected within $\set{M}$.
\subsubsection{Coordinated motion of multiple robots}
Consider $N > 1$ robots. The coordinated motion among multiple robots is characterized by the predefined parametric relations $\Delta^{[i,j]}_{(\cdot)}$, $i \ne j \in \mathbb{Z}_{1}^{N}$, between neighboring robots, for which the neighboring relations are encoded in the undirected communication graph $\mathcal{G}$. Note that the subscript in $\Delta^{[i,j]}_{(\cdot)}$ specifies which parameter of the desired set is considered (in the case of one parameter, this subscript is omitted for simplicity).

In the previous section, we only consider one parameter (i.e., $k=1$) $w_1$ to be communicated among robots. In this case, these robots exhibit the behavior of \emph{chasing each other} on one-dimensional desired paths $\set{P}^{[i]}$. Dealing with a desired set $\set{M}$ with $k>1$ gives us more freedom for the coordinated motion of robots. For simplicity, we consider the case of $k=2$ (i.e., the desired set $\set{M}$ is a two-dimensional surface), in the following example.
\begin{figure}[tb]
	\centering
	\subfigure[]{
		\includegraphics[width=0.47\columnwidth]{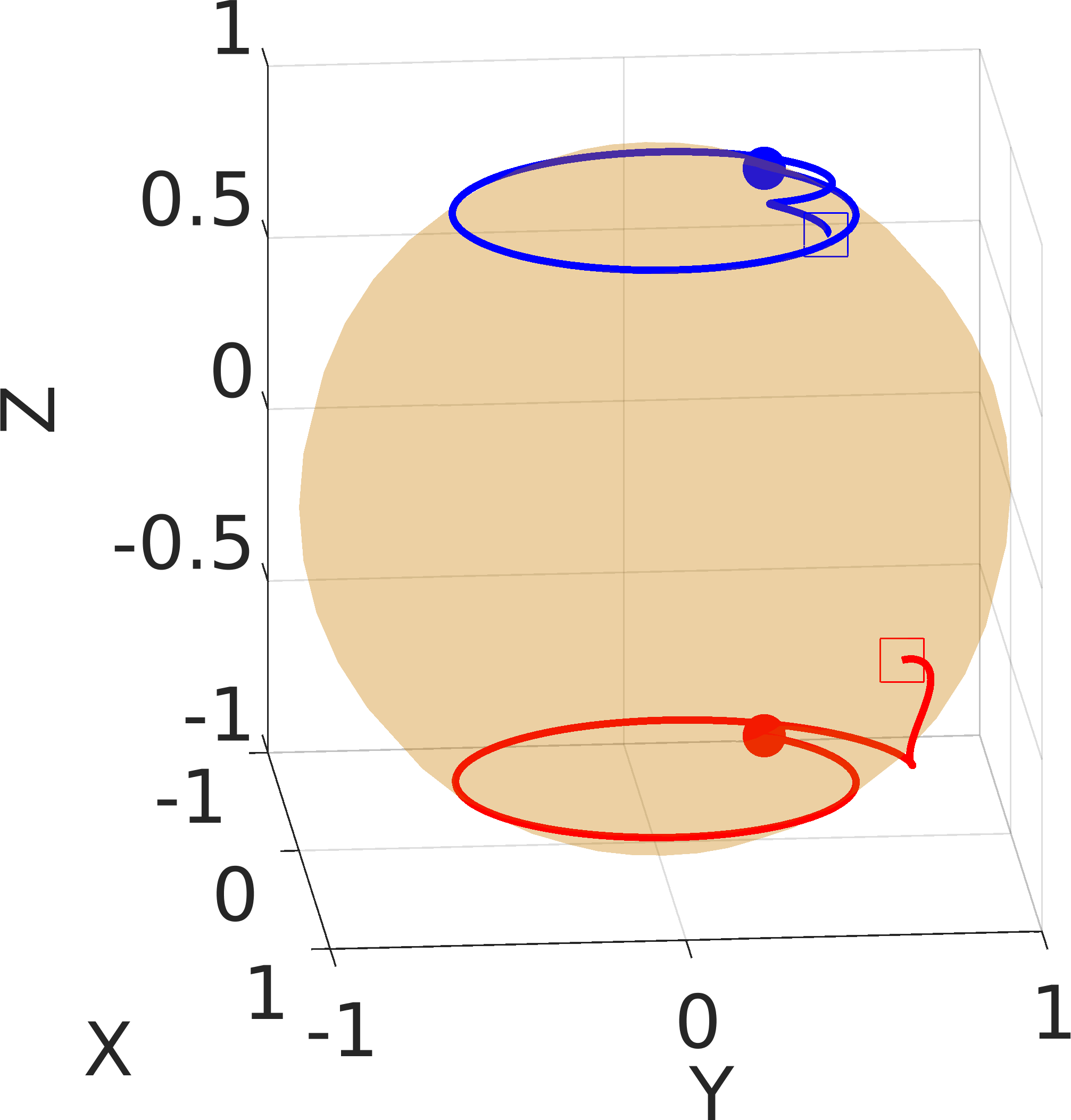}
		\label{fig:h6ex_ode20210614_1412_n3N2_ex2_1}
	}\hspace{-1em}
	\subfigure[]{
		\includegraphics[width=0.47\columnwidth]{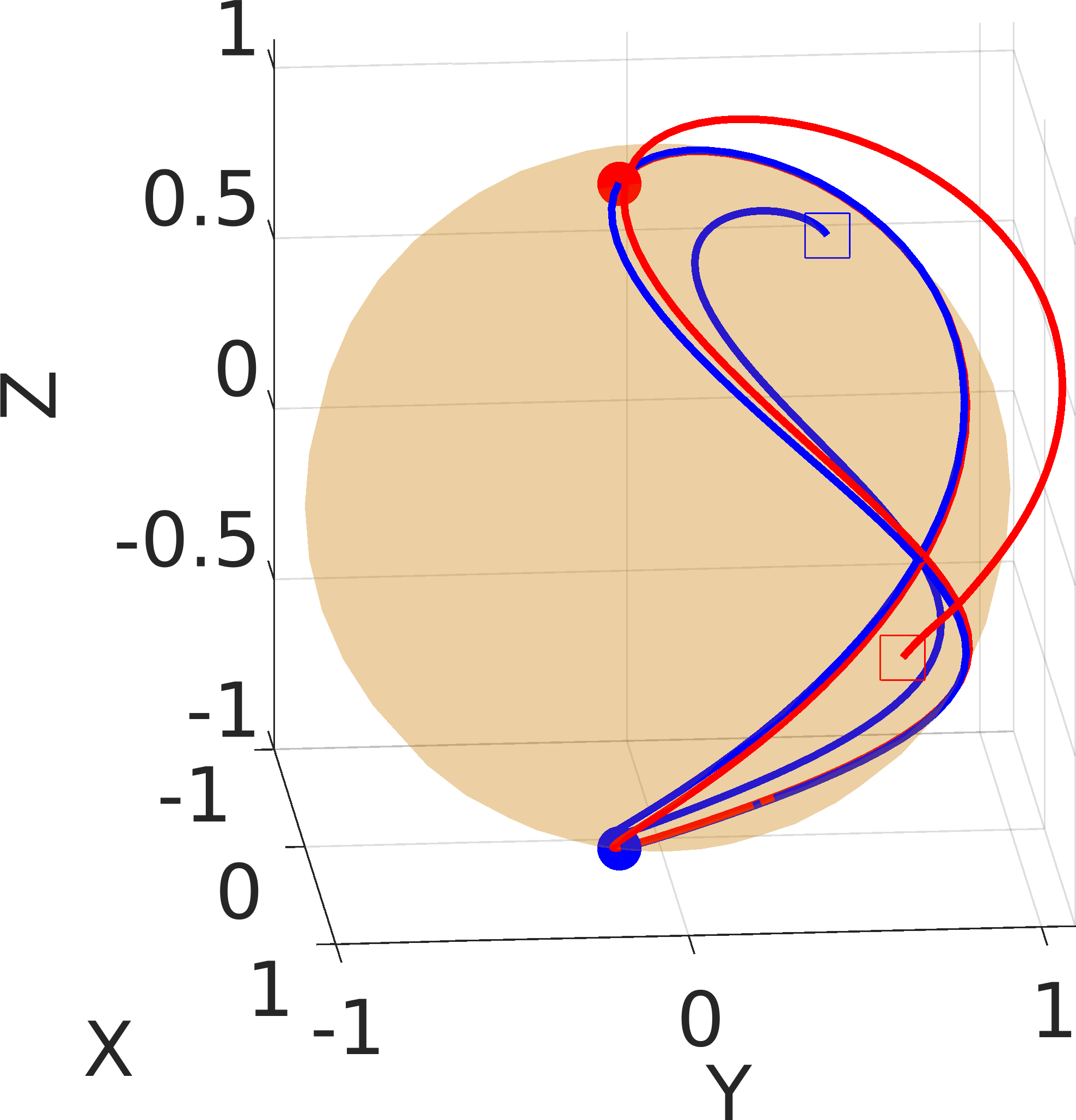}
		\label{fig:h6ex_ode20210614_1412_n3N2_ex2_2}
	}
	\caption{Simulation results of two robots coordinating their motions on the sphere described by \eqref{eq: Psphere}. The blue and red dots represent Robot $1$ and $2$, respectively, and the square symbols represent the initial positions. The desired parametric differences are $\Delta^{[1,2]}_1=\Delta^{[1,2]}_2=-\pi$. \subref{fig:h6ex_ode20210614_1412_n3N2_ex2_1} The desired parametric speeds are $\dot{w}^{*}_1=0$ and $\dot{w}^{*}_2=1$ for both robots. \subref{fig:h6ex_ode20210614_1412_n3N2_ex2_2} The desired parametric speeds are $\dot{w}^{*}_1=\dot{w}^{*}_2=1$ for both robots. }
	\label{fig:h6exode202106141412n3n2ex21}
\end{figure}
\begin{example}	 \label{example2}
	Two robots coordinate their motions in the desired set: the sphere described by \eqref{eq: Psphere}. Specifically, we require that the difference between each of the two parameters, $\mw{1}{1}$ and $\mw{1}{2}$, of Robot $1$, and each of those of Robot $2$, $\mw{2}{1}$ and $\mw{2}{2}$, is $-\pi$.  Namely, the desired parametric differences are $\Delta^{[1,2]}_1=\Delta^{[1,2]}_2=-\pi$. We also require that, at the steady state, they move at specific parametric speeds: The desired speeds in the first and the second parameter are $\dot{w}^{*}_1=0$ and $\dot{w}^{*}_2=1$, respectively in the first scenario, and  $\dot{w}^{*}_1=\dot{w}^{*}_2=1$ in the second scenario for both robots. We use the coordinating guiding vector field that will be introduced later to produce the simulation results shown in Fig. \ref{fig:h6exode202106141412n3n2ex21}. %
\end{example}
In the next subsection, we formally define the problem of motion coordination of robots on surfaces, and provide the technical detail of the coordinating guiding vector field for surface navigation as employed in Example \ref{example2}.

\subsection{Mathematical problem formulation} \label{subsec:surfproblemform}
In previous sections, the $i$-th robot is required to follow a one-dimensional desired path, which is parameterized by one parameter $w^{[i]}$. In contrast, to converge to a two-dimensional surface, we need to use two parameters denoted by $\mw{i}{1}$ and $\mw{i}{2}$, respectively. Specifically, suppose the $i$-th robot is required to converge to a two-dimensional surface $\set{S}^{[i]} \subseteq \mbr[n]$, which is parameterized by $n$ parametric equations:
\begin{equation} \label{eq_param_func_surf}
\mx{i}{1} = \mf{i}{1}(\mw{i}{1},\mw{i}{2})	\quad \dots  \quad 	 \mx{i}{n} = \mf{i}{n}(\mw{i}{1},\mw{i}{2}),
\end{equation}
where $\mx{i}{j}$ is the $j$-th coordinate, $\mf{i}{j} \in C^2$ is the $j$-th parametric function for the $i$-th robot, for $i\in \mathbb{Z}_{1}^{N}, j\in \mathbb{Z}_{1}^{n}$, and $\mw{i}{1},\mw{i}{2}$ are two parameters of the surface. To derive the corresponding guiding vector field, we use the parameters $\mw{i}{1}$ and $\mw{i}{2}$ as two additional virtual coordinates, and the surface $\set{S}^{[i]}$ is described by
\[
\set{S}^{[i]} \defeq \{ \bm{\xi^{[i]}} \in \mbr[n+2]: \mphi{i}{1}(\bm{\xi^{[i]}})=0, \dots,  \mphi{i}{n}(\bm{\xi^{[i]}})=0 \},
\]
where $\bm{\xi^{[i]}} \defeq (\mx{i}{1}, \dots, \mx{i}{n}, \mw{i}{1}, \mw{i}{2}) \in \mbr[n+2]$ denotes the \emph{generalized coordinate} of the $i$-th robot. Note that the $(n+1)$-th and the $(n+2)$-th entries of $\bm{\xi^{[i]}}$ are the additional virtual coordinates $\mw{i}{1}$ and $\mw{i}{2}$, respectively. The functions $\mphi{i}{j}$ are 
\[
\mphi{i}{j}(\mx{i}{1},\dots,\mx{i}{n}, \mw{i}{1}, \mw{i}{2})=\mx{i}{j}-\mf{i}{j}(\mw{i}{1}, \mw{i}{2})
\]
for $i\in \mathbb{Z}_{1}^{N}$ and $j\in \mathbb{Z}_{1}^{n}$. We define 
$
\bm{\Phi^{[i]}}(\bm{\xi^{[i]}}) \defeq \transpose{(\mphi{i}{1}(\bm{\xi^{[i]}}), \dots, \mphi{i}{n}(\bm{\xi^{[i]}}))} \in \mbr[n].
$
Observe that $\bm{\xi^{[i]}} \in \set{S}^{[i]}$ if and only if $\norm{\bm{\Phi^{[i]}}(\bm{\xi^{[i]}})}=0$. Therefore, we can similarly use $\bm{\Phi^{[i]}}(\bm{\xi^{[i]}})$, called the \emph{surface-convergence error}, to quantify the distance to the desired surface $\set{S}^{[i]}$. %

We use the same structure as the guiding vector field discussed before. However, since the dimensions of the states become $n+2$, the original wedge product is not well-defined. Therefore, we need to introduce an \emph{extra vector} $\bm{v}=(v_{1},\dots,v_{n+2})^\top \in \mbr[n+2]$ into the wedge product. Specifically, the new guiding vector field $\bm{\vfsf^{[i]}}: \mbr[n+2] \to \mbr[n+2]$ corresponding to the surface $\set{S}^{[i]}$, called the \emph{surface-navigation vector field}, is
\begin{multline} \label{eq_gvf_surf}
\bm{\vfsf^{[i]}}(\bm{\xi^{[i]}}) = \wedge(\bm{\nabla \mphi{i}{1}}(\bm{\xi^{[i]}}), \dots, \bm{\nabla \mphi{i}{n}}(\bm{\xi^{[i]}}), \bm{v}) - \\ \sum_{j=1}^{n} \mk{i}{j} \mphi{i}{j}(\bm{\xi^{[i]}})  \bm{\nabla \mphi{i}{j}}(\bm{\xi^{[i]}}).
\end{multline}
In theory, the extra vector $\bm{v} \in \mbr[n+2]$ can be arbitrarily chosen since the first term $\wedge(\bm{\nabla \mphi{i}{1}}(\bm{\xi^{[i]}}), \dots, \bm{\nabla \mphi{i}{n}}(\bm{\xi^{[i]}}), \bm{v})$ of \eqref{eq_gvf_surf} is always orthogonal to each of the scaled gradient $\mk{i}{j} \mphi{i}{j}(\bm{\xi^{[i]}})  \bm{\nabla \mphi{i}{j}}(\bm{\xi^{[i]}})$ in the second term of \eqref{eq_gvf_surf}, and this orthogonality is crucial for showing the convergence of robots' trajectories to the desired surface. However, for simplicity, we choose $v_{1} = \dots = v_{n} \equiv 0$ while $v_{n+1},v_{n+2}$ can be zero or non-zero constant values. This choice results in the following simple explicit expression of \eqref{eq_gvf_surf}:
\begin{multline} \label{eq_vfpf_surf}
\bm{\vfsf^{[i]}}(\bm{\xi^{[i]}}) 
= \\ 
	\scalemath{0.9}{
		(-1)^n \matr{ v_{n+2} \mfdotwone{i}{1} - v_{n+1} \mfdotwtwo{i}{1} \\
	\vdots \\
	v_{n+2} \mfdotwone{i}{n} - v_{n+1} \mfdotwtwo{i}{n} \\ 
	v_{n+2} \\
	-v_{n+1} } + 
\matr{-\mk{i}{1} \mphi{i}{1} \\ 
	\vdots \\
	-\mk{i}{n} \mphi{i}{n} \\
	\sum_{j=1}^{n} \mk{i}{j} \mphi{i}{j} \mfdotwone{i}{j} \\
	\sum_{j=1}^{n} \mk{i}{j} \mphi{i}{j} \mfdotwtwo{i}{j} },
}
\end{multline}
where $\mfdotwone{i}{j} \defeq \frac{\partial \mf{i}{j}}{\partial \mw{i}{1}}$ and $\mfdotwtwo{i}{j} \defeq \frac{\partial \mf{i}{j}}{\partial \mw{i}{2}}$ for $i \in \mathbb{Z}_{1}^{N}$ and $j \in \mathbb{Z}_{1}^{n}$. Thanks to this choice of the extra vector $\bm{v}$, it affects only the last two entries of the first term of \eqref{eq_vfpf_surf}. Once the robots are on the desired surface, the second term of \eqref{eq_vfpf_surf}  becomes zero. Therefore, the last two entries of the first term in \eqref{eq_vfpf_surf} equals the \emph{parametric speeds $\mdotw{i}{1}$ and $\mdotw{i}{2}$}, which represent how fast and in which direction the robots navigate on the surface in terms of the parameters $\mw{i}{1}$ and $\mw{i}{2}$.

To achieve coordination in $\mw{i}{1}$ and $\mw{i}{2}$, thus indirectly coordinate the positions of robots, we similarly introduce two time-invariant \emph{coordination components} $\bm{\vfcrone^{[i]}}, \bm{\vfcrtwo^{[i]}}: \mbr[N] \to \mbr[n+2]$ for the $i$-th robot:
\begin{subequations}\label{eq_vfco_surf}
	\begin{align} 
	&\bm{\vfcrone^{[i]}}(\bm{\mw{\cdot}{1}}) = \transpose{\big( 0, \cdots, 0, \mc{i}{1}(\bm{\mw{\cdot}{1}}), 0\big)},  \\ 
	&\bm{\vfcrtwo^{[i]}}(\bm{\mw{\cdot}{2}}) = \transpose{\big( 0, \cdots, 0, 0, \mc{i}{2}(\bm{\mw{\cdot}{2}}) \big)},  
	\end{align}
\end{subequations}
where $\bm{\mw{\cdot}{1}}=\transpose{(\mw{1}{1}, \dots, \mw{N}{1})}$, $\bm{\mw{\cdot}{2}}=\transpose{(\mw{1}{2}, \dots, \mw{N}{2})}$, and $\mc{i}{1}, \mc{i}{2}: \mbr[N] \to \mbr$ are the time-invariant \emph{coordination functions} to be designed later\footnote{As the first footnote, we do not require the knowledge of all virtual coordinates, but only those of the neighbors $\mw{j}{1}$ or $\mw{j}{2}$ for $j \in \mathcal{N}_{i}$.}, which enables coordination among robots through the local interactions via the \emph{neighboring} virtual coordinates $\mw{j}{1}$ and $\mw{j}{2}$ for $j \in \mathcal{N}_{i}$.  Specifically, we want the virtual coordinates of multiple robots $\mw{i}{1}(t) - \mw{j}{1}(t)$ and $\mw{i}{2}(t) - \mw{j}{2}(t)$ to converge to $\Delta^{[i,j]}_{1}$ and $\Delta^{[i,j]}_{2}$ for $(i, j) \in \set{E}$, where $\Delta^{[i,j]}_{1},\Delta^{[i,j]}_{2} \in \mbr$ are the desired differences between $\mw{i}{1}$ and $\mw{j}{1}$, satisfying $\Delta^{[i,j]}_{1}=-\Delta^{[j,i]}_{1}$ and $\Delta^{[i,j]}_{2}=-\Delta^{[j,i]}_{2}$. It is naturally assumed that $\Delta^{[i,j]}_{1}$ and $\Delta^{[i,j]}_{2}$ are chosen appropriately such that the resulting formation is feasible.

We design the $i$-th \emph{coordinating guiding vector field} $\bm{\vfcb^{[i]}}: \mbr[n+2N] \to \mbr[n+2]$ to be the weighted sum of the surface-navigation vector field $\bm{\vfsf^{[i]}}$ and the coordination components $\bm{\vfcrone^{[i]}}$ and $\bm{\vfcrtwo^{[i]}}$ as below:
\begin{multline}	\label{eq_vf_combined_surf}
\bm{\vfcb^{[i]}}(\bm{\xi^{[i]}}, \bm{\mw{\cdot}{1}}, \bm{\mw{\cdot}{2}}) = \bm{\vfsf^{[i]}}(\bm{\xi^{[i]}}) + k_{c1} \bm{\vfcrone^{[i]}}(\bm{\mw{\cdot}{1}}) + \\ k_{c2} \bm{\vfcrtwo^{[i]}}(\bm{\mw{\cdot}{2}}),
\end{multline}
where $k_{c1},k_{c2}>0$ are parameters to adjust the contribution of $\bm{\vfsf^{[i]}}$,  $\bm{\vfcrone^{[i]}}$ and $\bm{\vfcrtwo^{[i]}}$ to $\bm{\vfcb^{[i]}}$. 

We stack all the robot states as a vector $\bm{\xi} \defeq \transpose{(\transpose{\bm{\xi^{[1]}}}, \dots, \transpose{\bm{\xi^{[N]}}})} \in \mbr[(n+2)N]$ and stack all the coordinating guiding vector fields as $\bm{\vfcb}(t, \bm{\xi}) \defeq \transpose{( {\bm{\vfcb^{[1]}}}^\top, \dots, {\bm{\vfcb^{[N]}}}^\top )} \in \mbr[(n+2)N]$. We need to investigate the integral curves of $\bm{\vfcb}(\bm{\xi})$; i.e., the trajectories of the differential equation: $\dot{\bm{\xi}} = \bm{\vfcb}(\bm{\xi})$, given an initial condition $\bm{\xi_0} \in \mbr[(n+2)N]$ at $t=t_0 \ge 0$. %
Since the coordination functions in \eqref{eq_vfco_surf} are time-invariant, the system \eqref{eq_ode_surf} is an autonomous system: 
\begin{equation} \label{eq_ode_surf}
	\dot{\bm{\xi}} = \bm{\vfcb}(\bm{\xi}). 
\end{equation}
Now we can formally formulate the problem.
\begin{problem}[Multi-robot surface navigation] \label{problem1_surf}
	Design the coordinating guiding vector field $\bm{\vfcb^{[i]}}$ in \eqref{eq_vf_combined_surf} for $i\in \mathbb{Z}_{1}^{N}$, such that the trajectories of \eqref{eq_ode_surf}, given an initial condition $\bm{\xi_0} \in \mbr[(n+2)N]$ at $t=t_0 \ge 0$, fulfill the following three control objectives: 
	\begin{enumerate}
		\item (\textbf{Surface convergence}) Robot $i$'s surface-convergence errors tend to to zero asymptotically for $i\in \mathbb{Z}_{1}^{N}$. Namely, $\norm{\bm{\Phi^{[i]}}(\bm{\xi^{[i]}}(t)) } \to 0$ as $t \to \infty$ for $i\in \mathbb{Z}_{1}^{N}$.
		\item (\textbf{Motion coordination}) Each robot's motion is coordinated distributedly subject to the communication graph $\set{G}$ (i.e., Robot $i$ can communicate with Robot $j$ if and only if $(i, j) \in \set{E}$) such that their additional virtual coordinates satisfy $\mw{i}{1}(t) - \mw{j}{1}(t) - \Delta^{[i,j]}_{1} \to 0$ and $\mw{i}{2}(t) - \mw{j}{2}(t) - \Delta^{[i,j]}_{2} \to 0$ as $t \to \infty$ for $(i, j) \in \set{E}$.
		\item (\textbf{Surface maneuvering})  Given desired parametric speeds $\dot{w}^{*}_{1} \in \mbr[]$ and $\dot{w}^{*}_{2} \in \mbr[]$, the robot motion can achieve these speeds; i.e., $\mdotw{i}{1}(t) \to \dot{w}^{*}_{1}$ and $\mdotw{i}{2}(t) \to \dot{w}^{*}_{2}$ as $t \to \infty$ for $i \in \mathbb{Z}_{1}^{N}$.
	\end{enumerate}
\end{problem}
Given the surface-navigation vector field in \eqref{eq_vfpf_surf}, we will design the coordination function $\mc{i}{1}(\cdot), \mc{i}{2}(\cdot)$ later such that they coordinate the robots' motions but do not affect the surface-navigation performance.

\subsection{Time-invariant coordination component}  \label{subsec:surf_inv_coco}
In this subsection, we will design time-invariant coordination components $\mc{i}{1}(\bm{\mw{\cdot}{1}}),\mc{i}{2}(\bm{\mw{\cdot}{2}})$ and analyze the trajectories of \eqref{eq_ode_surf}, which is an autonomous system.

\subsubsection{Coordination component}
Given the desired surface $\set{S}^{[i]}$, we can design the desired parametric differences $\Delta^{[i,j]}_{1}$ and $\Delta^{[i,j]}_{2}$ starting from particular reference configurations $\bm{w^{*}_{1}} := \transpose{(\mw{1}{1}^*, \cdots, \mw{N}{1}^*)}$ and $\bm{w^{*}_{2}} := \transpose{(\mw{1}{2}^*, \cdots, \mw{N}{2}^*)}$. Hence, $\bm{\Delta}^{*}_{1} = D^\top \bm{w^{*}_{1}} \in \mbr[|\set{E}|]$ and $\bm{\Delta}^{*}_{2} = D^\top \bm{w^{*}_{2}} \in \mbr[|\set{E}|]$ are the stacked vectors of $\Delta^{[i,j]}_{1}, \Delta^{[i,j]}_{2}, (i,j)\in\mathcal{E}$, respectively, where $D \in \mbr[N \times |\set{E}|]$ is an incidence matrix. We employ the following consensus control algorithm:
\begin{subequations} \label{eq_coordination_surf}
	\begin{align} 
	\mc{i}{1} &=  - \sum_{j\in\mathcal{N}_{i}} \big( \mw{i}{1} - \mw{j}{1} - \Delta^{[i,j]}_{1} \big), \\
	\mc{i}{2} &=  - \sum_{j\in\mathcal{N}_{i}} \big( \mw{i}{2} - \mw{j}{2} - \Delta^{[i,j]}_{2} \big), 
	\end{align}
\end{subequations}
for $i \in \mathbb{Z}_{1}^{N}$. Equations \eqref{eq_coordination_surf} can be rewritten compactly as
\begin{subequations} \label{eq_coordination_stack_surf}
	\begin{align} 
	&\bm{\mc{\cdot}{1}}(\bm{\mw{\cdot}{1}}) = -L (\bm{\mw{\cdot}{1}} - \bm{w^{*}_{1}}) = -L \bm{\mtildew{\cdot}{1}}, \\
	&\bm{\mc{\cdot}{2}}(\bm{\mw{\cdot}{2}}) = -L (\bm{\mw{\cdot}{2}} - \bm{w^{*}_{2}}) = -L \bm{\mtildew{\cdot}{2}},
	\end{align}
\end{subequations}
where $\bm{\mc{\cdot}{1}}(\bm{\mw{\cdot}{1}})=(\mc{1}{1}(\bm{\mw{\cdot}{1}}),\dots,\mc{N}{1}(\bm{\mw{\cdot}{1}}))^{\top}$, $\bm{\mc{\cdot}{2}}(\bm{\mw{\cdot}{2}})=(\mc{1}{2}(\bm{\mw{\cdot}{2}}),\dots,\mc{N}{2}(\bm{\mw{\cdot}{2}}))^{\top}$, $L=L(\set{G})$ is the Laplacian matrix and 
$
\bm{\mtildew{\cdot}{1}} = \bm{\mw{\cdot}{1}} - \bm{w^{*}_{1}}$, 
$
\bm{\mtildew{\cdot}{2}} = \bm{\mw{\cdot}{2}} - \bm{w^{*}_{2}}.
$
Combining \eqref{eq_vfpf_surf}, \eqref{eq_vfco_surf}, \eqref{eq_vf_combined_surf} and \eqref{eq_coordination_surf}, we attain the coordinating guiding vector field $\bm{\vfcb^{[i]}}$ for $i\in \mathbb{Z}_{1}^{N}$. The vector field $\bm{\vfcb^{[i]}}$ only takes as inputs Robot $i$'s own states and its neighbors' virtual coordinates $\mw{j}{1}$, $\mw{j}{2}$ for $j \in \mathcal{N}_{i}$.
\begin{remark} \label{remark_neighbor_surf}
	As in Remark \ref{remark_neighbor}, from \eqref{eq_vfco_surf}, \eqref{eq_vf_combined_surf} and \eqref{eq_coordination_surf}, one observes that information exchange only involves the coordination components $\mc{i}{1}$, $\mc{i}{2}$, and Robot $i$ transmits only two scalars $\mw{i}{1}$, $\mw{i}{2}$ to the neighboring Robot $j \in \mathcal{N}_{i}$.
\end{remark}

\subsubsection{Convergence analysis}  
In this subsection, we show that the coordinating guiding vector field \eqref{eq_vf_combined_surf} enables multiple robots to navigate to their desired surfaces while they are coordinated by the virtual coordinates such that $\mw{i}{1}(t) - \mw{j}{1}(t)$ converges to $\Delta^{[i,j]}_{1}$, and $\mw{i}{2}(t) - \mw{j}{2}(t)$ converges to $\Delta^{[i,j]}_{2}$ for $(i,j) \in \set{E}$ as $t\to\infty$.
To achieve motion coordination on a surface, we impose the following mild assumption:
\begin{assump} \label{assump_bounded_f_deri_surf}
	The first derivatives $\mfdotwone{i}{j}(\cdot)\defeq \frac{\partial \mf{i}{j}}{\partial \mw{i}{1}}$, $\mfdotwtwo{i}{j}(\cdot)\defeq \frac{\partial \mf{i}{j}}{\partial \mw{i}{2}}$, and the second derivatives $\mfddotw{1}{1}{i}{j}(\cdot)$, $\mfddotw{2}{1}{i}{j}(\cdot)$, $\mfddotw{1}{2}{i}{j}(\cdot)$, $\mfddotw{2}{2}{i}{j}(\cdot)$, where $\partial_{w_{k}}^{w_{l}} \mf{i}{j}(\cdot) \defeq \frac{\partial^2 \mf{i}{j}}{\partial \mw{i}{l} \partial \mw{i}{k} }$ for $k,l \in \{1,2\}$, are bounded for all $i\in \mathbb{Z}_{1}^{N}, j\in \mathbb{Z}_{1}^{n}$. %
\end{assump}

Similarly, we define the \emph{composite error vector} $\bm{e}$ for the surface navigation to be 
\begin{equation} \label{eq_error_surf}
	\bm{e} = \transpose{( \transpose{\bm{\Phi}}, \transpose{(\transpose{D} \bm{\mtildew{\cdot}{1}})}, \transpose{(\transpose{D} \bm{\mtildew{\cdot}{2}})} )} \in \mbr[nN+2|\set{E}|].
\end{equation}	
Now we arrive at the following theorem:
\begin{theorem}[Motion coordination on surfaces] \label{thm_vf_invariant_surf}
	Under Assumptions \ref{assump_graph} and \ref{assump_bounded_f_deri_surf}, and given constant desired parametric differences $\Delta^{[i,j]}_1$, $\Delta^{[i,j]}_2$ for $(i,j) \in \set{E}$, the coordinating guiding vector fields $\bm{\vfcb^{[i]}}$ for $i\in \mathbb{Z}_{1}^{N}$ designed by combining \eqref{eq_vfpf_surf}, \eqref{eq_vfco_surf}, \eqref{eq_vf_combined_surf} and \eqref{eq_coordination_stack_surf}, and choosing
	\[
		\bm{v}=(-1)^{n+1} (0,\dots,0,\dot{w}^{*}_{2}, -\dot{w}^{*}_{1})
	\]
	in \eqref{eq_gvf_surf}, solve Problem \ref{problem1_surf} globally in the sense that the initial composite error $\norm{\bm{e}(t=0)}$ in \eqref{eq_error_surf} can be arbitrarily large. %
\end{theorem}
\begin{proof}
	See Appendix \ref{app:thm_vf_invariant_surf}.
\end{proof}
\begin{remark}
	The above analysis can be further extended for more parameters. For example, to define motion coordination confined in a three-dimensional manifold (e.g., a cube), one may choose three parameters $w_i, i=1,2,3$. The design methodology and analysis techniques for the two-parameter case can be directly adopted for this case with more cumbersome notations. In addition, if one uses $m \ge 2$ parameters, then one also needs to choose $m-1$ extra vectors $\bm{v_{1}}, \dots, \bm{v_{m-1}}$ as what has been shown in \eqref{eq_gvf_surf}, such that the wedge product is well-defined. %
\end{remark}

\section{Extending the vector field to incorporate collision avoidance} \label{sec:collision}
The focus of our work is the introduction and rigorous analysis of coordinating guiding vector fields for following paths or navigating surfaces with motion coordination as discussed above. However, to demonstrate the flexibility and practicality of our approach, in this section, we briefly explain how our proposed approach can incorporate an existing collision avoidance algorithm. Namely, we can modify the \emph{nominal} guiding vector field \eqref{eq_vf_combined} in a minimally invasive way using \emph{safety barrier certificates} \cite{wang2017safety}. Without loss of generality, we only consider the coordinating guiding vector field for path following. Namely, the number of parameters is $1$, and thus there is only one virtual coordinate for each robot. 

For $i,j\in \mathbb{Z}_{1}^{N}$, $i \ne j$, we define a function 
\[
	\mathfrak{h}_{ij}(\bm{\xi^{[i]}}, \bm{\xi^{[j]}}) = \norm{P(\bm{\xi^{[i]}}-\bm{\xi^{[j]}})}^2 - R^2,
\]
where $R>0$ is the safety distance, $P$ is the projection matrix $P = I - \left[ \begin{smallmatrix} \bm{0} & \bm{0} \\ \bm{0} & 1\end{smallmatrix} \right] \in \mbr[(n+1) \times (n+1)]$ and $I$ is the identity matrix. The function $\mathfrak{h}_{ij}$ reflects whether two robots keep a safety distance. Therefore, we can define the \emph{safety set} $\mathfrak{S}$:
\begin{multline}\label{eq_safeset}
	\mathfrak{S} = \{ ( {\bm{\xi^{[1]}}}^\top, \dots, {\bm{\xi^{[N]}}}^\top)^\top \in \mbr[(n+1)N] : \\ \mathfrak{h}_{ij}(\bm{\xi^{[i]}}, \bm{\xi^{[j]}}) \ge 0,  \forall i \ne j \}.
\end{multline}
Using  $\mathfrak{h}_{ij}$, we can define a barrier function \cite{ames2016control}:
\begin{equation} \label{eq_cbf}
	B_{ij}(\bm{\xi^{[i]}}, \bm{\xi^{[j]}}) \defeq \frac{1}{\mathfrak{h}_{ij}(\bm{\xi^{[i]}}, \bm{\xi^{[j]}})}
\end{equation}
for $i \ne j$, $i,j\in \mathbb{Z}_{1}^{N}$. We want to modify the \emph{nominal vector field} $\bm{\vfcb^{[i]}}$ such that the new vector field denoted by $\bm{\vfcbnew^{[i]}}$ also enables collision-avoidance, which is encoded in the safety barrier certificate. Specifically, the new vector field $\bm{\vfcbnew^{[i]}}$ is obtained by minimizing $\frac{1}{2}\sum_{i=1}^{N} \norm{\bm{\vfcbnew^{[i]}}-\bm{\vfcb^{[i]}}}^2$ subject to the constraint 
$
	\dot{B}_{ij} \le \alpha / B_{ij} 
$
where $\alpha>0$ is a constant, for $j<i$ and $i,j\in \mathbb{Z}_{1}^{N}$. Using \eqref{eq_cbf} and the equivalence
$
	\dot{B}_{ij} = - \dot{\mathfrak{h}}_{ij} / \mathfrak{h}_{ij}^2 =  - [2 (\bm{\xi^{[i]}}-\bm{\xi^{[j]}})^\top P^\top (\bm{\vfcbnew^{[i]}} - \bm{\vfcbnew^{[j]}} )] / \mathfrak{h}^2_{ij},
$
the constraint $\dot{B}_{ij} \le \alpha / B_{ij} $ can be rewritten as 
\begin{equation} \label{eq_c_constraint}
	 (\bm{\xi^{[j]}}-\bm{\xi^{[i]}})^\top P^\top \bm{\vfcbnew^{[i]}} + (\bm{\xi^{[i]}}-\bm{\xi^{[j]}})^\top P^\top \bm{\vfcbnew^{[j]}} \le \frac{\alpha}{2} \mathfrak{h}^3_{ij}.
\end{equation}
Therefore, the new vector field $\bm{\vfcbnew} \defeq ( {\bm{\vfcbnew^{[1]}}}^\top, \dots, {\bm{\vfcbnew^{[N]}}}^\top )^\top \in \mbr[(n+1)N]$ is calculated by the quadratic program below:
\begin{equation} \tag{QP1} \label{eq_qp1}
	\begin{split}
		\min_{\bm{\vfcbnew} \in \mbr[(n+1)N]} \;& \frac{1}{2} \sum_{i=1}^{N} \norm{\bm{\vfcbnew^{[i]}}-\bm{\vfcb^{[i]}}}^2  \\
		\text{s.t.} &\quad \eqref{eq_c_constraint}, \quad\forall j \in \set{D}_{i},
	\end{split}
\end{equation}
where $\set{D}_{i}$ is the set of robots within distance $R$ with respect to Robot $i$; i.e., $\set{D}_{i} \defeq \{j \in \mathbb{Z}_{1}^{N} : \mathfrak{h}_{ij} \le 0, j \ne i \}$. Note that the definition of $\set{D}_{i}$ is based on geographical distances, different from the neighborhood $\mathcal{N}_{i}$, which is based on a communication graph $\set{G}$. If Robot $i$ keeps safety distances to all other robots (i.e., $\mathfrak{h}_{ij} \ge 0$ for all $j \ne i$), then the new vector field $\bm{\vfcbnew^{[i]}}$ is the same as the original one $\bm{\vfcb^{[i]}}$. In addition, since the virtual coordinate is not considered in the constraint \eqref{eq_c_constraint}, there always holds $\mvfcb{i}{n+1}=\vfcbnew^{[i]}_{n+1}$ for $i \in \mathbb{Z}_{1}^{N}$. Note that this optimization problem is \emph{centralized} in the sense that it requires the positions and the nominal vector fields of \emph{all} robots. We can change it to the following \emph{distributed} version:
\begin{equation} \tag{QP2} \label{eq_qp2}
	\begin{split}
	\min_{\bm{\vfcbnew^{[i]}} \in \mbr[n+1]} \;& \frac{1}{2} \norm{\bm{\vfcbnew^{[i]}}-\bm{\vfcb^{[i]}}}^2  \\
	\text{s.t.} \quad& (\bm{\xi^{[j]}}-\bm{\xi^{[i]}})^\top P^\top \bm{\vfcbnew^{[i]}} \le \frac{\alpha}{4} \mathfrak{h}^3_{ij}, \quad\forall j \in \set{D}_{i}
	\end{split}
\end{equation}
for each robot $i \in \mathbb{Z}_{1}^{N}$. In this optimization problem, Robot $i$ only needs to measure\footnote{In practice, the measurement is always inaccurate. However, the robustness property of control barrier functions against perturbation \cite{xu2015robustness} mitigate the consequence of inaccurate measurement.} the positions $\bm{p^{[j]}}=P \bm{\xi^{[j]}}$ of nearby robots for $j \in \set{D}_{i}$, and does \emph{not} need to obtain other robots' nominal vector field $\bm{\vfcb^{[j]}}$, which is only possible via communication. If all the constraints in \eqref{eq_qp2} are satisfied, so are those in \eqref{eq_qp1}. Therefore, a feasible solution of \eqref{eq_qp2} is also a feasible solution of \eqref{eq_qp1}. Assume that the initial positions of robots are in the safety set $\mathfrak{S}$, then it is guaranteed that all the robots will always be in the safety set $\mathfrak{S}$ \cite{wang2017safety}, and one observes that the quadratic program \eqref{eq_qp2} is always feasible, since $\bm{\vfcbnew^{[i]}}=\bm{0}$ remains a trivial solution. This implies that a robot can always keep stationary (i.e., $\dot{\bm{p}}^{[i]} = P \bm{\vfcbnew^{[i]}}=\bm{0}$) to maintain a safety distance to other robots even if the nominal vector field $\bm{\vfcb^{[i]}}$ attempted to drive the robot to move closer to others. 
\begin{remark}
	The incorporation of the collision avoidance behavior may slow down the convergence speed of the path-following/surface-navigation behavior. Robots may not even move if deadlocks persistently appear \cite[Section VII]{wang2017safety}. In addition, to make sure that the coordinated path-following/surface-navigation behavior and the collision avoidance behavior are feasible, one should assume that the desired path/surface and the desired parametric distances have been appropriately designed such that when all robots follow the path or navigate the surface separated by the desired parametric distances, the Euclidean distance between each pair of them is greater than the safety distance $R$.
\end{remark}

\begin{figure*}[!t]
\small
\begin{multline} 	\label{eq_jsf3}
J(\bm{\vfsf^{[i]}}) =  \frac{\partial \bm{\vfsf^{[i]}}}{\partial \bm{\zeta^{[i]}}} = \\
\scalemath{0.81}{
	\left[
	\begin{array}{ c  : c : c : c  : c : c  }
	\begin{matrix}
	-\mk{i}{1}             &    0      & 0                 \\
	0         & -\mk{i}{2}            & 0                  \\
	0                 &    0                &  -\mk{i}{3}           \\
	\mk{i}{1} \partial_{w_1} \mf{i}{1}  &   \mk{i}{2} \partial_{w_1} \mf{i}{2}   & \mk{i}{3} \partial_{w_1} \mf{i}{3}  \\
	\mk{i}{1} \partial_{w_2} \mf{i}{1}  &   \mk{i}{2} \partial_{w_2} \mf{i}{2}   & \mk{i}{3} \partial_{w_2} \mf{i}{3} 
	\end{matrix}
	& 
	\begin{matrix}
	\bm{0}_{5 \times (i-1)}
	\end{matrix}
	&
	\begin{matrix}
	-v_{n+2} \partial_{w_1}^{w_1} \mf{i}{1} + v_{n+1} \partial_{w_2}^{w_1} \mf{i}{1} + k_1 \partial_{w_1} \mf{i}{1} \\
	-v_{n+2} \partial_{w_1}^{w_1} \mf{i}{2} + v_{n+1} \partial_{w_2}^{w_1} \mf{i}{2} + k_1 \partial_{w_1} \mf{i}{2} \\
	-v_{n+2} \partial_{w_1}^{w_1} \mf{i}{3} + v_{n+1} \partial_{w_2}^{w_1} \mf{i}{3} + k_1 \partial_{w_1} \mf{i}{3} \\
	\sum_{j=1}^{3}  [ \mk{i}{j}  \mphi{i}{j} \partial_{w_1}^{w_1} \mf{i}{j} -  \mk{i}{j} (\partial_{w_1} \mf{i}{j} )^2 ] \\
	\sum_{j=1}^{3}  [ \mk{i}{j}  \mphi{i}{j} \partial_{w_2}^{w_1} \mf{i}{j} -  \mk{i}{j} \partial_{w_1} \mf{i}{j} \, \partial_{w_2} \mf{i}{j} ]
	\end{matrix}
	&
	\begin{matrix}
	\bm{0}_{5 \times (N-1)}
	\end{matrix}
	&
	\begin{matrix}
	-v_{n+2} \partial_{w_1}^{w_2} \mf{i}{1} + v_{n+1} \partial_{w_2}^{w_2} \mf{i}{1} + k_1 \partial_{w_2} \mf{i}{1} \\
	-v_{n+2} \partial_{w_1}^{w_2} \mf{i}{2} + v_{n+1} \partial_{w_2}^{w_2} \mf{i}{2} + k_1 \partial_{w_2} \mf{i}{2} \\
	-v_{n+2} \partial_{w_1}^{w_2} \mf{i}{3} + v_{n+1} \partial_{w_2}^{w_2} \mf{i}{3} + k_1 \partial_{w_2} \mf{i}{3} \\
	\sum_{j=1}^{3}  [ \mk{i}{j}  \mphi{i}{j} \partial_{w_1}^{w_2} \mf{i}{j} -  \mk{i}{j} \partial_{w_1} \mf{i}{j} \, \partial_{w_2} \mf{i}{j} ] \\
	\sum_{j=1}^{3}  [ \mk{i}{j}  \mphi{i}{j} \partial_{w_2}^{w_2} \mf{i}{j} -  \mk{i}{j} (\partial_{w_2} \mf{i}{j} )^2 ]
	\end{matrix}
	&
	\begin{matrix}
	\bm{0}_{5 \times (N-i)}
	\end{matrix}
	\end{array}
	\right]	.
}
\end{multline}
\begin{align} \label{eq_jco3}
J( \bm{ \vfcrone^{[i]} }) = 
\left[ \arraycolsep=1.5pt \def\arraystretch{1.5}
\begin{array}{c : c}
\begin{matrix} 
\bm{0}_{5 \times 3}
\end{matrix}
&
\begin{matrix} 
\bm{0}_{3 \times 2N} \\ \hdashline[2pt/2pt]
\begin{matrix} a_{i1} & \cdots & a_{i(i-1)} & \; -\sum_{j=1}^{N} a_{ij} \; & a_{i(i+1)} & \cdots & a_{iN}  & \bm{0}_{1 \times N} \end{matrix} \\ \hdashline[2pt/2pt]
\bm{0}_{1 \times 2N} 
\end{matrix}	
\end{array}
\right]
=
\left[ \arraycolsep=1.5pt \def\arraystretch{1.5}
\begin{array}{c : c}
\begin{matrix} 
\bm{0}_{5 \times 3}
\end{matrix}
&
\begin{matrix}
\bm{0}_{3 \times 2 N} \\ \hdashline[2pt/2pt]
\begin{matrix} - \transpose{\bm{b_i}} L & \bm{0}_{1 \times N} \end{matrix} \\ \hdashline[2pt/2pt]
\bm{0}_{1 \times 2N} 
\end{matrix}	
\end{array}
\right].
\end{align}	
\begin{align} \label{eq_jco4}
J( \bm{ \vfcrtwo^{[i]} }) = 
\left[ \arraycolsep=1.5pt \def\arraystretch{1.5}
\begin{array}{c : c}
\begin{matrix} 
\bm{0}_{5 \times 3}
\end{matrix}
&
\begin{matrix} 
\bm{0}_{4 \times 2N} \\ \hdashline[2pt/2pt]
\begin{matrix} \bm{0}_{1 \times N} & a_{i1} & \cdots & a_{i(i-1)} & \; -\sum_{j=1}^{N} a_{ij} \; & a_{i(i+1)} & \cdots & a_{iN}  \end{matrix} 
\end{matrix}	
\end{array}
\right]
=
\left[ \arraycolsep=1.5pt \def\arraystretch{1.5}
\begin{array}{c : c}
\begin{matrix} 
\bm{0}_{5 \times 3}
\end{matrix}
&
\begin{matrix} 
\bm{0}_{4 \times 2 N} \\ \hdashline[2pt/2pt]
\begin{matrix}  \bm{0}_{1 \times N} & - \transpose{\bm{b_i}} L  \end{matrix}
\end{matrix}	
\end{array}
\right].
\end{align}	
\hrulefill
\vspace*{4pt}
\normalsize
\end{figure*}

\vspace{1em}
\section{Saturated controller for a Dubins-car-like model} \label{sec:controlalg}
If a robot's dynamics can be approximately modeled by the single-integrator model, then the coordinating guiding vector field in \eqref{eq_vf_combined} can be used directly as the velocity input to the robot. For the unicycle model, one can use feedback linearization to transform it into the single-integrator model \cite{yun1992} to utilize the guiding vector field directly. However, we will design a controller for a Dubins-car-like model without using the feedback linearization technique. Note that the control algorithm design idea in this section is applicable to robot models whose motions are characterized by the robot's orientations, such as the car-like model and the underwater glider model \cite{Siciliano2007}. These models (approximately) represent many different robotic systems in reality, and thus the design methodology is widely applicable.

Different from the unicycle model, which allows backwards or stationary motion, we use the following Dubins-car-like 3D model that describes the fixed-wing aircraft dynamics:
\begin{align} \label{model}
\dot{p}^{[i]}_1 = v \cos \theta^{[i]}, \;
\dot{p}^{[i]}_2 = v \sin \theta^{[i]} , \;
\dot{p}^{[i]}_3 = u^{[i]}_{z} ,  \;
\dot{\theta}^{[i]} = u^{[i]}_{\theta},
\end{align}
where $v$ is a constant speed, $\transpose{(p^{[i]}_1,p^{[i]}_2,p^{[i]}_3)} \in \mbr[3]$ is the position of the $i$-th aircraft's center of mass, $\theta^{[i]}$ is the yaw angle, and $u^{[i]}_{z}$ and $u^{[i]}_{\theta}$ are two control inputs to be designed. Since the essential role of a guiding vector field is to provide the desired yaw angle to guide the flight of a fixed-wing aircraft, the core idea behind the control algorithm design is to align the aircraft's flying direction with that given by the guiding vector field. In the sequel, we consider the case where there are two parameters and design a control algorithm for fixed-wing aircraft to navigate on a surface\footnote{For the one-parameter case, see \cite[Theorem 2]{yao2021multi}.}. Therefore, the coordinating guiding vector field has two additional coordinates, and thus one needs to add two additional virtual coordinates $p^{[i]}_4$ and $p^{[i]}_5$ such that the aircraft's generalized position is $(p^{[i]}_1,p^{[i]}_2,p^{[i]}_3,p^{[i]}_4,p^{[i]}_5) \in \mbr[5]$. Correspondingly, its generalized velocity is $(\dot{p}^{[i]}_1, \dot{p}^{[i]}_2 ,\dot{p}^{[i]}_3, \dot{p}^{[i]}_4, \dot{p}^{[i]}_5) = (v \cos\theta^{[i]}, v\sin\theta^{[i]}, u^{[i]}_{z}, \dot{p}^{[i]}_4, \dot{p}^{[i]}_5)$, where $\dot{p}^{[i]}_4, \dot{p}^{[i]}_5$ are virtual, acting as extra design freedom. To align the aircraft heading $v (\cos\theta^{[i]}, \sin\theta^{[i]})$ with the counterpart of the coordinating guiding vector field $\bm{\vfcb^{[i]}}$, we need to ``partially normalize'' the vector field $\bm{\vfcb^{[i]}}$ such that its first two entries form a vector of the same length as $v$; that is, 
\[
v \bm{\underline{\vfcb^{[i]}}} \defeq v \bm{\vfcb^{[i]}} / \sqrt{\mvfcb{i}{1}^2 + \mvfcb{i}{2}^2}.
\]
Subsequently, we need to design the yaw angular control input $u^{[i]}_{\theta}$ such that the aircraft heading $v (\cos\theta^{[i]}, \sin\theta^{[i]})$ gradually aligns with the vector formed by the first two entries of $v \bm{\underline{\vfcb^{[i]}}}$ [i.e., $v ({\underline{\vfcb^{[i]}}}{}_{1}, {\underline{\vfcb^{[i]}}}{}_{2})$]. For the last three entries of the generalized velocity $(\dot{p}^{[i]}_1, \dot{p}^{[i]}_2 ,\dot{p}^{[i]}_3, \dot{p}^{[i]}_4, \dot{p}^{[i]}_5)$, one can simply equate them with those of the ``partially normalized'' vector field $v \bm{\underline{\vfcb^{[i]}}}$  respectively; i.e.,
\begin{equation}\label{eq_u34}
\begin{split}
&\dot{p}^{[i]}_3 = u^{[i]}_{z}= v \mvfcb{i}{3}/ \sqrt{\mvfcb{i}{1}^2 + \mvfcb{i}{2}^2}\\
&\dot{p}^{[i]}_4 = v \mvfcb{i}{4} / \sqrt{\mvfcb{i}{1}^2 + \mvfcb{i}{2}^2} \\
&\dot{p}^{[i]}_5 = v \mvfcb{i}{5} / \sqrt{\mvfcb{i}{1}^2 + \mvfcb{i}{2}^2}.
\end{split}
\end{equation}
This control algorithm design method \cite{yao2021singularity,rezende2018robust,yaothesis2021} is extended here to handle the issue with the actuator saturation in the yaw angular control input $u^{[i]}_{\theta}$ as described below. First, we define the saturation function $\sat: \mbr[] \to \mbr[]$ by $\sat(x)=x$ for $x \in [a,b]$, $\sat(x)=a$ for $x \in (-\infty, a)$ and $\sat(x)=b$ for $x \in (b, \infty)$, where $a,b \in \mbr[], a<b$ are some constants. Although the saturation function $\sat$ is not differentiable, it is Lipschitz continuous. For convenience, we call the time interval when $\sat(x(t))=b$ the \emph{upper saturation period}, and the time interval when $\sat(x(t))=a$ the \emph{lower saturation period}. We use the notation $\normv{\bm{v}}$ to denote the normalization of a vector $\bm{v}$ (i.e., $\normv{\bm{v}}=\bm{v} / \norm{\bm{v}}$). We also define %
\begin{equation} \label{eq_vfcbp}
	\bm{\vfcb^{[i]}_p} = \transpose{(\mhatvfcb{i}{1}, \mhatvfcb{i}{2})},
\end{equation}
which is the vector formed by the first two entries of the normalized vector field $\normv{\bm{\vfcb^{[i]}}}$. Therefore, $\normv{\bm{\vfcb^{[i]}_p}}$ represents the orientation given by the vector field $\bm{\vfcb^{[i]}}$. In addition, one can easily calculate that $\normv{\bm{\vfcb^{[i]}_p}} = ({\underline{\vfcb^{[i]}}}{}_{1}, {\underline{\vfcb^{[i]}}}{}_{2})$. 

Suppose we are given physical desired surfaces $\phy{\set{S}}^{[i]} \subseteq \mbr[3]$ parameterized by \eqref{eq_param_func_surf}, where $n=3$, and the coordinating guiding vector field $\bm{\vfcb^{[i]}}: \mbr[3+2N] \to \mbr[3+2]$ in \eqref{eq_vf_combined_surf}. Denote the orientation of the aircraft by $\bm{h^{[i]}} = \normv{\bm{h^{[i]}}}=\transpose{(\cos \theta^{[i]}, \sin \theta^{[i]})}$,  the (signed) angle difference directed from $\normv{\bm{\vfcb^{[i]}_p}}$ to $\normv{\bm{h^{[i]}}}$ by $\sigma^{[i]} \in (-\pi, \pi]$ and define the rotation matrix $E=\left[\begin{smallmatrix}0 & -1 \\ 1 & 0\end{smallmatrix}\right]$. 
The following theorem states that the aircraft's orientation $\normv{\bm{h^{[i]}}}$ will converge to that of the vector field $\normv{\bm{\vfcb^{[i]}_p}}$ asymptotically (i.e., $\sigma^{[i]}$ converges to zero).

\begin{figure}[tb]
	\centering
	\begin{tikzpicture}[scale=0.8]
	\node[right] (vfcb) at (1.5,0) {$\normv{\bm{\vfcb^{[i]}_p}}$};
	\node[right] (hnode) at (1.2,1.2) {$\normv{\bm{h^{[i]}}}$} ;
	\draw [->, >=stealth, out=80, in=-30, looseness=1] (vfcb.east) to  node[right]{$\dot{\theta}^{[i]}_d>0$} (hnode.east);
	\draw [->, >=stealth] (0, 0) -- (1.5,0);
	\draw [->, >=stealth] (1.5,0) coordinate(A) -- (0,0) coordinate(B) -- (1.2,1.2)  coordinate(C) pic[draw,->, "$\sigma^{[i]}$", angle eccentricity=1.8]{angle};
	\end{tikzpicture}
	\caption{The signed angle $\sigma^{[i]}$.}
	\label{fig: beta}
\end{figure}
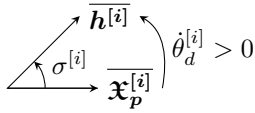

\begin{theorem}	\label{thm_guidance}
	Assume that the vector field satisfies $\mvfcb{i}{1}(\bm{\xi^{[i]}})^2 + \mvfcb{i}{2}(\bm{\xi^{[i]}})^2 > \gamma>0$ for $\bm{\xi^{[i]}} \in \mbr[3+2]$, $i\in \mathbb{Z}_{1}^{N}$, where $\gamma$ is a positive constant.  Let the angular velocity control input $u^{[i]}_{\theta}$  in model \eqref{model} be
	\begin{align} 
	\dot{\theta}^{[i]}=u^{[i]}_{\theta} = \sat( \dot{\theta}^{[i]}_d - k_\theta \transpose{\normv{\bm{h^{[i]}}}} E \normv{\bm{\vfcb^{[i]}_p}} ), \label{eq_u_theta} 
	\end{align}
	where 
	\begin{equation} \label{eq_dotthetad}
	\dot{\theta}^{[i]}_d = -\transpose{\normv{\bm{\vfcb^{[i]}_p}}} E \bm{\dot{\vfcb}^{[i]}_p} / \norm{\bm{\vfcb^{[i]}_p}},
	\end{equation}
	$k_\theta>0$ is a constant, and $a<0$, $b>0$ are constants for the saturation function $\sat$. If the angle difference $\sigma^{[i]}$ satisfies the following conditions:
	\begin{enumerate}[leftmargin=*]
		\item The initial angle difference $\sigma^{[i]}(t=0) \ne \pi$;
		\item \label{cond2} $\sigma^{[i]}(t) \in [0, \pi)$ during the upper saturation period,  and $\sigma^{[i]}(t) \in (-\pi, 0]$ during the lower saturation period,
	\end{enumerate}
	then $\sigma^{[i]}$ will vanish asymptotically (i.e., $\sigma^{[i]}(t) \to 0$). 
\end{theorem}
\begin{proof}
		See Appendix \ref{app:thm_guidance}.
\end{proof}
\begin{remark} 
The quantity $\bm{\dot{\vfcb}^{[i]}_p}$ in \eqref{eq_dotthetad} can be calculated by 
\begin{equation} \label{eq_dotvfcbp}
	\bm{\dot{\vfcb}^{[i]}_p} = J(\bm{\vfcb^{[i]}_p}) \bm{\dot{\zeta}^{[i]}},
\end{equation}
where $J(\bm{\vfcb^{[i]}_p}) \in \mbr[2 \times (3+2N)]$ is the Jacobian matrix of $\bm{\vfcb^{[i]}_p}$ with respect to the generalized position $\bm{\zeta^{[i]}}=\transpose{(\transpose{\bm{\xi^{[i-w]}}}, \transpose{\bm{\mw{\cdot}{1}}}, \transpose{\bm{\mw{\cdot}{2}}} )} = \transpose{(\mx{i}{1}, \mx{i}{2}, \mx{i}{3}, \transpose{\bm{\mw{\cdot}{1}}}, \transpose{\bm{\mw{\cdot}{2}}} )}\in \mbr[3+2N]$, with $\bm{\xi^{[i-w]}}$ representing the vector obtained by deleting the last two entries (i.e., the virtual coordinates $w_{1}^{[i]}$ and $w_{2}^{[i]}$) of $\bm{\xi^{[i]}} \in \mbr[3+2]$.  In addition, we can simplify the computation of the Jacobian $\bm{\vfcb^{[i]}_p}$ to 
\begin{multline*}
	J(\bm{\vfcb^{[i]}_p})=F J(\normv{\bm{\vfcb^{[i]}}})= F(I-\normv{\bm{\vfcb^{[i]}}} \; \transpose{\normv{\bm{\vfcb^{[i]}}}}) J(\bm{\vfcb^{[i]}}) / \norm{\bm{\vfcb^{[i]}}} \\
	= F(I-\normv{\bm{\vfcb^{[i]}}} \; \transpose{\normv{\bm{\vfcb^{[i]}}}}) J( \bm{\vfsf^{[i]}} + k_{c1} \bm{\vfcrone^{[i]}} +  k_{c2} \bm{\vfcrtwo^{[i]}}  ) / \norm{\bm{\vfcb^{[i]}}} \\
	= \scalemath{0.8}{ F(I-\normv{\bm{\vfcb^{[i]}}} \; \transpose{\normv{\bm{\vfcb^{[i]}}}}) \big(  J(\bm{\vfsf^{[i]}}) + k_{c1} J(\bm{\vfcrone^{[i]}}) +  k_{c2} J(\bm{\vfcrtwo^{[i]}})  \big)  / \norm{\bm{\vfcb^{[i]}}} } ,
\end{multline*}
where $F =\left[ \begin{smallmatrix} 1 & 0 & 0 & 0 & 0 \\ 0 & 1 & 0 & 0 & 0 \end{smallmatrix} \right]$ and the Jacobians $J(\bm{\vfsf^{[i]}})$, $ J(\bm{\vfcrone^{[i]}})$ and  $ J(\bm{\vfcrtwo^{[i]}})$ are shown in \eqref{eq_jsf3}, \eqref{eq_jco3} and \eqref{eq_jco4}, respectively (where $\bm{b_i}$ is the basis vector with the $i$-th entry being $1$). Although \eqref{eq_dotvfcbp} contains states $\dot{w}_{1}^{[j]}$ and $\dot{w}_{2}^{[j]}$ from all robots, we emphasize that \emph{each robot only needs the information $w_{1}^{[j]}$, $w_{2}^{[j]}$, $\dot{w}_{1}^{[j]}$ and $\dot{w}_{2}^{[j]}$ from its neighbors (i.e., $j \in \mathcal{N}_{i}$), and thus the control algorithm is distributed}. This is because in \eqref{eq_jco3} and \eqref{eq_jco4}, those terms $a_{ij}$ become zero if $j$ is \emph{not} a neighbor of the $i$-th robot, and thereby the corresponding information $w_{1}^{[j]}$, $w_{2}^{[j]}$, $\dot{w}_{1}^{[j]}$ and $\dot{w}_{2}^{[j]}$ is \emph{not} required by the $i$-th robot. In particular, if the communication graph is a cycle, then each robot only needs the information of $w_{1}^{[j]}$, $w_{2}^{[j]}$, $\dot{w}_{1}^{[j]}$ and $\dot{w}_{2}^{[j]}$ from its \emph{two} neighbors, regardless of the size of the multi-robot system.
\end{remark}
\begin{remark}[{Interpretation of Condition \ref{cond2}}]
	In the proof of the theorem, it has been shown that 
	\[
		\dot{\normv{\bm{\vfcb^{[i]}_p}}} = \dot{\theta}^{[i]}_d E \normv{\bm{\vfcb^{[i]}_p}}.
	\]
	Since $E \normv{\bm{\vfcb^{[i]}_p}}$ is orthogonal to $\normv{\bm{\vfcb^{[i]}_p}}$, the quantity $\dot{\theta}^{[i]}_d$ in \eqref{eq_u_theta} encodes how fast the vector field changes its orientation along the trajectory of the aircraft (i.e., $\dot{\theta}^{[i]}_d$ is the \emph{change rate of the vector field orientation}). Since $\transpose{\normv{\bm{h^{[i]}}}} E \normv{\bm{\vfcb^{[i]}}}\le 1$ in \eqref{eq_u_theta}, saturation may happen due to the possibly large magnitude of the term $\dot{\theta}^{[i]}_d$. If upper saturation happens at $t=t_0$, the change rate of the vector field orientation $\dot{\theta}^{[i]}_d$ demands a faster change than the aircraft can achieve; therefore, if $\sigma^{[i]}(t_0)>0$, then the vector field orientation encoded by $\normv{\bm{\vfcb^{[i]}_p}}$ will ``chase'' the aircraft orientation vector $\normv{\bm{h^{[i]}}}$ such that the angle difference $\sigma^{[i]}(t)$ is decreasing (see Fig. \ref{fig: beta}). Thus as long as $\sigma^{[i]}(t_0)>0$, the Lyapunov function $V$ is still decreasing. However, if the saturation lasts for a long time such that $\normv{\bm{\vfcb^{[i]}_p}}$ overtakes the aircraft orientation vector $\normv{\bm{h^{[i]}}}$, the angle $\sigma^{[i]}(t)$ will become negative and violate Condition \ref{cond2} in Theorem \ref{thm_guidance}; thus the decreasing property of the Lyapunov function $V$ is not necessarily guaranteed; in this case, $\dot{V}> - k_\theta \left( \transpose{\normv{\bm{h^{[i]}}}} E \normv{\bm{\vfcb^{[i]}_p}} \right)^2 $, so $\dot{V}$ may be negative or positive. Therefore, although Condition \ref{cond2} in Theorem \ref{thm_guidance} might be \emph{difficult} to check in practice, it conveys the core message that the saturation, albeit allowed, should not last for a long time. However, this condition is only sufficient (not necessary), while in practice, violating this condition does not immediately entail instability of the algorithm. As the aircraft is guided by the vector field, the aircraft can re-orient its heading towards the desired path even if it temporarily deviates due to saturation or other constraints (such as path curvature), as long as the change rate of the vector field orientation $\dot{\theta}^{[i]}_d$ does not saturate the control input persistently. The subsequent fixed-wing aircraft experiment verifies the effectiveness of the control law.
\end{remark}
Nevertheless, we can remove Condition \ref{cond2} in Theorem \ref{thm_guidance} by imposing an upper bound on the magnitude of $\dot{\theta}^{[i]}_d$, as shown in Corollary \ref{coroll1}.
\begin{coroll} \label{coroll1}
	Suppose there exists a positive constant $d$ satisfying $d<\min\{-a, b\}$, where $a, b$ are the threshold values of the saturation function $\sat(\cdot)$ in Theorem \ref{thm_guidance} (note that $a<0$), such that the change rate of the vector field orientation $|\dot{\theta}^{[i]}_d| \le d$. Let $\bar{k}_\theta \defeq  \min\{-a-d, b-d\}>0$. If the positive gain $k_\theta$ in \eqref{eq_u_theta} is chosen within the range $(0, \bar{k}_\theta)$, then the angle difference $\sigma^{[i]}$ converges to $0$ without requiring Condition \ref{cond2} in Theorem \ref{thm_guidance}.
\end{coroll}
\begin{proof}
	See Appendix \ref{app:coroll1}.
\end{proof}
To reduce the magnitude of $\dot{\theta}^{[i]}_d$ and avoid possible saturation, one can scale down the path parameter in the parametric functions in \eqref{eq_param_func} (e.g. by changing $\mf{i}{j}(w^{[i]})$ to $\mf{i}{j}(\beta w^{[i]})$, where $0<\beta<1$), or choose another desired path with a smaller curvature. Another approach to avoid input saturation is to add an additional constraint $|u^{[i]}_{\theta}| \le \min\{|a|, |b|\}$ in the quadratic program in Section \ref{sec:collision}.

\section{Simulations and experiments} \label{sec:simexp}

\subsection{Simulations}
\begin{figure}[tb]
	\centering
	\subfigure[]{
		\includegraphics[width=0.5\linewidth]{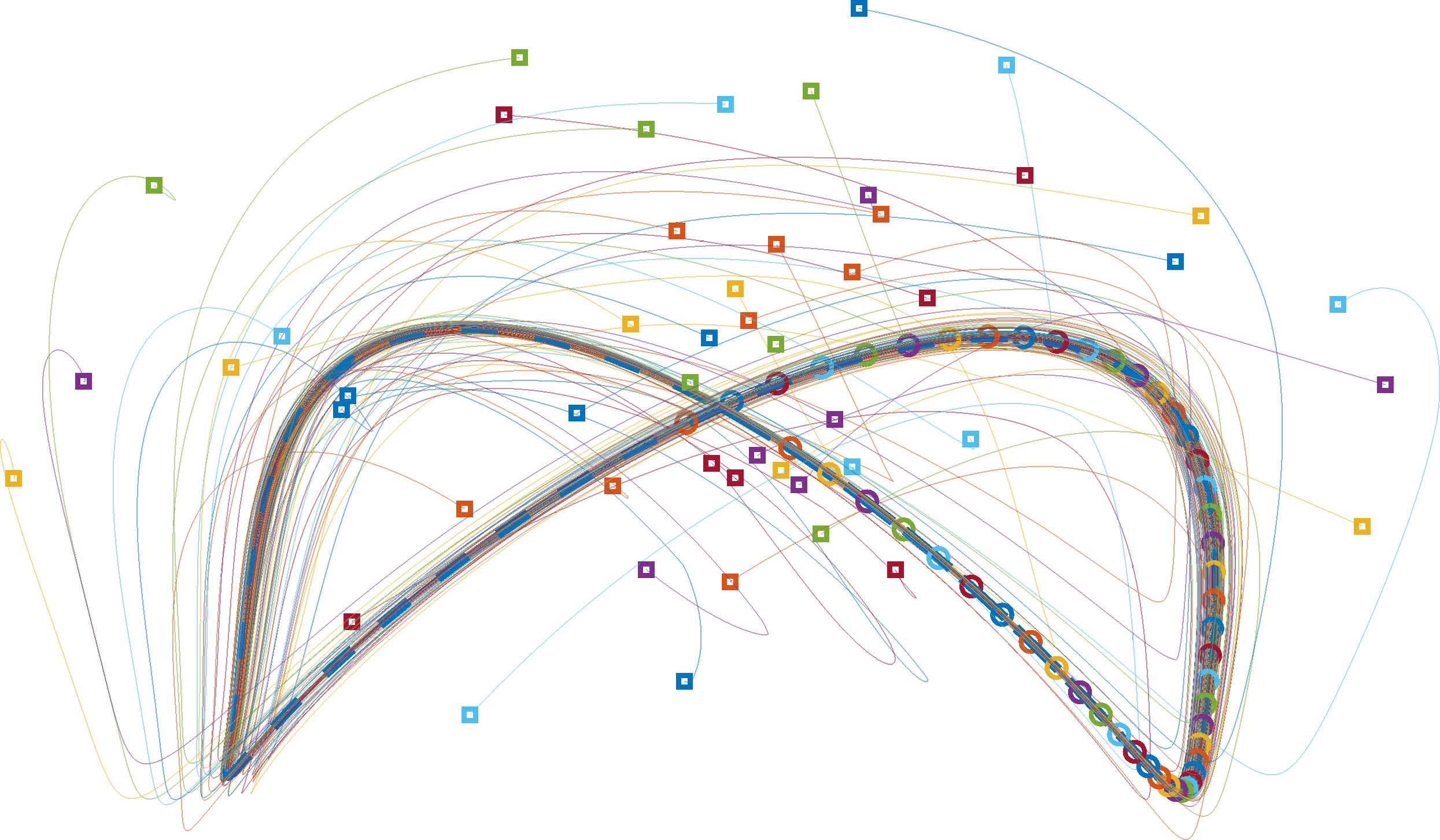}
		\label{fig:h2_201020_1353_n3N50_trjnew}
	} \hspace{-0.8em}
	\subfigure[]{
		\includegraphics[width=0.44\columnwidth]{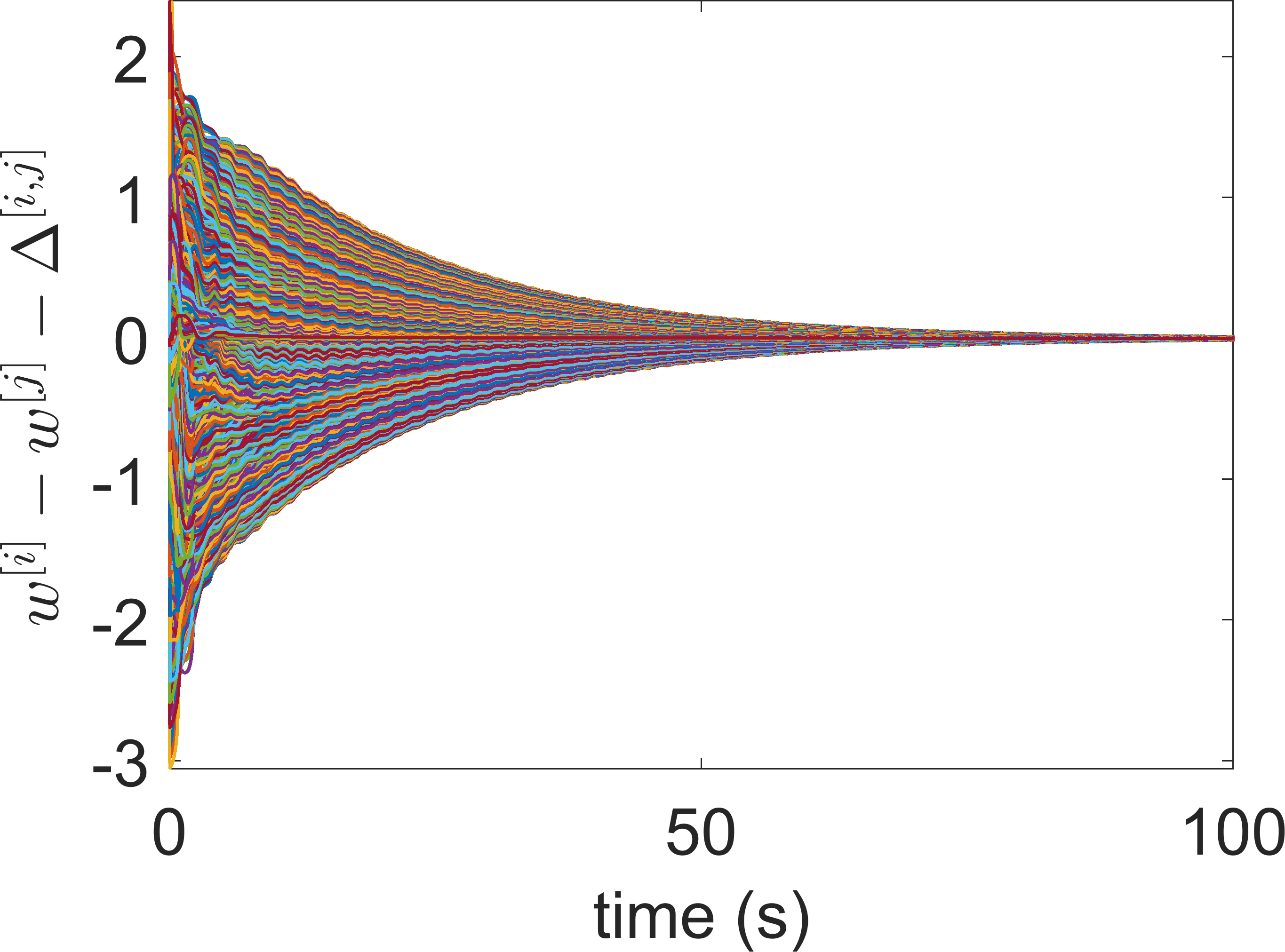}
		\label{fig:h2_201020_1353_n3N50_wnew}
	} \\
	\subfigure[]{
		\includegraphics[width=0.44\columnwidth]{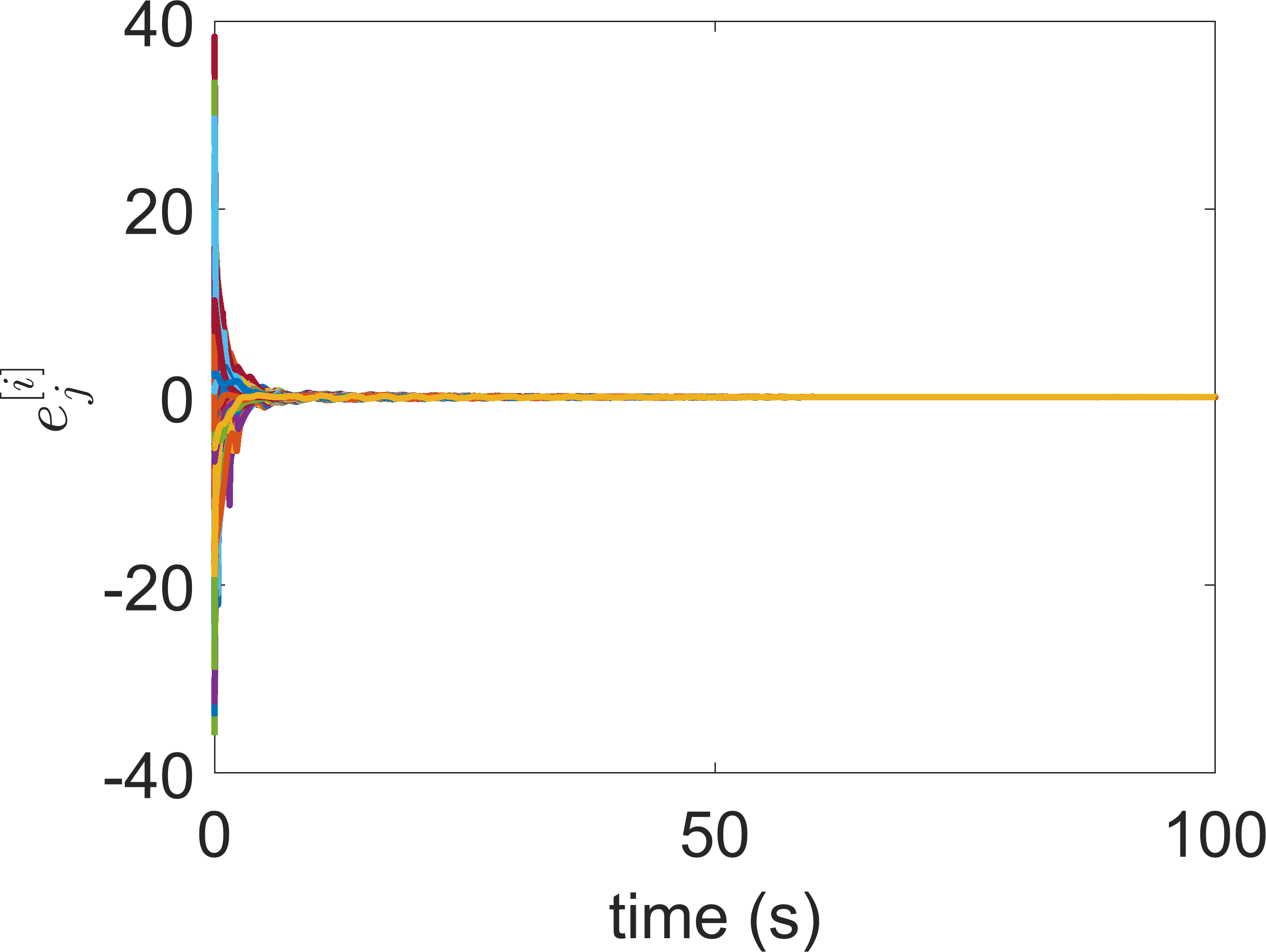}
		\label{fig:h2_201020_1353_n3N50_errornew}
	}	
	\subfigure[]{
		\includegraphics[width=0.46\columnwidth]{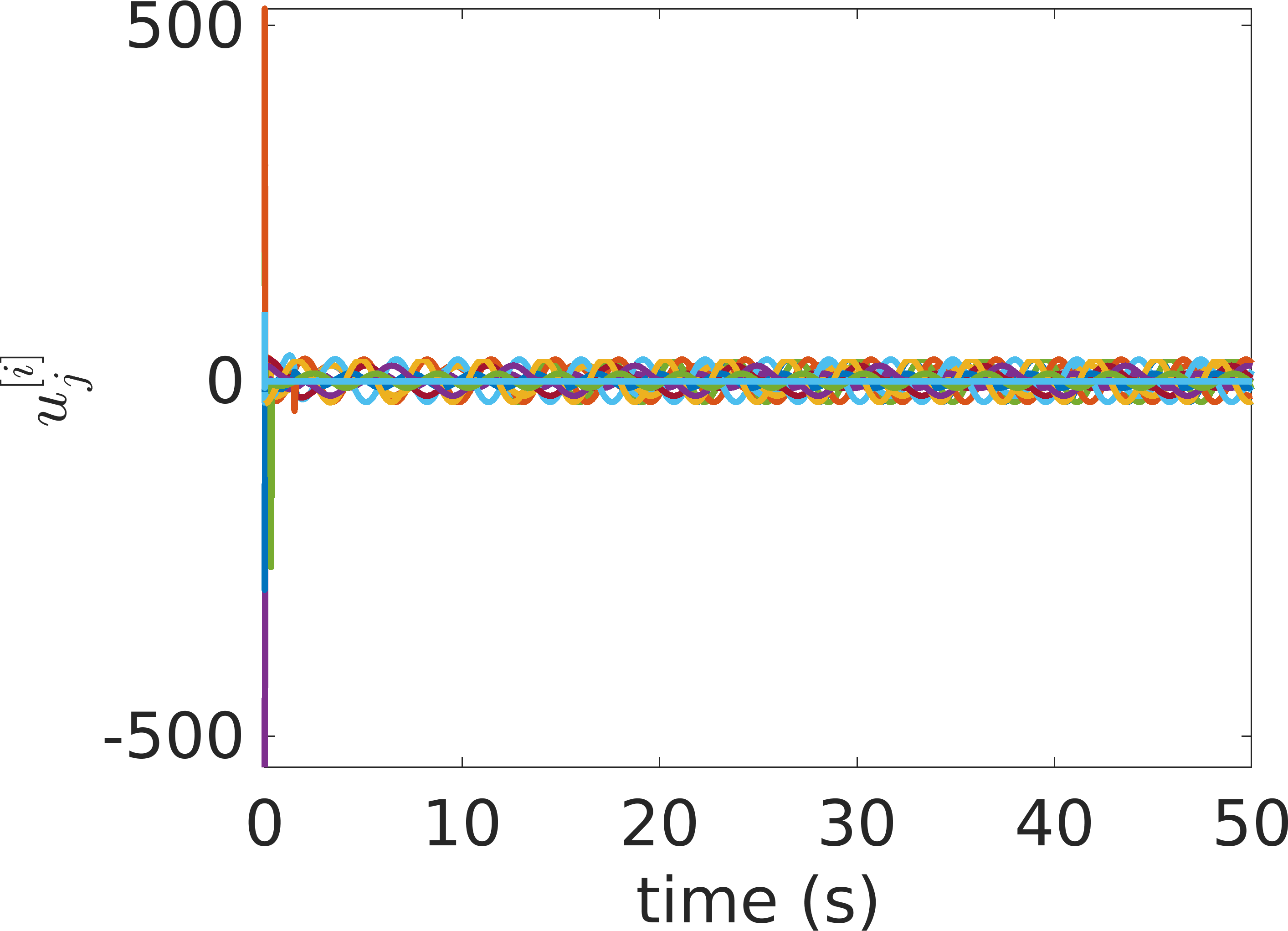}
		\label{fig:h2_201020_1353_n3N50_control}
	}
	\caption{The first simulation results. \subref{fig:h2_201020_1353_n3N50_trjnew} The trajectories of robots, where squares and circles symbolize the trajectories' initial and final positions respectively.  \subref{fig:h2_201020_1353_n3N50_wnew} The coordination errors $w^{[i]}-w^{[j]}-\Delta^{[i,j]}$ for $i,j\in \mathbb{Z}_{1}^{50},i<j$. \subref{fig:h2_201020_1353_n3N50_errornew} The path-following errors $\phi^{[i]}_j$ for $i\in \mathbb{Z}_{1}^{50}$ and $j\in \mathbb{Z}_{1}^{3}$. \subref{fig:h2_201020_1353_n3N50_control} The control inputs $u^{[i]}_j \defeq \vfcb^{[i]}_j$ for $i=1,13,25,37,50$, $j\in \mathbb{Z}_{1}^{3+1}$. Only 50 seconds of data are shown for clarity in \subref{fig:h2_201020_1353_n3N50_errornew}-\subref{fig:h2_201020_1353_n3N50_control}. }
	\label{fig:sim1}
\end{figure}
\begin{figure}[tb]
	\centering
	\subfigure[]{
		\includegraphics[width=0.29\columnwidth]{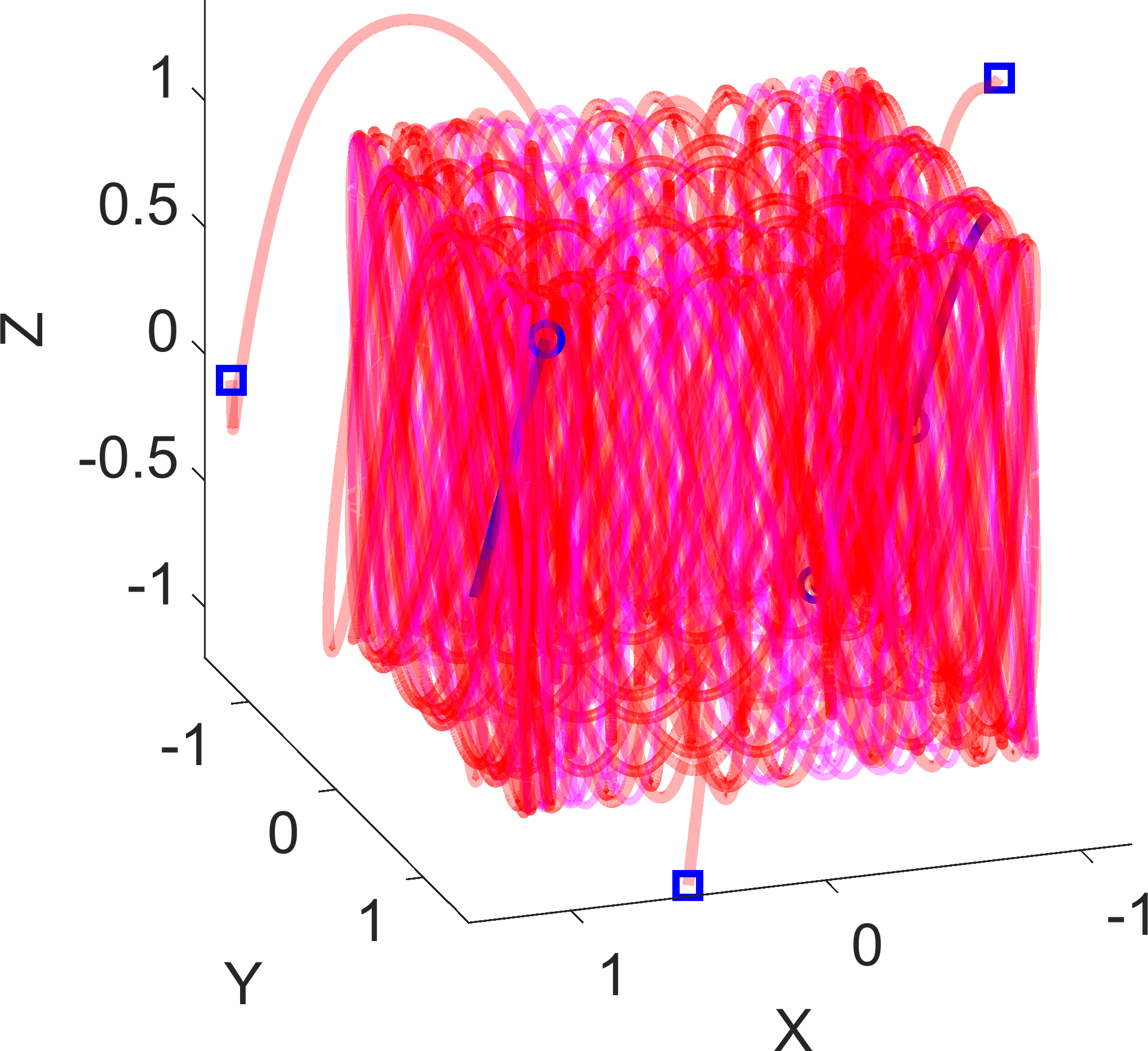}
		\label{fig:h3_201021_1003_n3N3_trj}
	}  \hspace{-1.2em}
	\subfigure[]{
		\includegraphics[width=0.32\columnwidth]{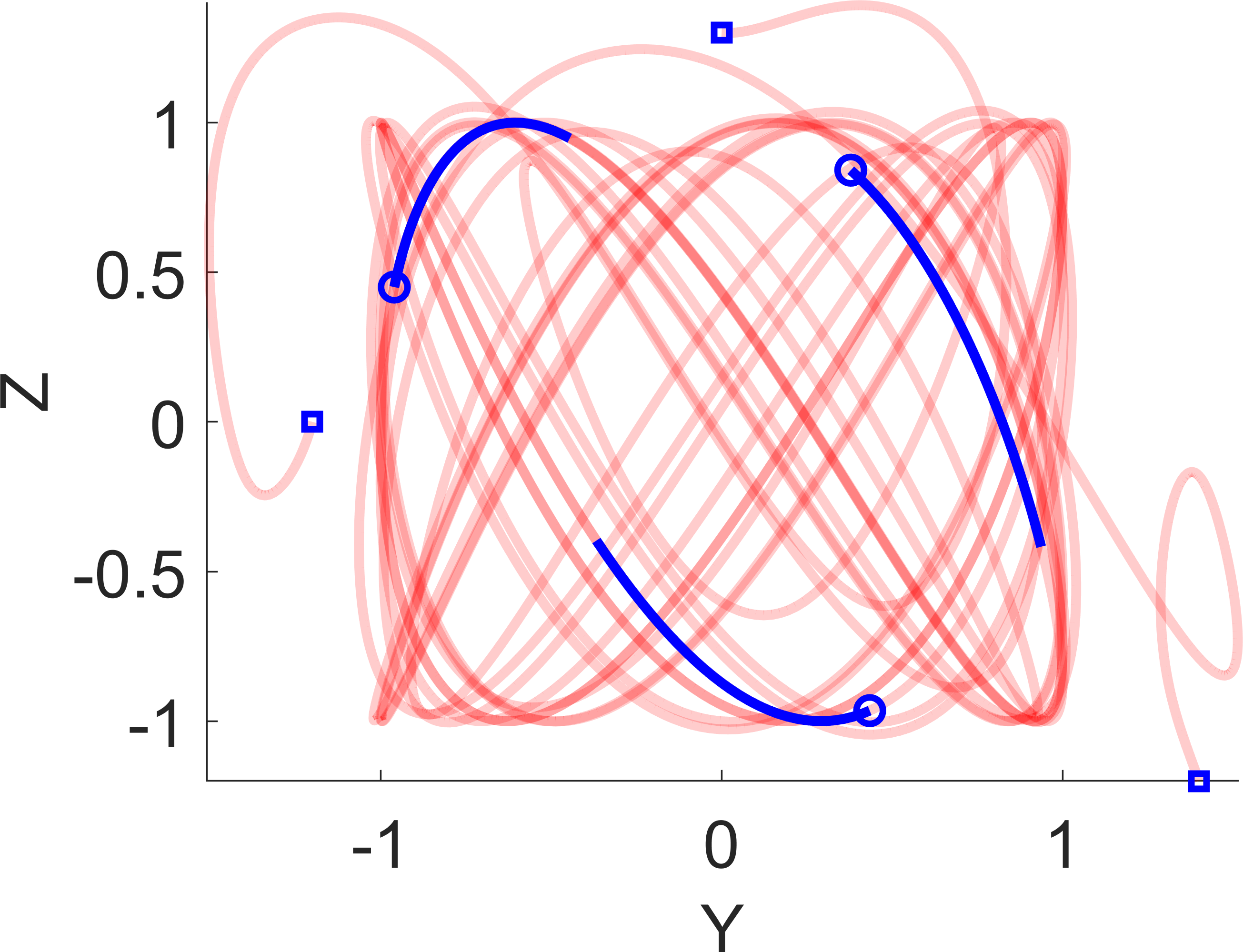}
		\label{fig:h3_201021_1003_n3N3_yz9-4}
	} \hspace{-1.2em}
	\subfigure[]{
		\includegraphics[width=0.32\columnwidth]{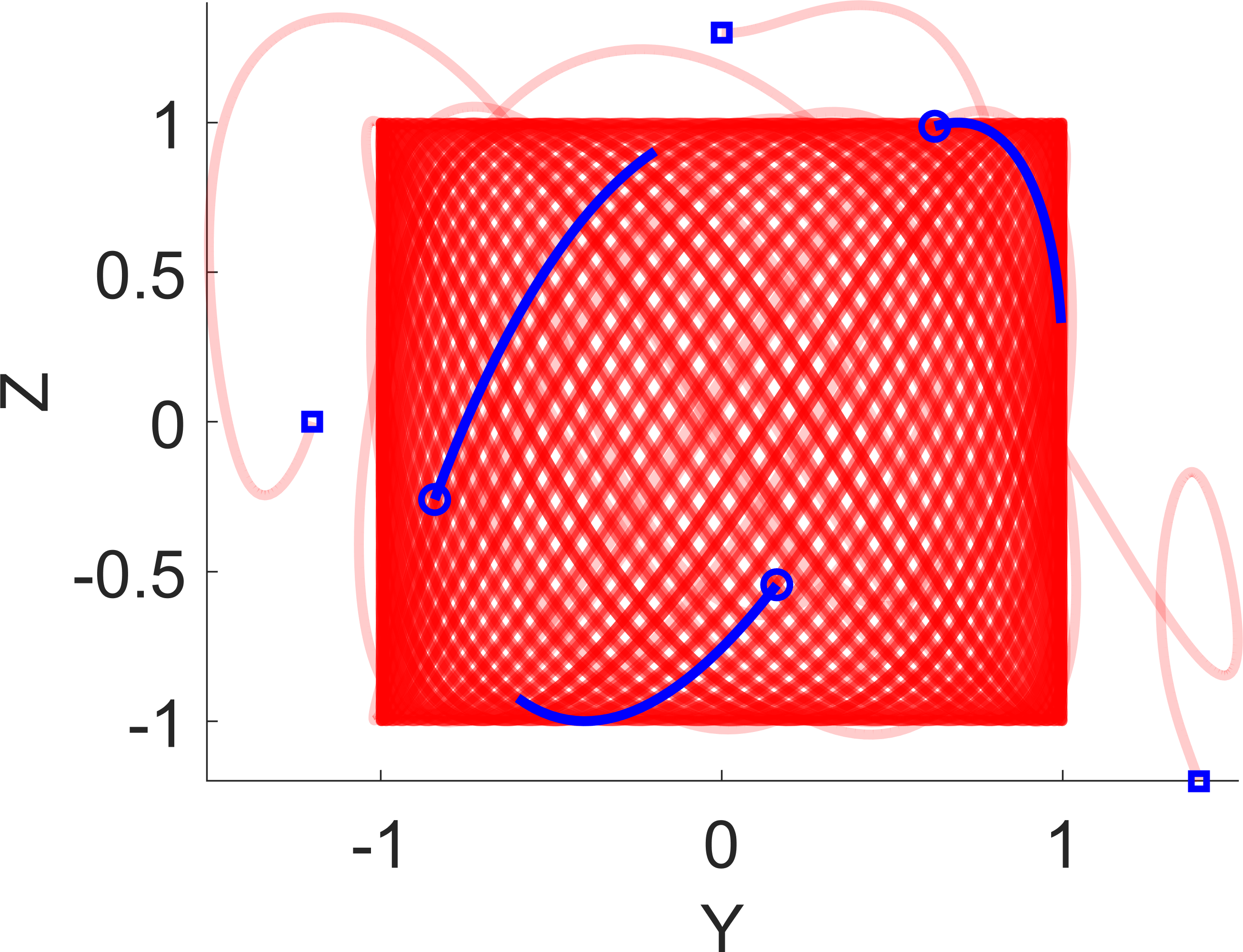}
		\label{fig:h3_201021_1003_n3N3_yz76-8}
	} \\
	\subfigure[]{
	\includegraphics[width=0.32\columnwidth]{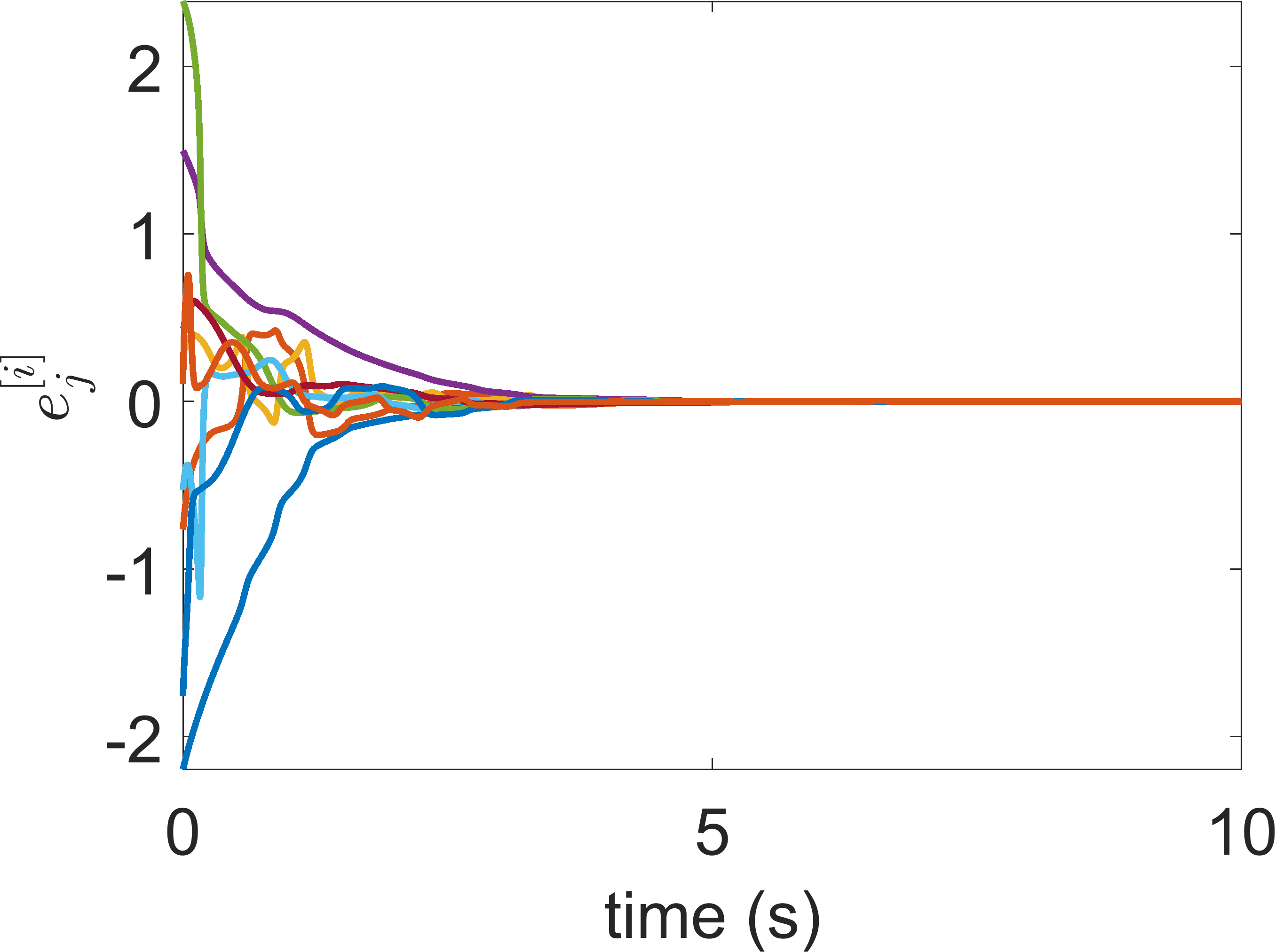}
	\label{fig:h3_201021_1003_n3N3_error}
	} \hspace{-1.2em}
	\subfigure[]{
		\includegraphics[width=0.32\columnwidth]{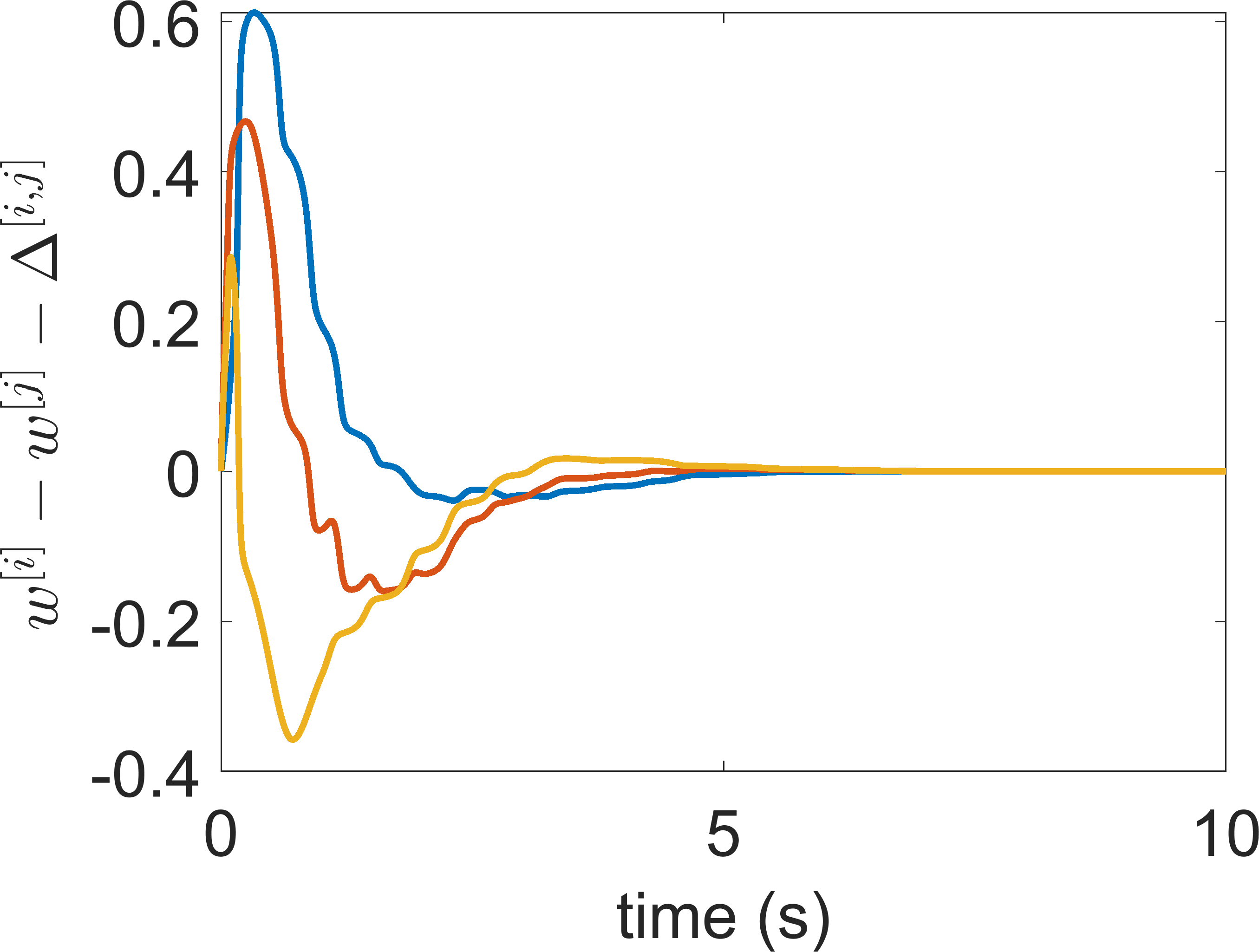}
		\label{fig:h3_201021_1003_n3N3_w}
	} \hspace{-1.2em}
	\subfigure[]{
		\includegraphics[width=0.32\columnwidth]{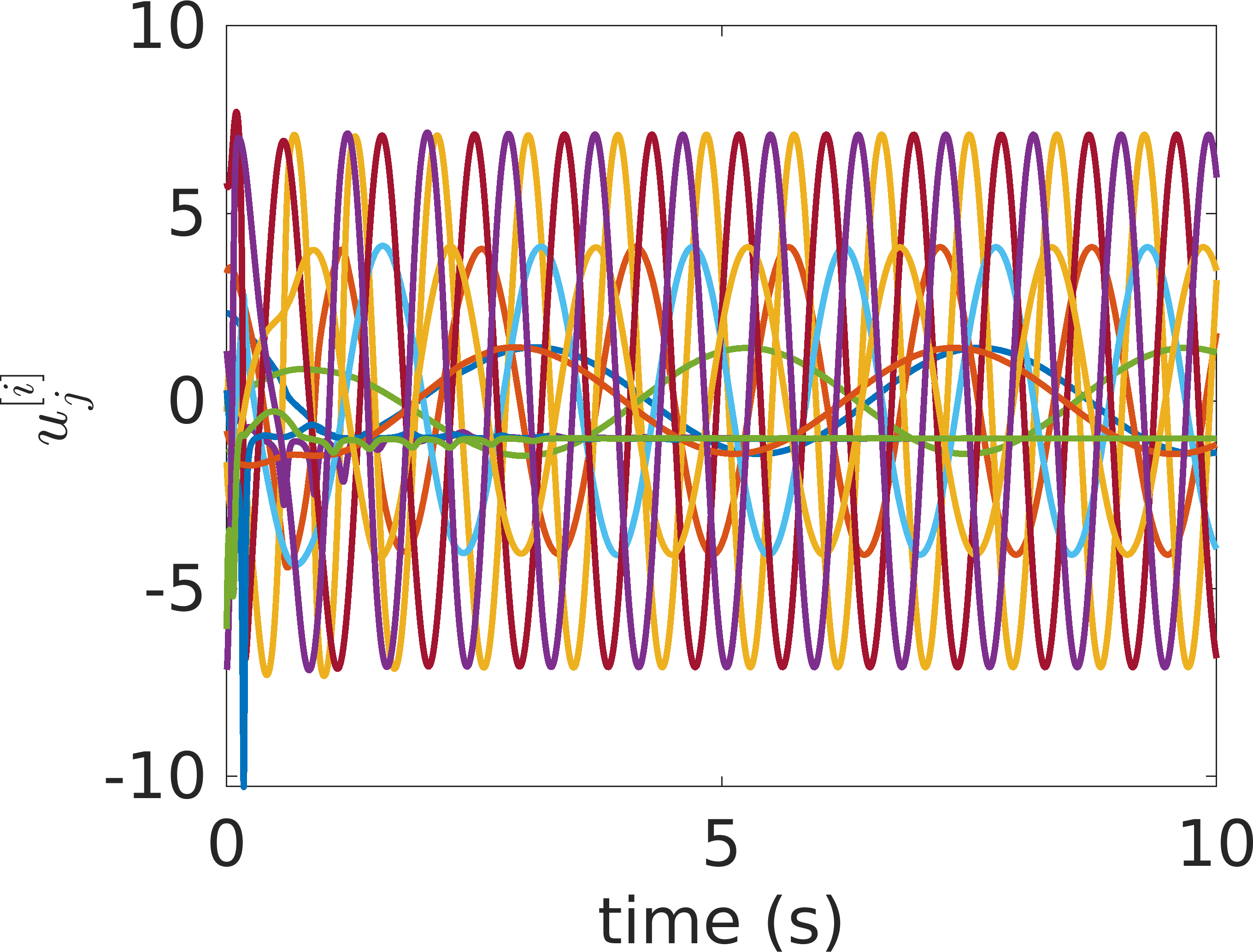}
		\label{fig:h3_201021_1003_n3N3_control}
	}
	\caption{The second simulation results. Squares and circles symbolize trajectories' initial and final positions, and the solid blue lines are the trajectories during the last $30$ time steps. \subref{fig:h3_201021_1003_n3N3_trj} The trajectories of three robots follow an open Lissajous curve with irrational coefficients. The magenta curve represents part of the Lissajous curve. \subref{fig:h3_201021_1003_n3N3_yz9-4} and \subref{fig:h3_201021_1003_n3N3_yz76-8} correspond to the Y-Z side views of the trajectories at time $9.4$ and $76.8$ seconds, respectively. \subref{fig:h3_201021_1003_n3N3_error} The path-following errors $\phi^{[i]}_j$ for $i\in \mathbb{Z}_{1}^{3}$ and $j\in \mathbb{Z}_{1}^{3}$. \subref{fig:h3_201021_1003_n3N3_w} The coordination error $w^{[i]}-w^{[j]}-\Delta^{[i,j]}$ for $i,j\in \mathbb{Z}_{1}^{3},i<j$. \subref{fig:h3_201021_1003_n3N3_control} The control inputs $u^{[i]}_j \defeq \vfcb^{[i]}_j$ for $i \in \mathbb{Z}_{1}^{3}$, $j\in \mathbb{Z}_{1}^{3+1}$. Only 10 seconds of data are shown for clarity in \subref{fig:h3_201021_1003_n3N3_error}-\subref{fig:h3_201021_1003_n3N3_control}.}
	\label{fig:sim2_1}
\end{figure}

\begin{figure}[tb]
	\centering
	\subfigure[]{
		\includegraphics[width=0.48\linewidth]{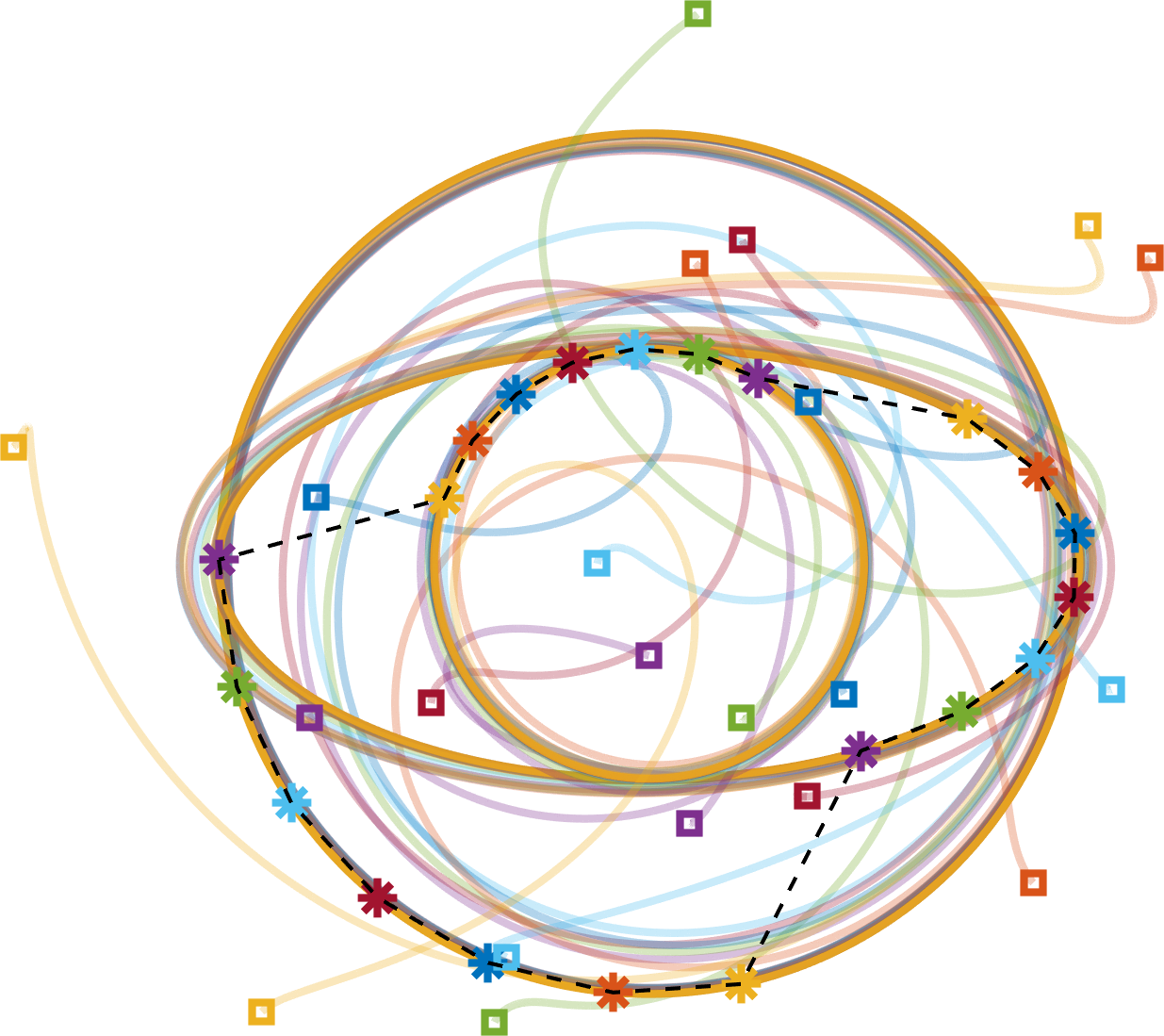}
		\label{fig:h4_201020_1843_n2N21_trj}
	} \hspace{-1.2em}
	\subfigure[]{
		\includegraphics[width=0.48\columnwidth]{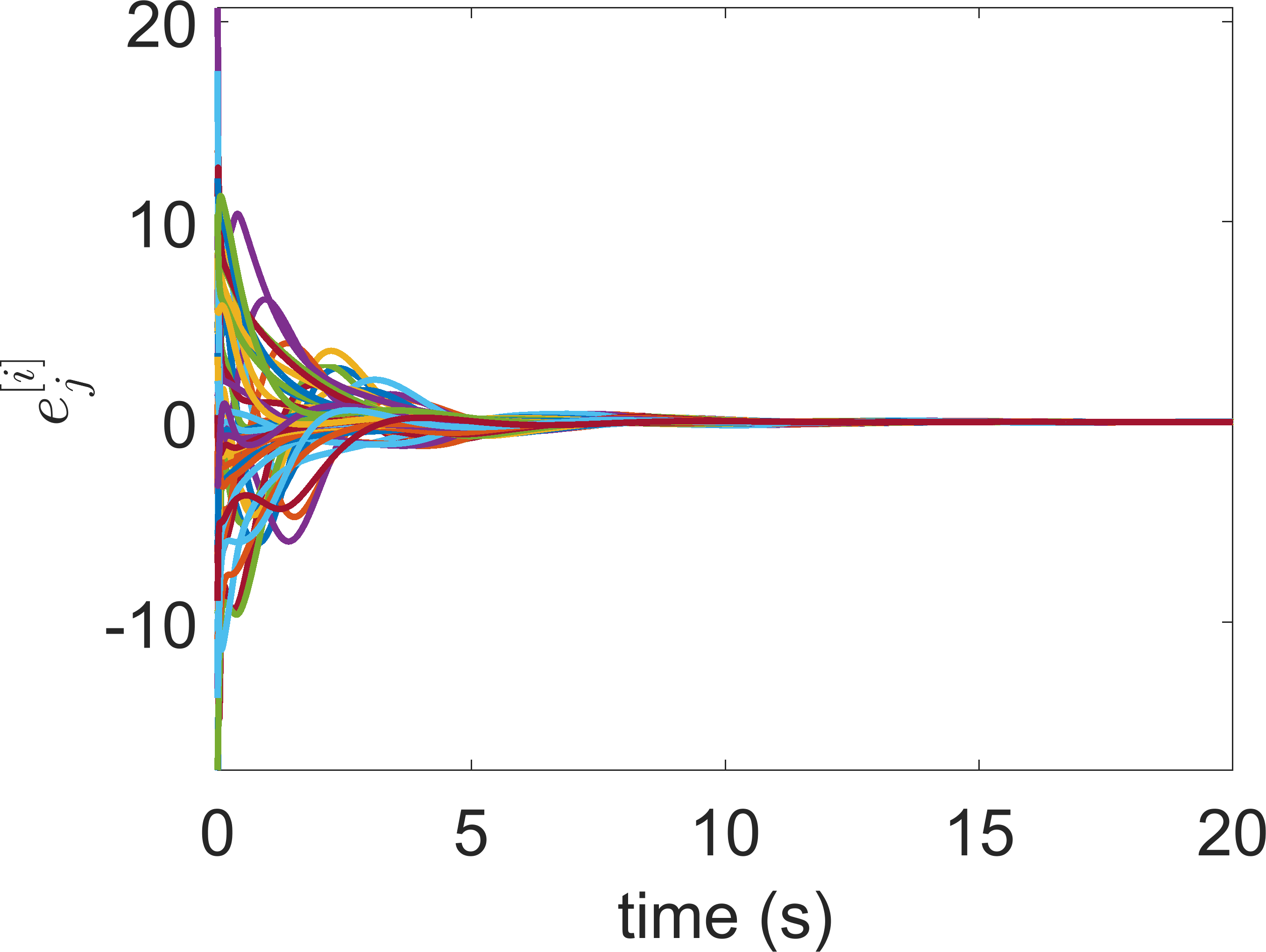}
		\label{fig:h4_201020_1843_n2N21_error}
	} \\
	\subfigure[]{
		\includegraphics[width=0.48\columnwidth]{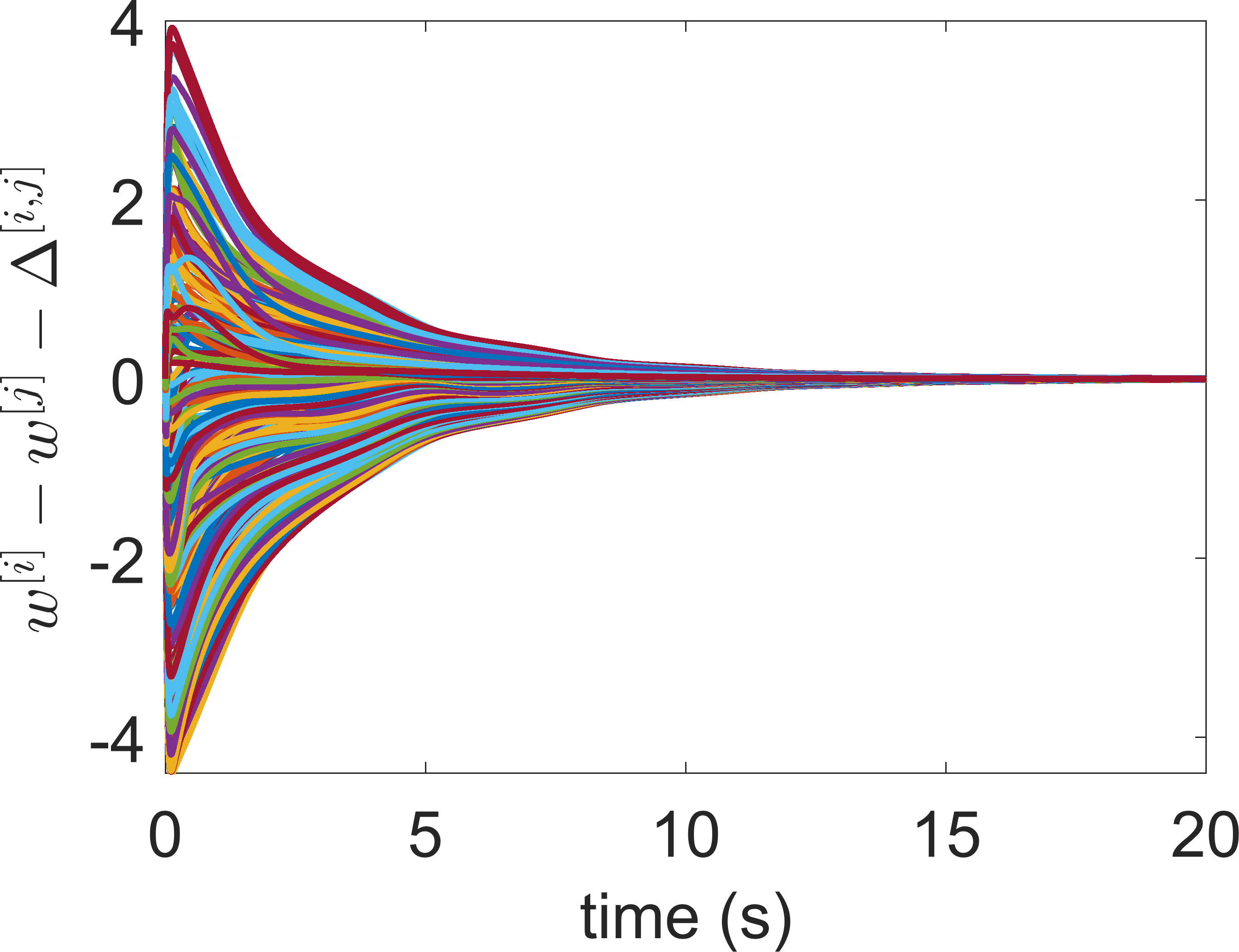}
		\label{fig:h4_201020_1843_n2N21_w}
	} \hspace{-1.2em}
	\subfigure[]{
		\includegraphics[width=0.48\columnwidth]{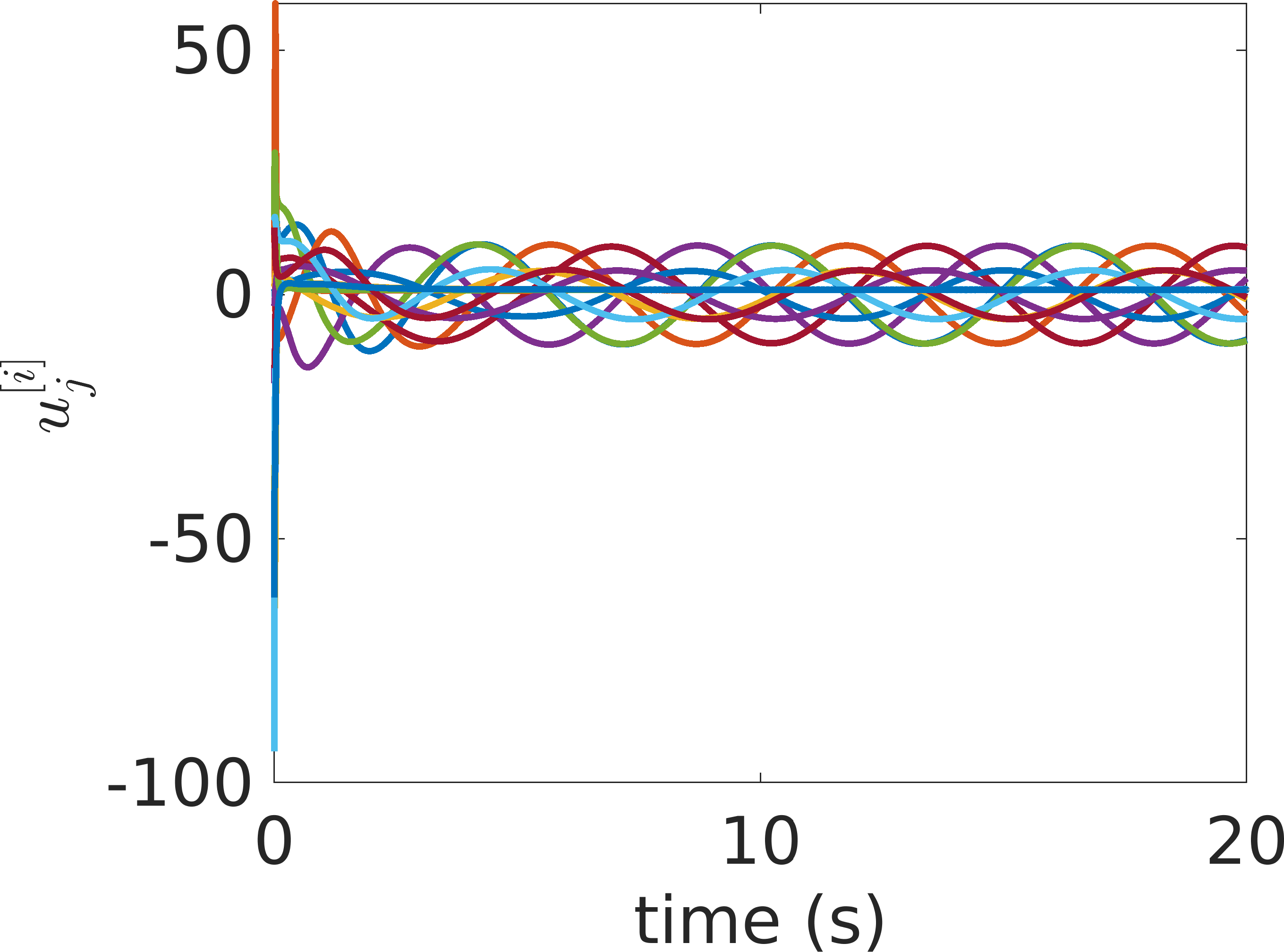}
		\label{fig:h4_201020_1843_n2N21_control}
	}
	\caption{The third simulation results. \subref{fig:h4_201020_1843_n2N21_trj} The trajectories of robots, where squares and $*$ symbolize the trajectories' initial and final positions respectively. The dashed black line shows the communication links between robots (i.e., a cycle graph). \subref{fig:h4_201020_1843_n2N21_error} The path-following errors $\phi^{[i]}_j$ for $i\in \mathbb{Z}_{1}^{21}$ and $j\in \mathbb{Z}_{1}^{2}$. \subref{fig:h4_201020_1843_n2N21_w} The coordination error $w^{[i]}-w^{[j]}-\Delta^{[i,j]}$ for $i,j\in \mathbb{Z}_{1}^{21},i<j$. \subref{fig:h4_201020_1843_n2N21_control} The control inputs $u^{[i]}_j \defeq \vfcb^{[i]}_j$ for $i=1,6,11,16,21$, $j\in \mathbb{Z}_{1}^{2+1}$.  Only 20 seconds of data are shown for clarity in \subref{fig:h4_201020_1843_n2N21_error}-\subref{fig:h4_201020_1843_n2N21_control}. }
	\label{fig:sim3}
\end{figure}
\begin{figure}[htb]
	\centering
	\subfigure[]{
		\includegraphics[width=0.32\columnwidth]{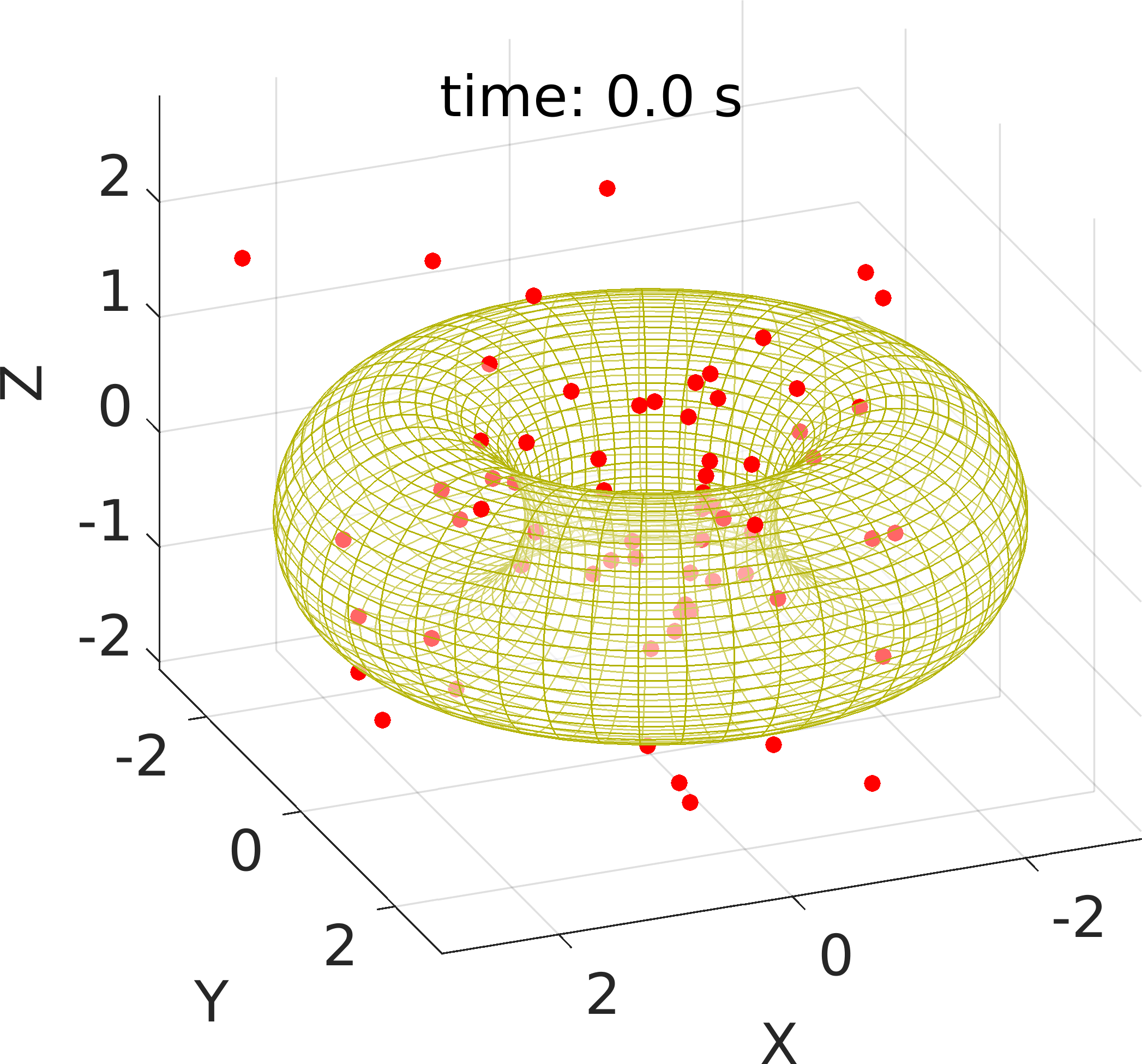}
		\label{fig:h8icra_ode20210429_1420_n3N67_trj1}
	}\hspace{-1.2em}
	\subfigure[]{
		\includegraphics[width=0.32\columnwidth]{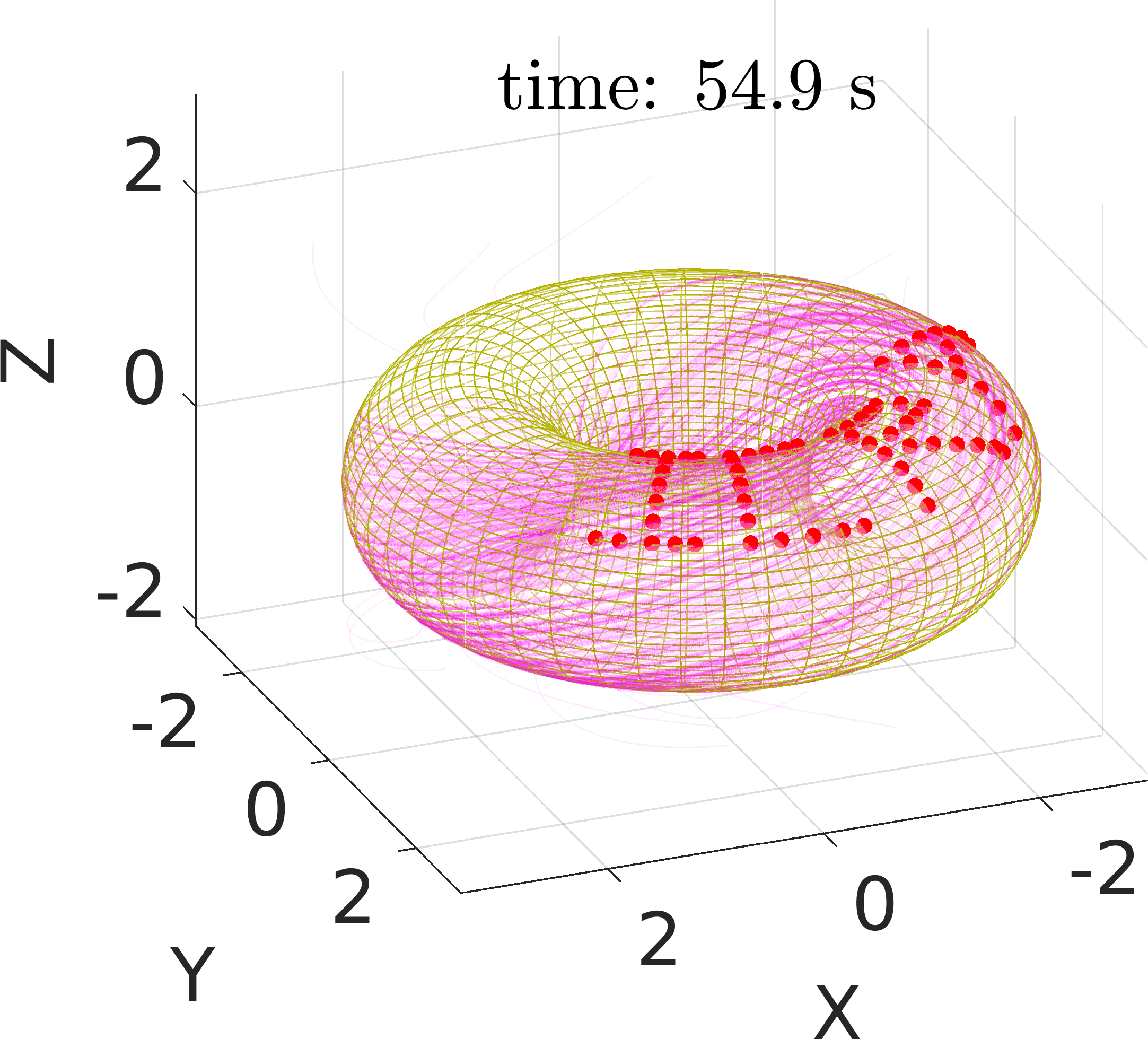}
		\label{fig:h8icra_ode20210429_1420_n3N67_trj2}
	} \hspace{-1.2em}
	\subfigure[]{
		\includegraphics[width=0.32\columnwidth]{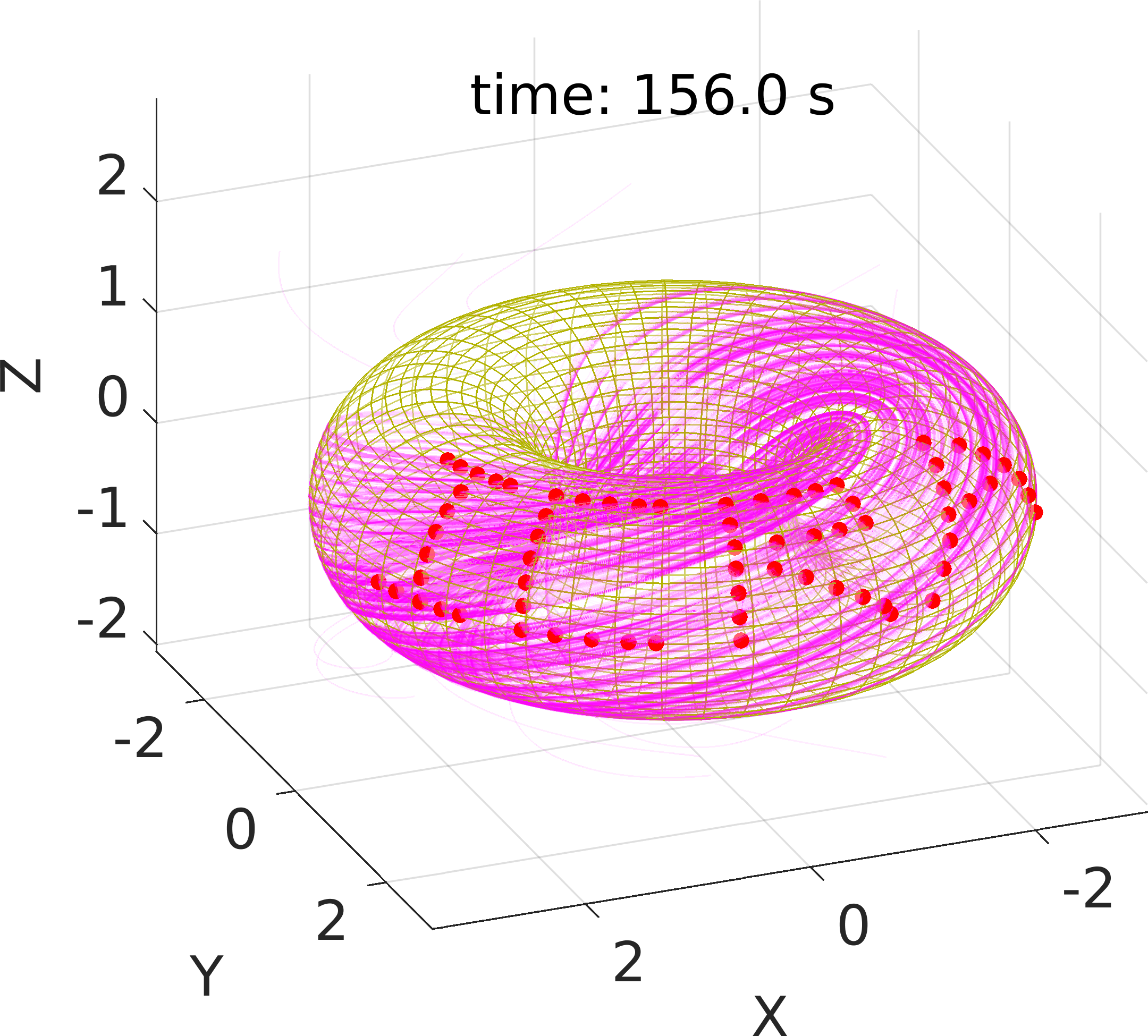}
		\label{fig:h8icra_ode20210429_1420_n3N67_trj3}
	} \\
	\subfigure[]{
		\includegraphics[width=0.32\columnwidth]{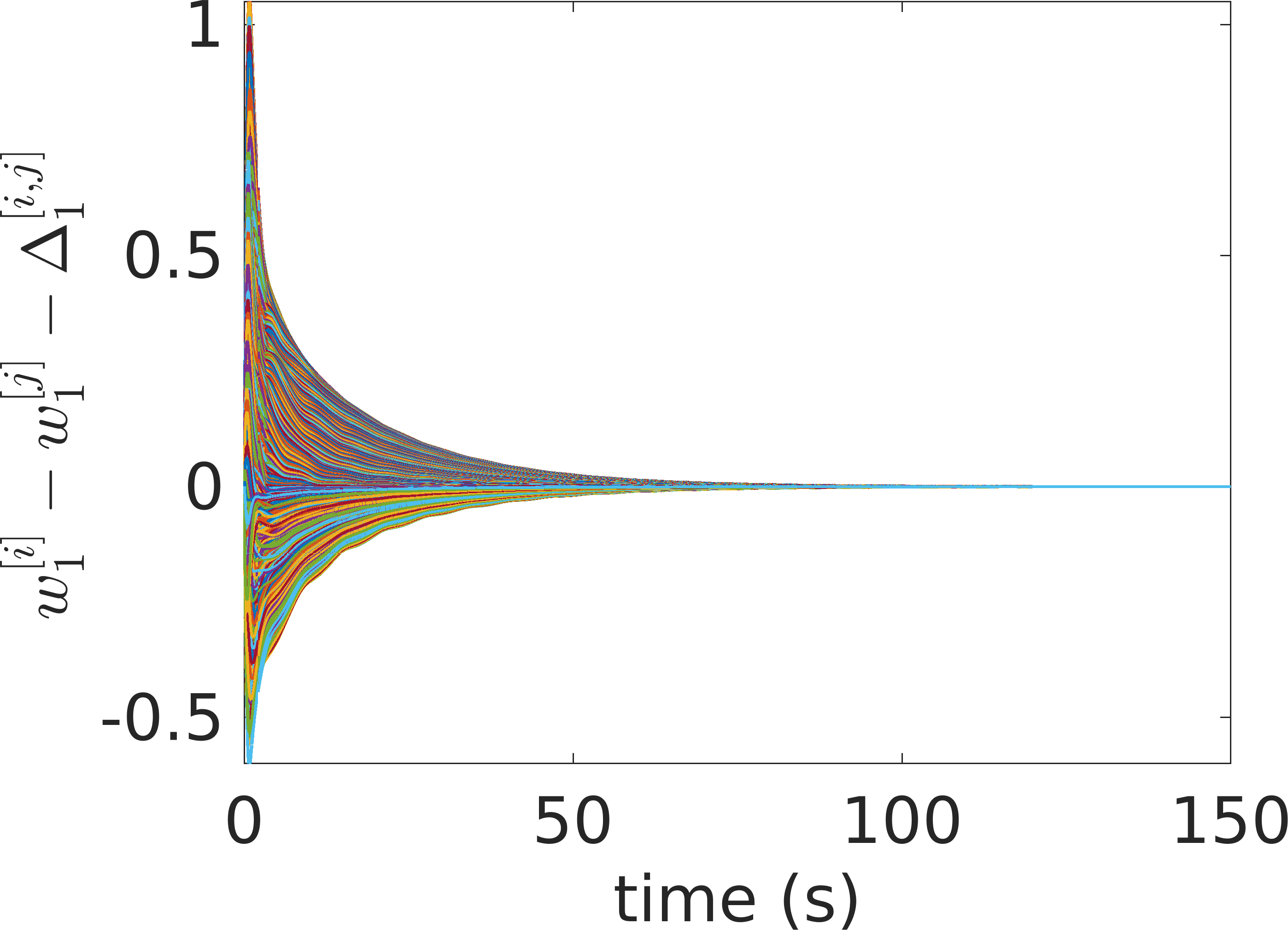}
		\label{fig:h8icra_ode20210429_1420_n3N67_w1diff}
	} \hspace{-1.2em}
	\subfigure[]{
		\includegraphics[width=0.32\columnwidth]{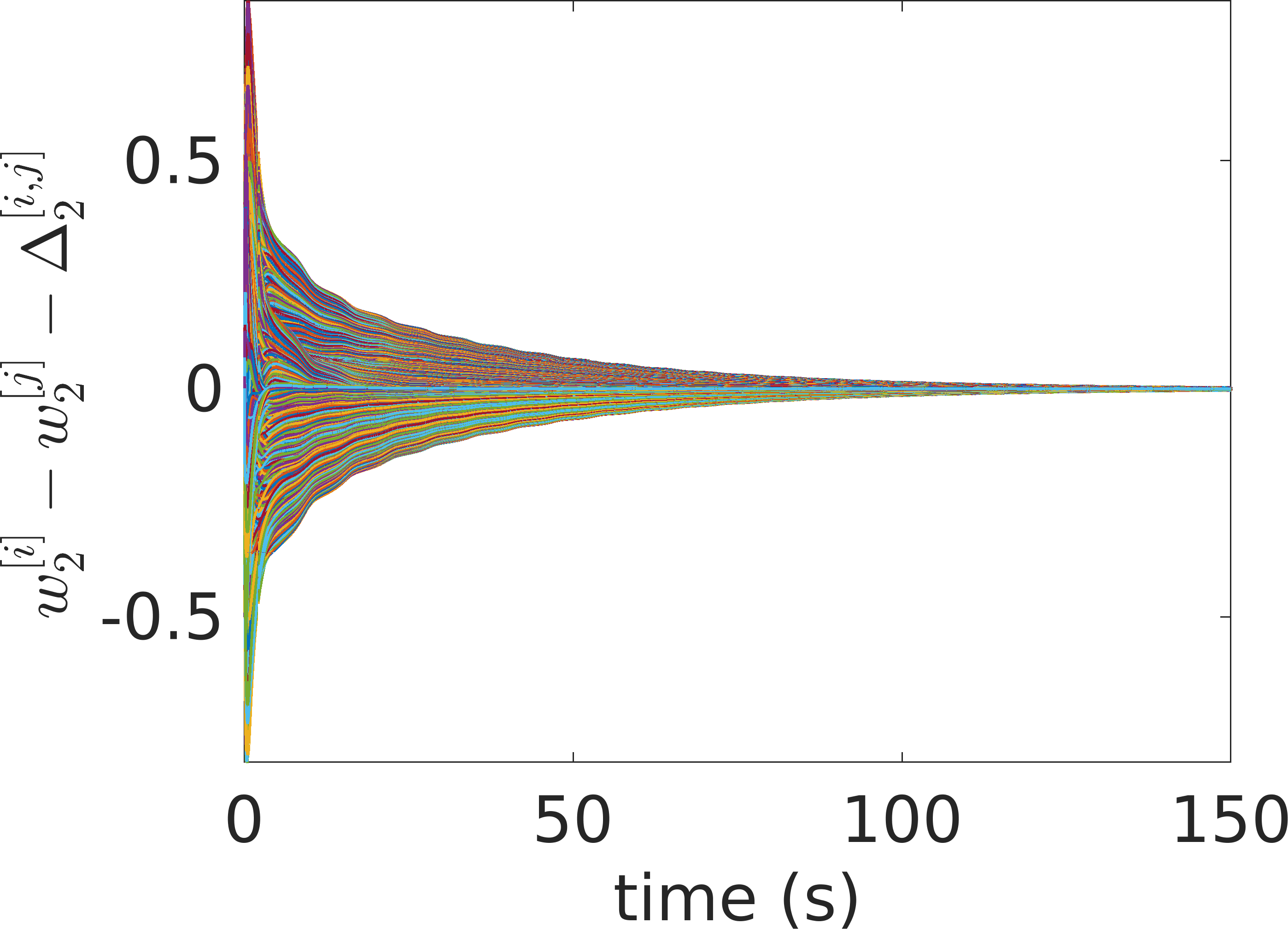}
		\label{fig:h8icra_ode20210429_1420_n3N67_w2diff}
	} \hspace{-1.2em}
	\subfigure[]{
		\includegraphics[width=0.32\columnwidth]{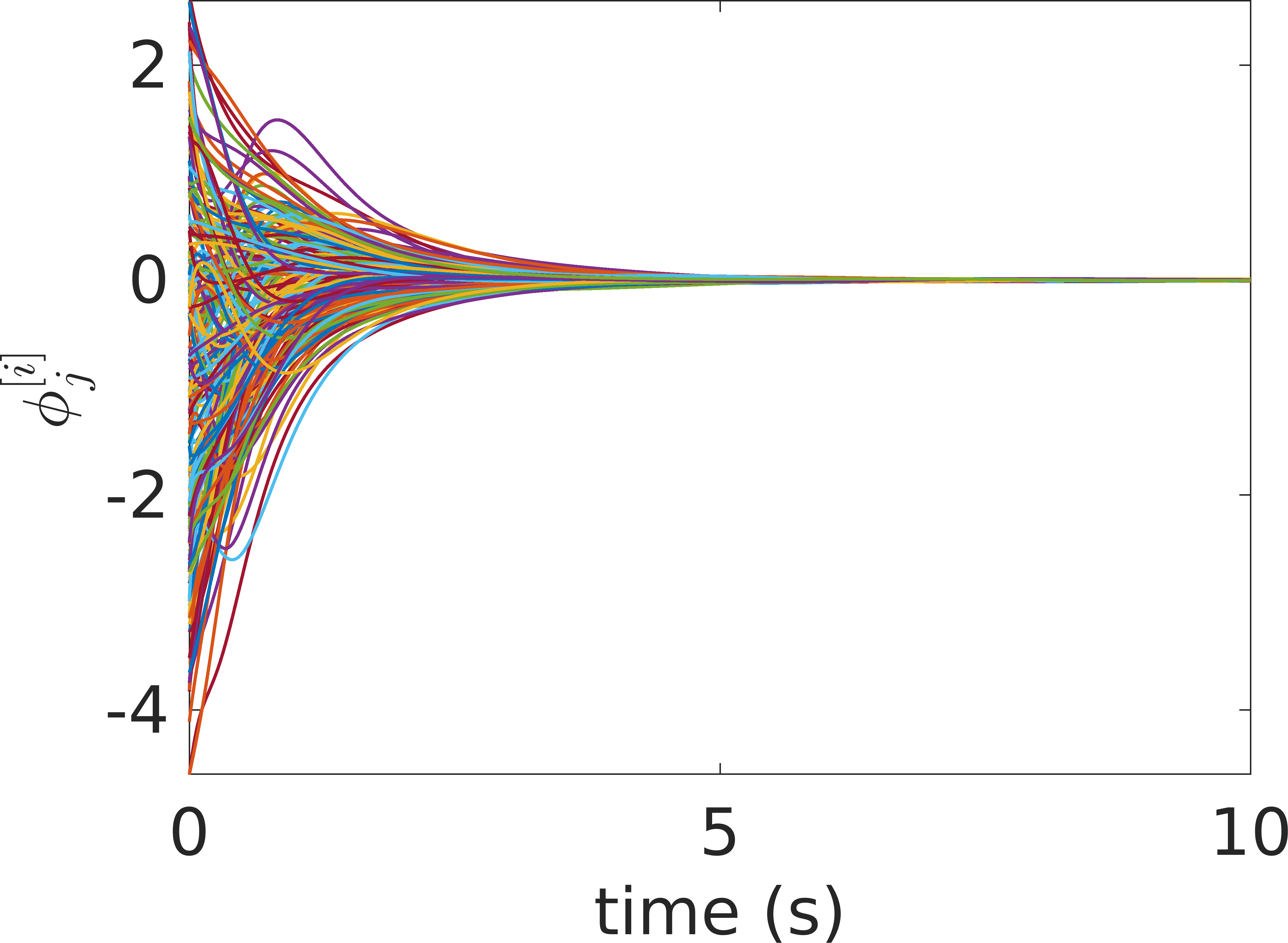}
		\label{fig:h8icra_ode20210429_1420_n3N67_error}
	} \\
	\subfigure[]{
		\includegraphics[width=0.32\columnwidth]{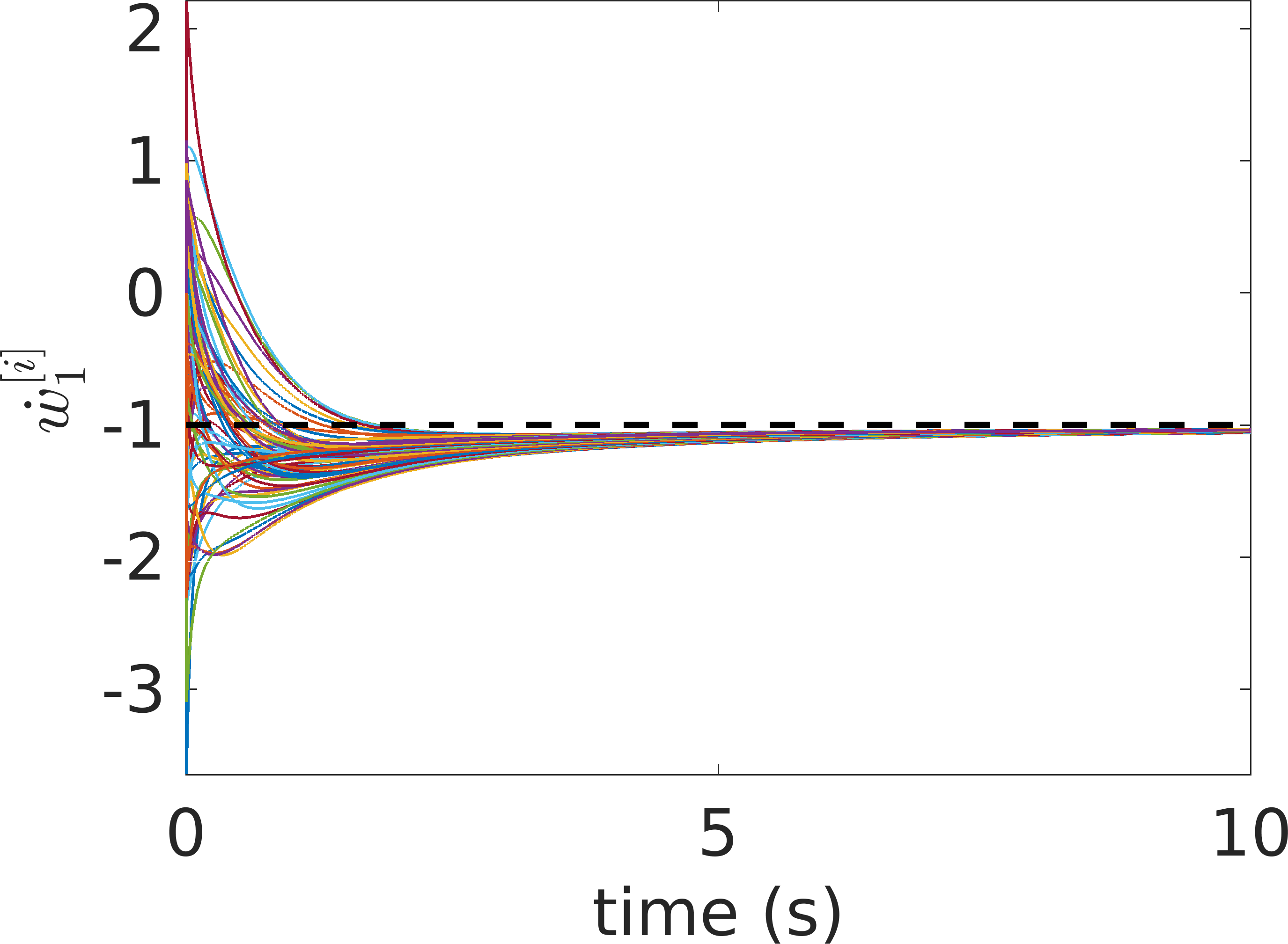}
		\label{fig:h8icra_ode20210429_1420_n3N67_wdot1}
	} \hspace{-1.2em}
	\subfigure[]{
		\includegraphics[width=0.32\columnwidth]{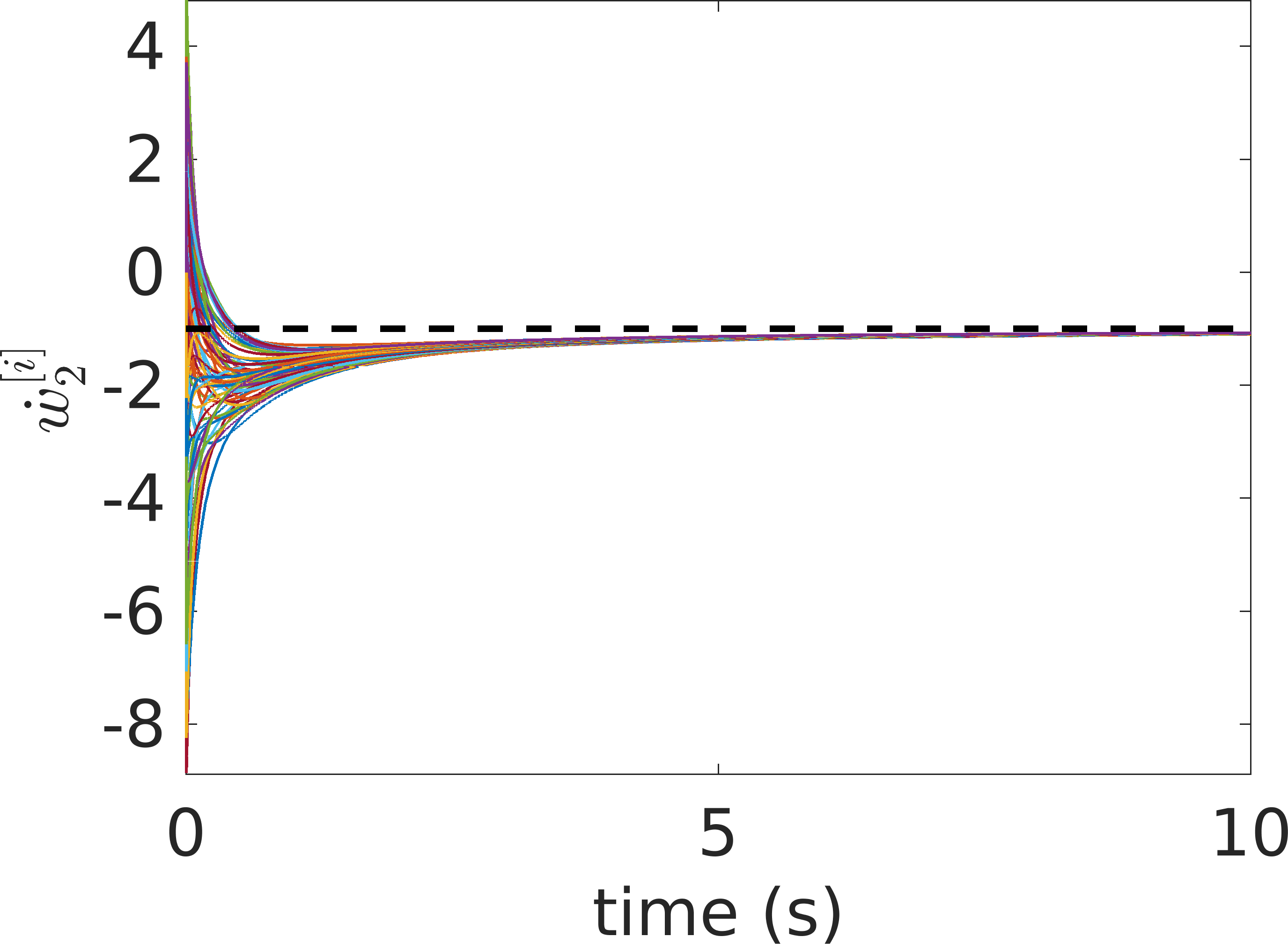}
		\label{fig:h8icra_ode20210429_1420_n3N67_wdot2}
	} \hspace{-1.2em}
	\subfigure[]{
		\includegraphics[width=0.32\columnwidth]{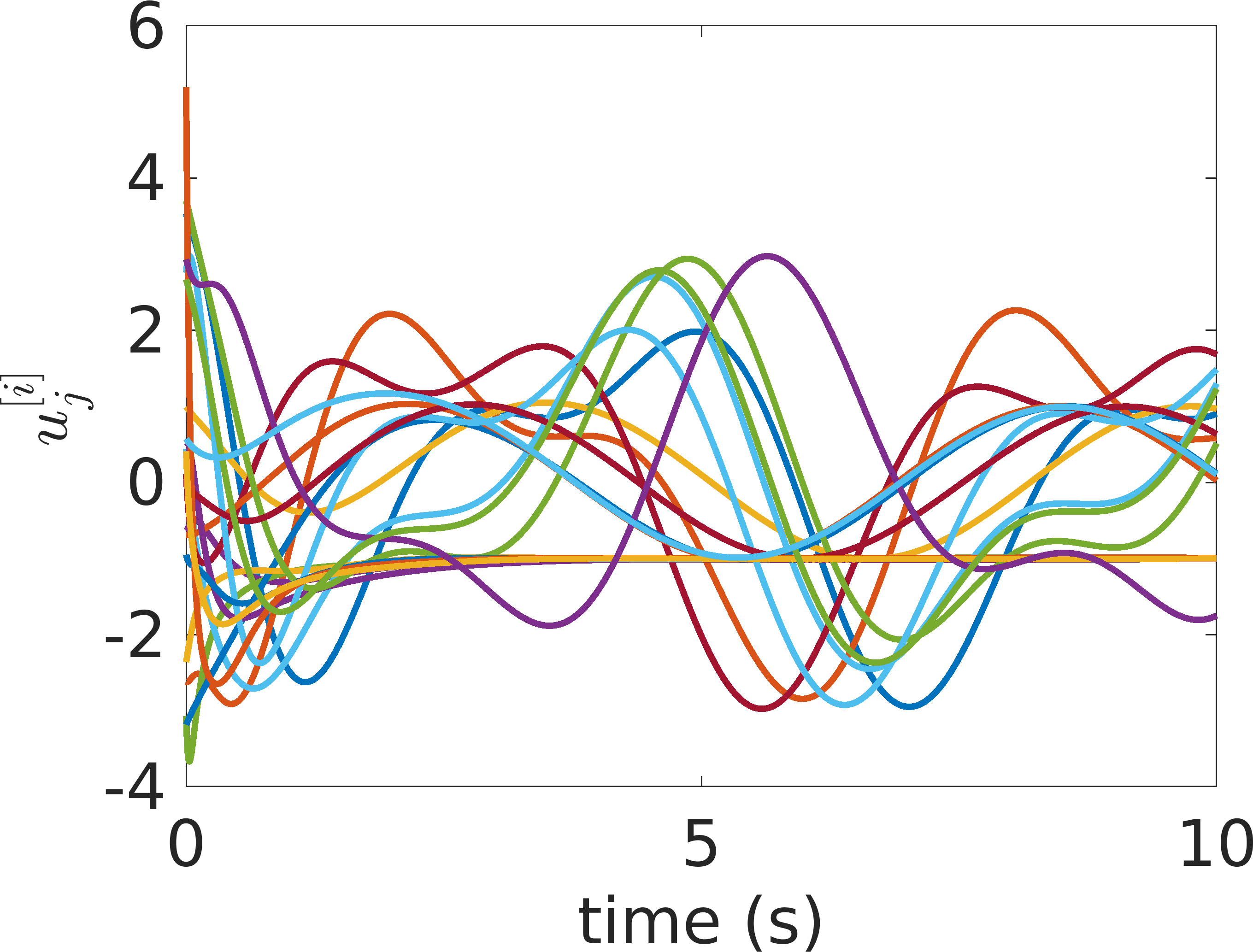}
		\label{fig:h8icra_ode20210429_1420_n3N67_control}
	}
	\caption{The fourth simulation results. \subref{fig:h8icra_ode20210429_1420_n3N67_trj1}-\subref{fig:h8icra_ode20210429_1420_n3N67_trj3} Sixty-seven robots, represented by red dots, converge to and maneuver on a torus while they form the pattern ``\texttt{ICRA}'' at different time instances. The magenta curves are the trajectories of robots. \subref{fig:h8icra_ode20210429_1420_n3N67_w1diff}-\subref{fig:h8icra_ode20210429_1420_n3N67_w2diff} The coordination errors $\mw{i}{1}-\mw{j}{1}-\Delta^{[i,j]}_{1}$ and $\mw{i}{2}-\mw{j}{2}-\Delta^{[i,j]}_{2}$ for the two parameters for $i,j\in \mathbb{Z}_{1}^{67},i<j$ converge to zero eventually. \subref{fig:h8icra_ode20210429_1420_n3N67_error} The surface-convergence errors $\mphi{i}{j}$ for $i\in \mathbb{Z}_{1}^{67}$ and $j\in \mathbb{Z}_{1}^{3}$ converge to zero eventually. \subref{fig:h8icra_ode20210429_1420_n3N67_wdot1}-\subref{fig:h8icra_ode20210429_1420_n3N67_wdot2} The parametric velocities $\mdotw{i}{1}$ and  $\mdotw{i}{2}$ converge to the desired values $\dot{w}^{*}_1=\dot{w}^{*}_2=-1$ denoted by black dashed lines. \subref{fig:h8icra_ode20210429_1420_n3N67_control} The control inputs $u^{[i]}_j \defeq \vfcb^{[i]}_j$ for $i=1,17,34,50,67$, $j\in \mathbb{Z}_{1}^{3+1}$. Only $10$ seconds of data are shown for clarity in \subref{fig:h8icra_ode20210429_1420_n3N67_error}-\subref{fig:h8icra_ode20210429_1420_n3N67_control}. }
	\label{fig:sim4}
\end{figure}
In all simulations, the communication topology is a \emph{cycle graph}, and thus each robot is only allowed to communicate with its two adjacent neighbors. %

In the first simulation, we let $N=50$ robots follow a 3D (i.e., $n=3$) \emph{self-intersected} bent ``$\infty$''-shaped curve parameterized by $\mx{i}{1}=15 \sin(2 w^{[i]})$, $\mx{i}{2}=30 \sin(w^{[i]}) \sqrt{ 0.5 (1 - 0.5 \sin^2(w^{[i]})) }$ and $\mx{i}{3}=5 + 5 \cos(2 w^{[i]}) - 2$ for $i\in \mathbb{Z}_{1}^{N}$. The period of this closed curve is $T=2 \pi$ and the desired differences between two adjacent robots' virtual coordinates are %
$T/(2N)$. We construct the desired parametric differences $\Delta^{[i,j]}$ from the reference $w^{[i]*}=(i-1) T/(2N)$ for $i\in \mathbb{Z}_{1}^{N}$. The control gains for the coordinating vector field are: $\mk{i}{1}=\mk{i}{2}=\mk{i}{3}=1,  k_c=300$ for $i \in \mathbb{Z}_{1}^{N}$, where $k_c$ is large to accelerate the motion coordination. As shown in Fig. \ref{fig:sim1}, all robots follow the ``$\infty$''-shaped path successfully and keep desired positions (in terms of $w^{[i]}$) between each other. The path-following errors and the coordination errors converge to zero eventually.

In the second simulation, we aim to show that our algorithm is also applicable to path following control of complicated and \emph{open} curves (i.e., aperiodic curve), and demonstrate its potential application to volume coverage in 3D. To this end, we choose a 3D Lissajous curve (i.e., $n=3$) with \emph{irrational} coefficients, of which the parametric equations are $\mx{i}{1}=\cos(n_x w^{[i]}) + m_x$, $\mx{i}{2}=\cos(n_y w^{[i]}) + m_y$ and $\mx{i}{3}=\cos(n_z w^{[i]}) + m_z$ for  $i\in \mathbb{Z}_{1}^{N}$, and the coefficients are chosen as $n_x=\sqrt{2}, n_y=4.1, n_z=7.1, m_x=0.1, m_y=0.7, m_z=0$. This is an open curve bounded in a cube (as $n_x$ is irrational). Therefore, it is ideal for a volume coverage task. To clearly illustrate this idea, we choose $N=3$ robots and simulate for $100$ seconds. The control gains for the coordinating guiding vector field are: $\mk{i}{1}=\mk{i}{2}=\mk{i}{3}= k_c=1$ for $i \in \mathbb{Z}_{1}^{N}$. We construct the desired parametric differences $\Delta^{[i,j]}$ from the references $w^{[i]*}(t)=(i-1) 2\pi/N$, for $i\in \mathbb{Z}_{1}^{N}$. Since the Lissajous curve will fill the whole cube as the path parameter $w^{[i]}$ varies from $0$ to infinity, for illustration purpose, we only plot part of the desired path where $w^{[i]} \in [0, 30\pi]$ (see the magenta curve in Fig. \ref{fig:sim2_1}). As seen in Fig. \ref{fig:sim2_1}, the three robots tend to cover the whole volume of the unit cube, and the path-following errors and coordination errors are almost zero after $20$ seconds.

In the third simulation, we show how different robots can follow different desired paths while they still coordinate their motions to form some formation shapes. We let $N=21$ robots follow three different paths, where the first seven robots follow a large circle of radius $a>0$ parameterized by $\mx{i}{1}=a \cos w^{[i]}$, $\mx{i}{2}= a \sin w^{[i]}$ for $i=1,\dots,7$, the last seven robots follow a small circle of radius $0<b<a$ parameterized by $\mx{i}{1}=b \cos w^{[i]}$, $\mx{i}{2}= b \sin w^{[i]}$ for $i=15,\dots,21$, and the remaining seven robots follow an ellipse with a semi-major axis $a$ and a semi-minor axis $b$, parameterized by $\mx{i}{1}= a \cos w^{[i]}$, $\mx{i}{2} = b \sin w^{[i]}$ for $i=8, \dots, 14$. These three paths are concentric (see Fig. \ref{fig:sim3}). The robots are coordinated in a distributed way such that they are equally separated in the path parameter $w^{[i]}$; i.e., we construct the desired parametric differences $\Delta^{[i,j]}$ from the references $w^{[i]*}(t)=(i-1) 2\pi/N$, for $i\in \mathbb{Z}_{1}^{N}$. Other parameters are chosen as $a=10, b=5, \mk{i}{1}=\mk{i}{2}=1, k_c=100$ for $i\in \mathbb{Z}_{1}^{N}$. An interesting feature is that during the steady-state, these robots will not overlap with each other, since overlapping happens if any two of the robots' virtual coordinates are equal, while the distributed coordination guarantees that the adjacent neighbors satisfy $|w^{[i]} - w^{[j]}|=2\pi/N$ for $(i,j) \in \mathcal{E}$. These robots successfully generate varying formation shapes: they cluster in different parts of the desired paths. %

In the fourth simulation, we show how multiple robots converge to a torus and form a desired pattern while they maneuver on the torus. Specifically, we manually select $N=67$ reference points $(w^{[i]*}_1, w^{[i]*}_2), i \in \mathbb{Z}_{1}^{N}$ such that they form a pattern resembling the four letters ``\texttt{ICRA}''. Then based on these reference points, one can calculate the desired parametric differences $\bm{\Delta^{*}_{1}}, \bm{\Delta^{*}_{2}} \in \mbr[67]$. Note that Robot $i$ only needs to know the desired parametric differences with respect to its neighbors; i.e., $\Delta^{[i,j]}_1, \Delta^{[i,j]}_2$ for $j \in \mathcal{N}_{i}$. These $67$ robots are required to converge to a torus of which the parametric equations are $\mx{i}{1}=(2+\cos \mw{i}{1}) \cos \mw{i}{2}$, $\mx{i}{2}=(2+\cos \mw{i}{1} ) \sin \mw{i}{2}$, $ \mx{i}{3} = \sin \mw{i}{1}$ for $i \in \mathbb{Z}_{1}^{N}$ (hence $n=3$ in Theorem \ref{thm_vf_invariant_surf}). They are also required to maneuver on the torus in the sense that $\dot{w}^{*}_{1}=\dot{w}^{*}_{2}=-1$. By Theorem \ref{thm_vf_invariant_surf}, the extra vector $\bm{v}$ is thus chosen as $\bm{v}=(0,0,0,-1,1)$. Other parameters are chosen as $\mk{i}{1}=\mk{i}{2}=1, k_c=10$ for $i\in \mathbb{Z}_{1}^{N}$. The simulation results are shown in Fig. \ref{fig:sim4}.

\begin{figure}[tb]
	\centering
	\includegraphics[width=1\columnwidth]{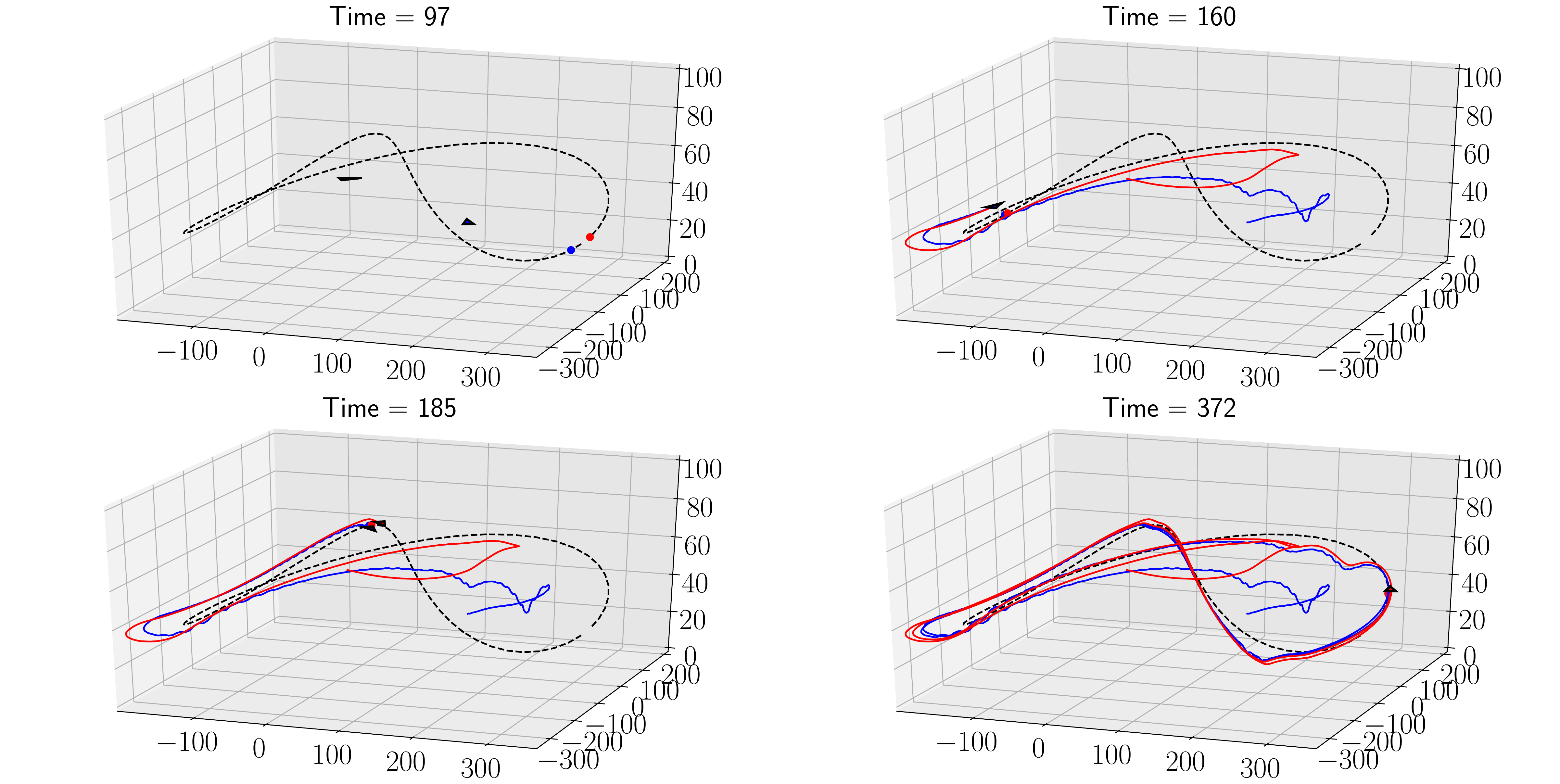}
	\caption{Two aircraft (blue and red trajectories) fly together and follow a 3D bent ``$\infty$''-shaped path. Although the subfigures show that the aircraft's positions are overlapped, in reality, they are different since the aircraft's GPS receptors are biased with respect to each other by one meter in the XY plane. The red ($i=1$) and blue ($i=2$) dots represent $(f^{[i]}_{1}, f^{[i]}_{2}, f^{[i]}_{3})$. The displayed positions are in meters.}
	\label{fig: lissaplane}
	\vspace{-0.5em}
\end{figure}
\begin{figure}[tb]
	\centering
	\includegraphics[width=0.7\columnwidth]{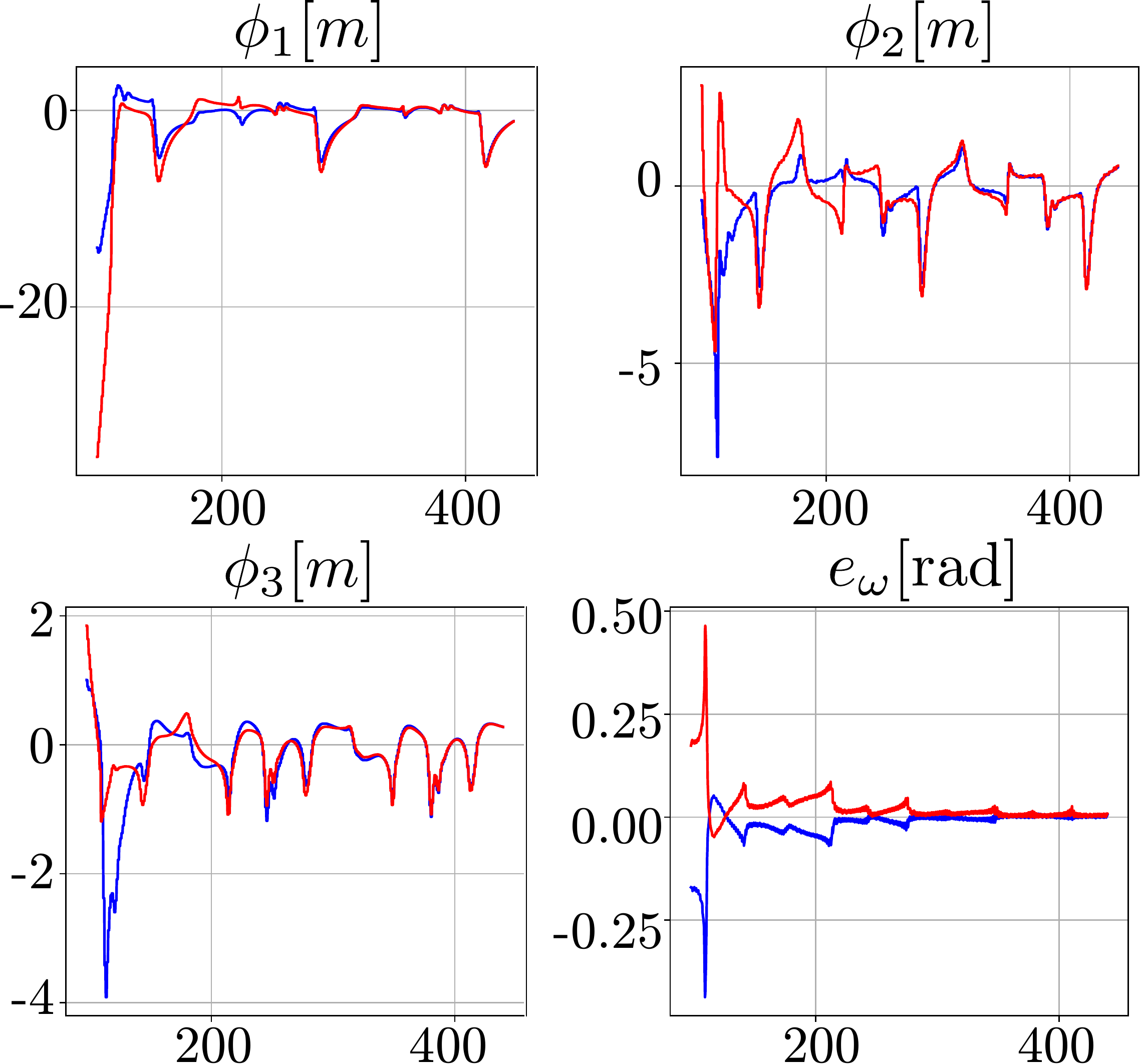}
	\caption{Path-following errors of the two aircraft (blue and red) to the desired path, and the coordination error $e_w$ with respect to the desired $\Delta^{[i,j]} = 0$. The horizontal axes represent time in seconds.}
	\label{fig: planesphi}
	\vspace{-1em}
\end{figure}

\subsection{Experiments with a pair of fixed-wing aircraft}
\subsubsection{Rendezvous on a 3D Lissajous curve}
In this experiment\footnote{The code is available at \url{https://github.com/noether/paparazzi/tree/gvf_param_multi/sw/airborne/modules/guidance/gvf_parametric}.}, two autonomous fixed-wing aircraft (Autonomous Opterra 1.2m similar to the Sonicmodel in Fig. \ref{fig: planes2}) are employed to validate Theorems \ref{thm_vf_invariant} and \ref{thm_guidance}. The aircraft are equipped with open-source software/hardware components developed with \emph{Paparazzi} \cite{gati2013open}, an open-source research community for the development of autonomous, mostly aerial, vehicles. %
We choose the following 3D Lissajous curve $\mf{i}{1}(w) = 225 \cos(w^{[i]}), \mf{i}{2}(w^{[i]}) = 225\cos(2 w^{[i]} + \pi/2), \mf{i}{3}(w^{[i]})= -20\cos(2 w^{[i]})$, for $i=1,2$, which is a bent ``$\infty$''-shaped  path. The mission requires both aircraft to have $\Delta^{[1,2]} =\Delta^{[2,1]}= 0$; i.e., to fly in a tight formation. The overlapping at steady state is avoided by biasing the GPS measurement of one aircraft by a constant distance of one meter in the horizontal plane; i.e., when the aircraft achieve $\Delta^{[1,2]}=\Delta^{[2,1]} = 0$, they are displaced physically. We choose $\mk{i}{1} = \mk{i}{2} = 0.002, \mk{i}{3} = 0.0025$, $k_c=0.01, k_\theta = 1$, for $i=1,2$, and the communication frequency is $10$ Hz. In the experiment, the weather forecast reported 14 degrees Celsius and a South wind of 10 km/h. In Fig. \ref{fig: lissaplane}, the telemetry shows that both aircraft fly in a tight formation and follow the path, and Fig. \ref{fig: planesphi} depicts the path-following and coordination errors.

\subsubsection{Rendezvous on a torus surface}
\begin{figure}[tb]
	\centering
	\includegraphics[width=\columnwidth]{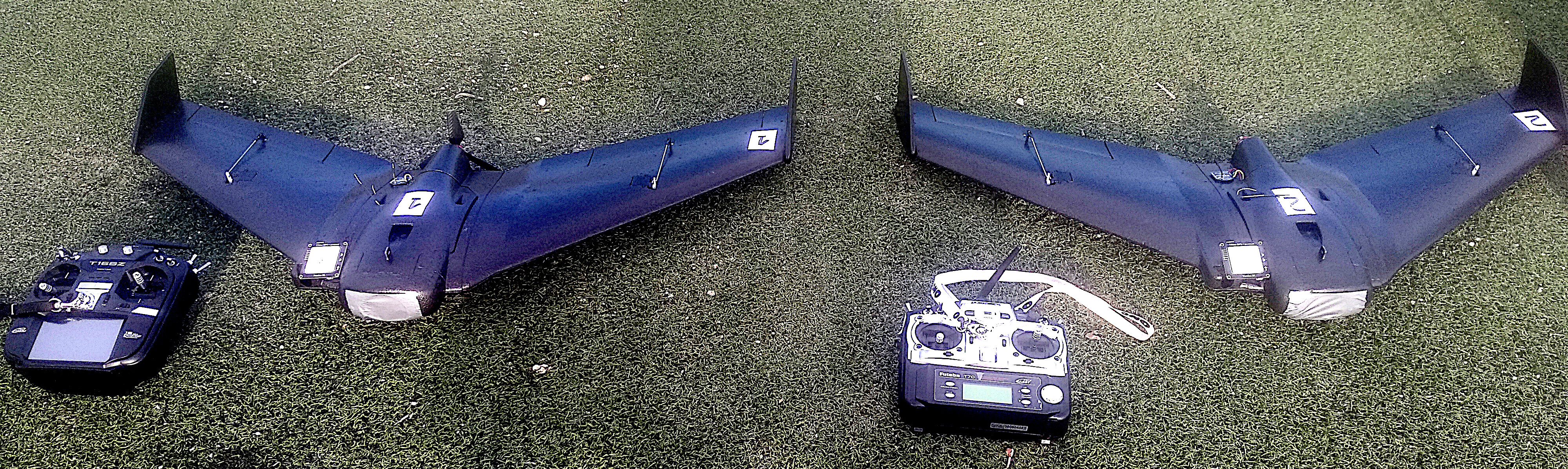}
	\includegraphics[width=\columnwidth]{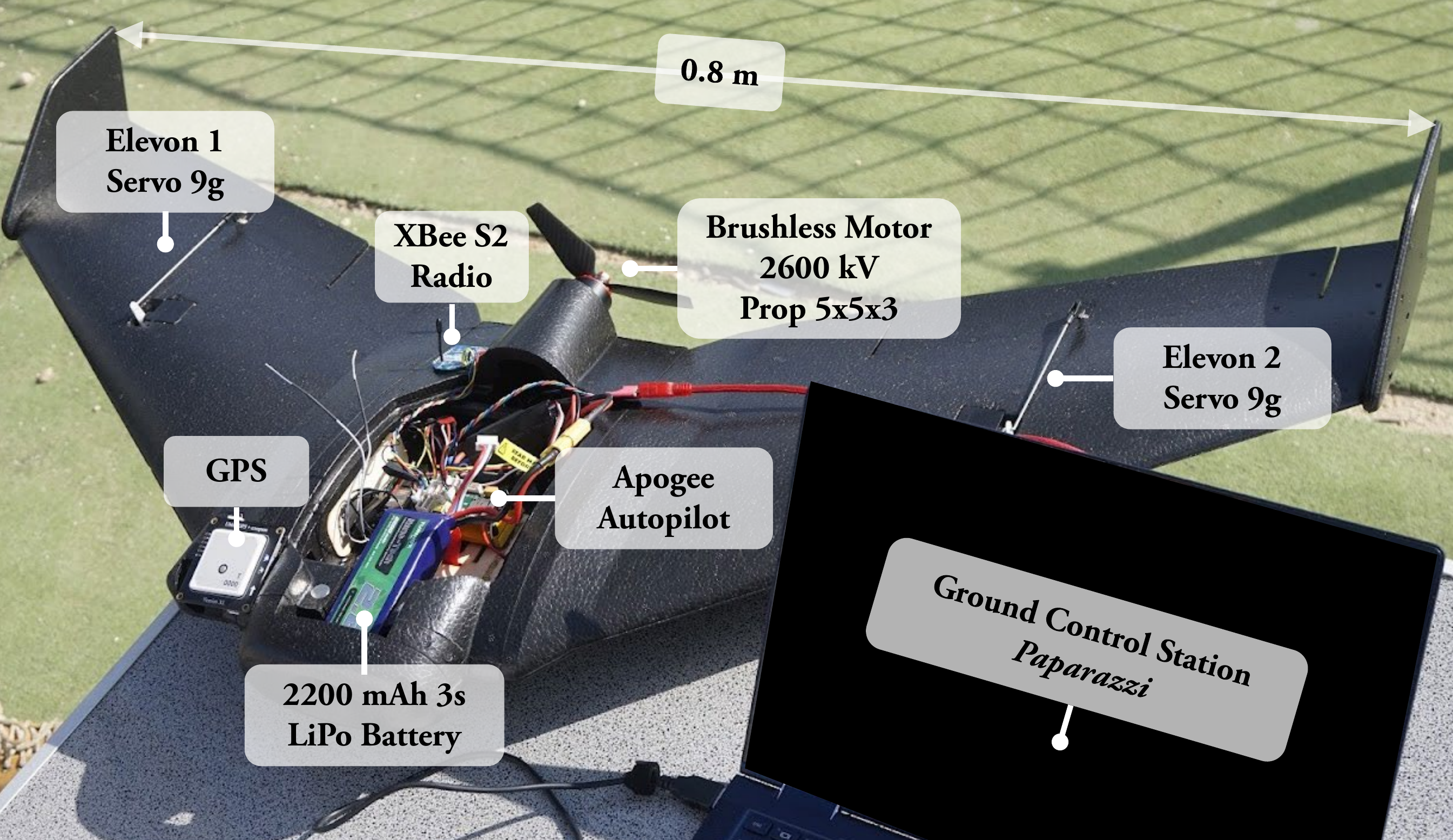}
	\caption{Two aircraft SonicModell employed for the rendezvous on a torus.}
	\label{fig: planes2}
	\vspace{-1em}
\end{figure}

\begin{figure}[tb]
	\centering
	\includegraphics[width=1\columnwidth]{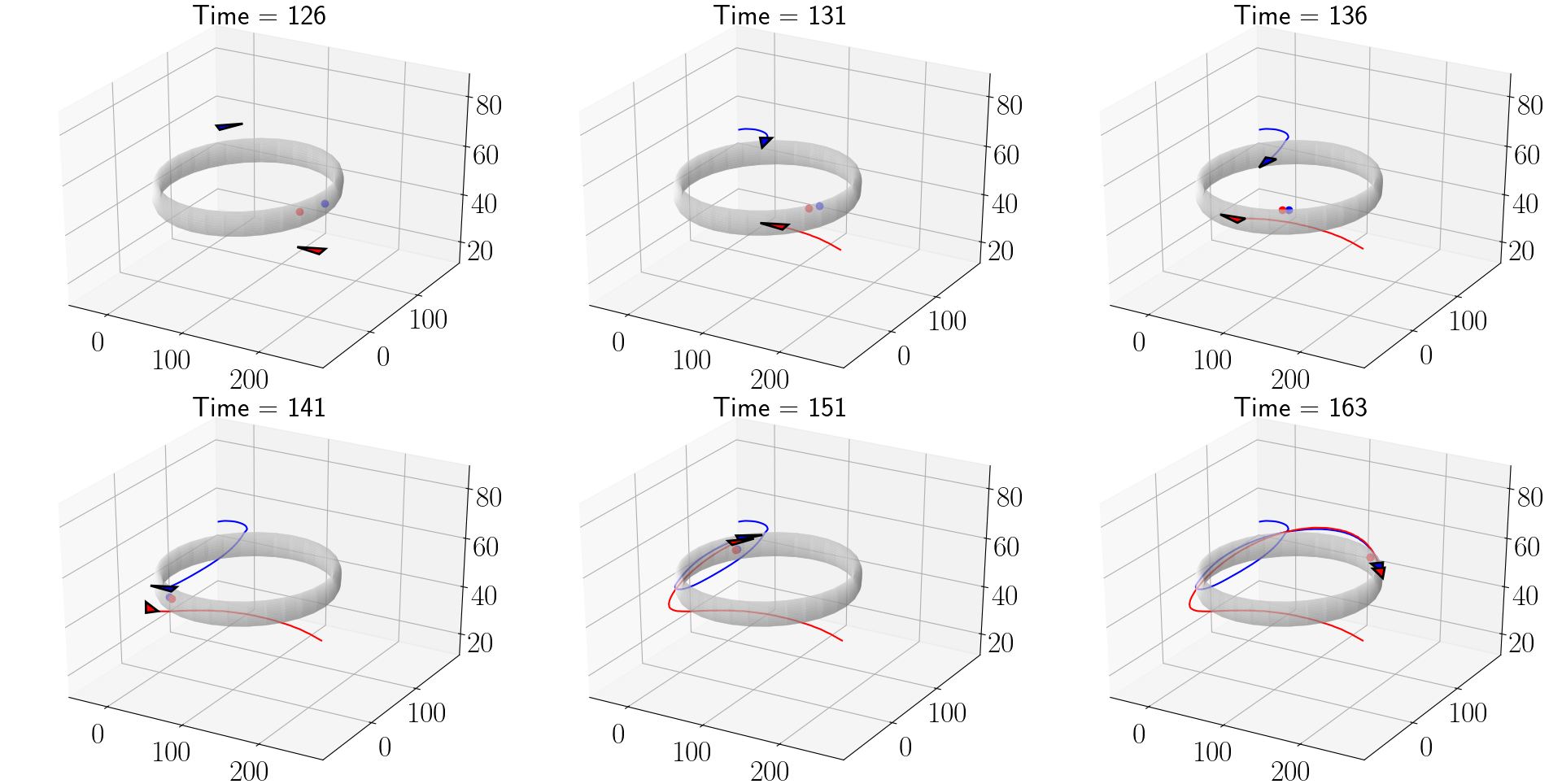}
	\includegraphics[width=1\columnwidth]{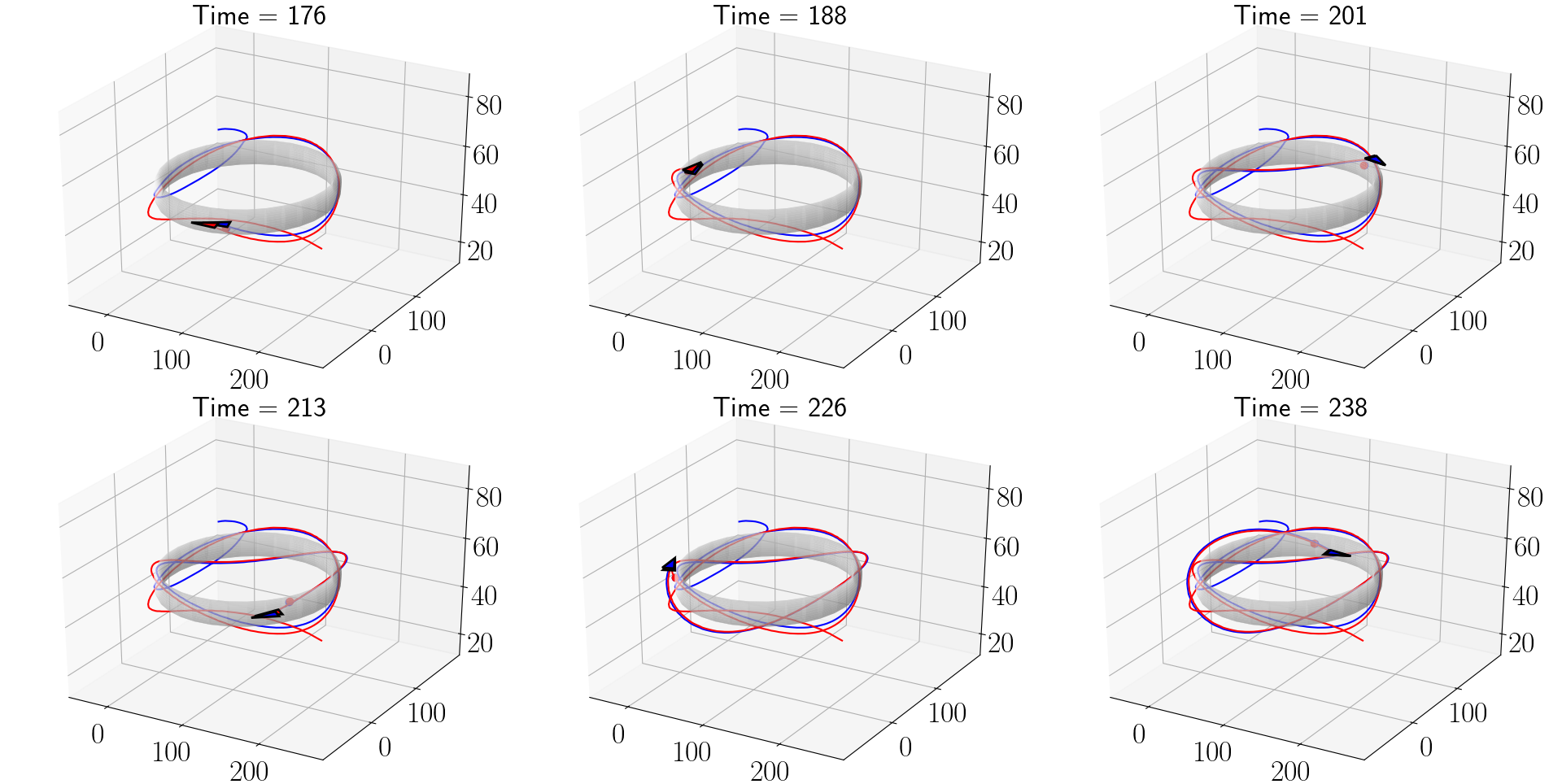}
	\caption{Two aircraft (blue and red trajectories) rendezvous on a torus surface, where the desired trajectory on the torus corresponds to $\dot{w}_2^* = 2 \dot{w}_1^*$. Although the synchronization is done on the same surface, one aircraft has its actual position biased in order to avoid collisions during the rendezvous. The red ($i=1$) and blue ($i=2$) dots represent $(f^{[i]}_{1}, f^{[i]}_{2}, f^{[i]}_{3})$. The displayed positions are in meters.}
	\label{fig: planes2XYZ}
	\vspace{-1em}
\end{figure}

\begin{figure}[tb]
	\centering
	\includegraphics[width=\columnwidth]{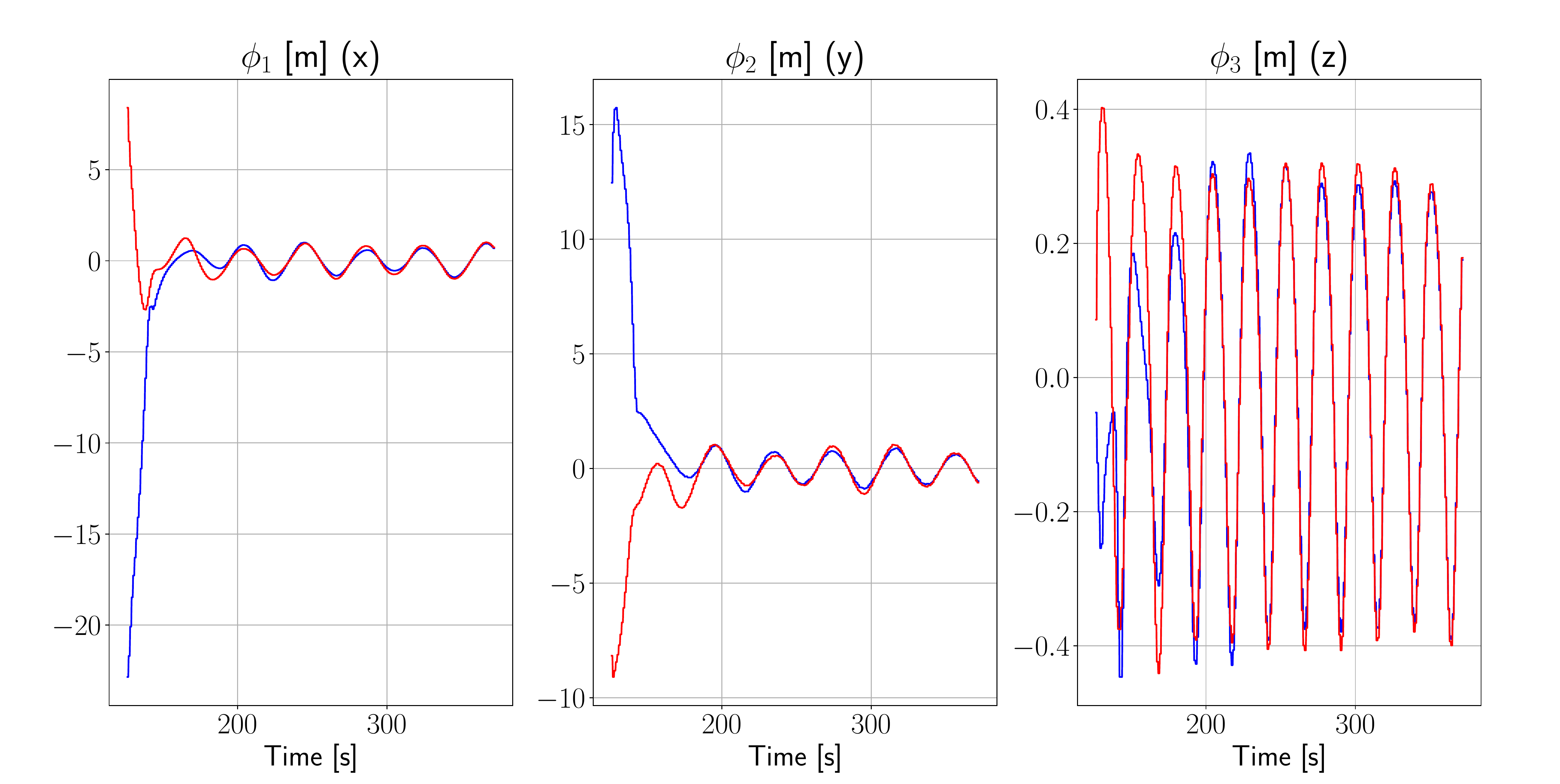}
	\includegraphics[width=\columnwidth]{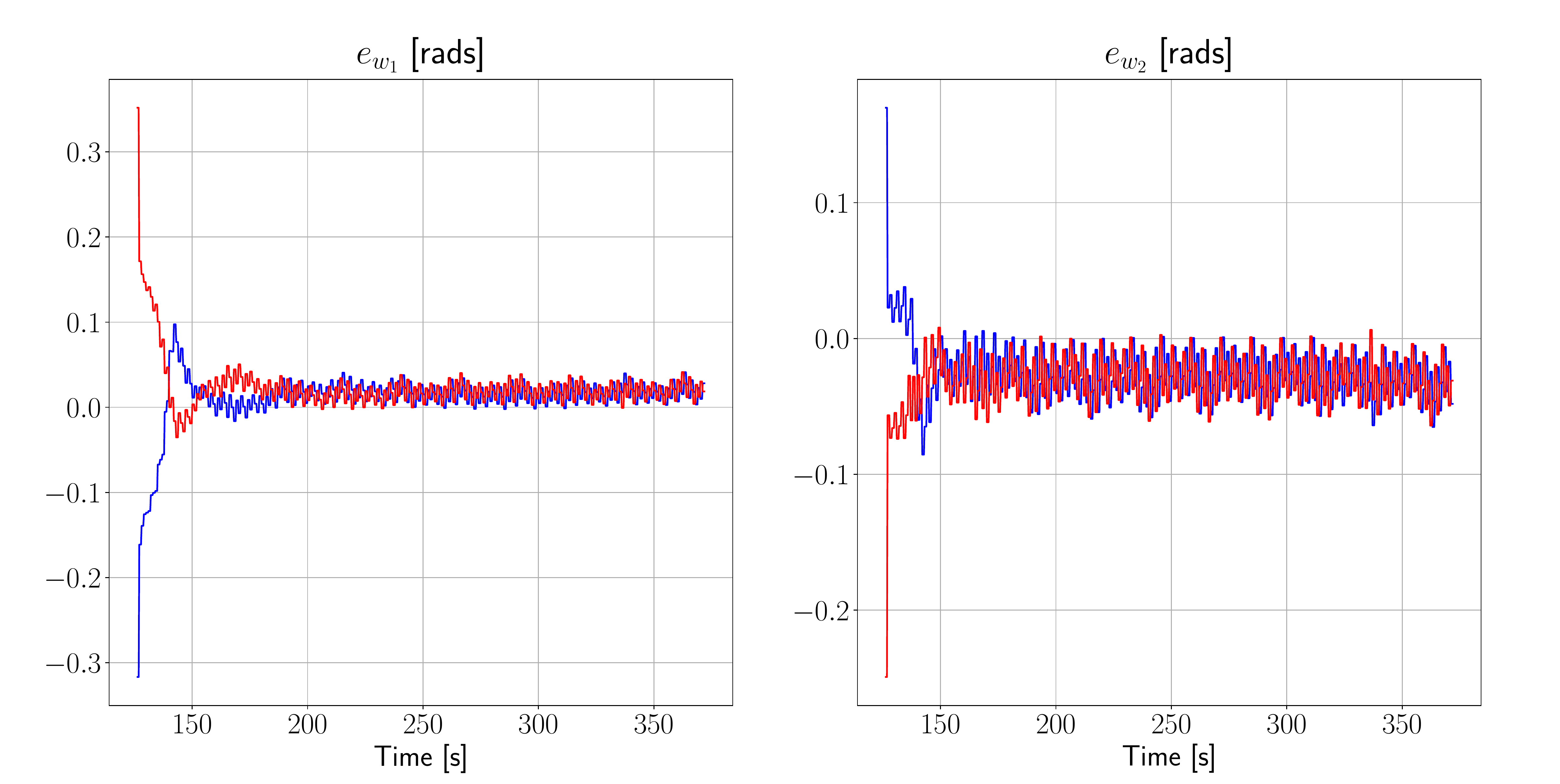}
	\caption{Path-following errors (in XYZ) of the two aircraft (blue and red) to the desired torus surface, and errors $e_{w_1}, e_{w_2}$ with respect to the desired $\Delta_{1}^{[1,2]} =\Delta_{2}^{[1,2]}= 0$. The horizontal axes represent time in seconds.}
	\label{fig: planes2phi}
	\vspace{-1em}
\end{figure}

Two autonomous fixed-wing aircraft (SonicModell in Fig. \ref{fig: planes2}) are employed to validate Theorems \ref{thm_vf_invariant_surf} and \ref{thm_guidance}. In particular, we proceed to show the rendezvous of two aircraft on a torus surface\footnote{The code is available at \url{https://github.com/noether/paparazzi/tree/gvf_param_multi/sw/airborne/modules/guidance/gvf_parametric_surf}. It is designed such that a generic surface can be added by an experienced user.}. %
The torus surface is parametrized with $f_1^{[i]}(w_1,w_2) = (100 + 5\cos(w_2^{[i]})\cos(w_1^{[i]}), f_2^{[i]}(w_1,w_2) = (100 + 5\cos(w_2^{[i]})\sin(w_1^{[i]}), f_3^{[i]}(w_1,w_2) = 5\sin(w_2^{[i]}) + 50$ meters, for $i=1,2$, and we set $\dot{w}_2^* = 2 \dot{w}_1^* = 0.01$; i.e., for each loop in the horizontal dimension, the aircraft will accomplish two loops in the vertical dimension. The rendezvous sets $\Delta_{1}^{[1,2]} =\Delta_{2}^{[1,2]}= 0$, and we have chosen $\mk{i}{1} = \mk{i}{2} = \mk{i}{3} = 0.003$, $k_{c1} = k_{c2} =0.01, k_\theta = 1$, for $i=1,2$, and the air-to-air communication frequency is $10$ Hz. In the experiment, the weather forecast reported 16 degrees Celsius and an East wind of 10 km/h at the radio-control club of Arroyomolinos, Spain. In Fig. \ref{fig: planes2XYZ}, the telemetry shows that both aircraft converge to fly together and follows the surface with the planned trajectory eventually. The oscillation around zero for $\phi_{1,2,3}$ is due to the difficulty on the vertical speed controller for the selected type of aircraft that affects the airspeed. Note that if the \emph{horizontal} speed changes, then the aircraft wait for each other and they do not track precisely the desired $\Delta_{1}^{[1,2]}$ and $\Delta_{2}^{[1,2]}$. Nevertheless, Fig. \ref{fig: planes2phi} shows that the error for the desired altitude does not exceed half a meter, in the horizontal dimensions is around a meter, and the errors for the rendezvous are around one degree in $w_1$ and less than one degree in $w_2$.


\begin{figure}[tb]
	\centering
	\includegraphics[width=\columnwidth]{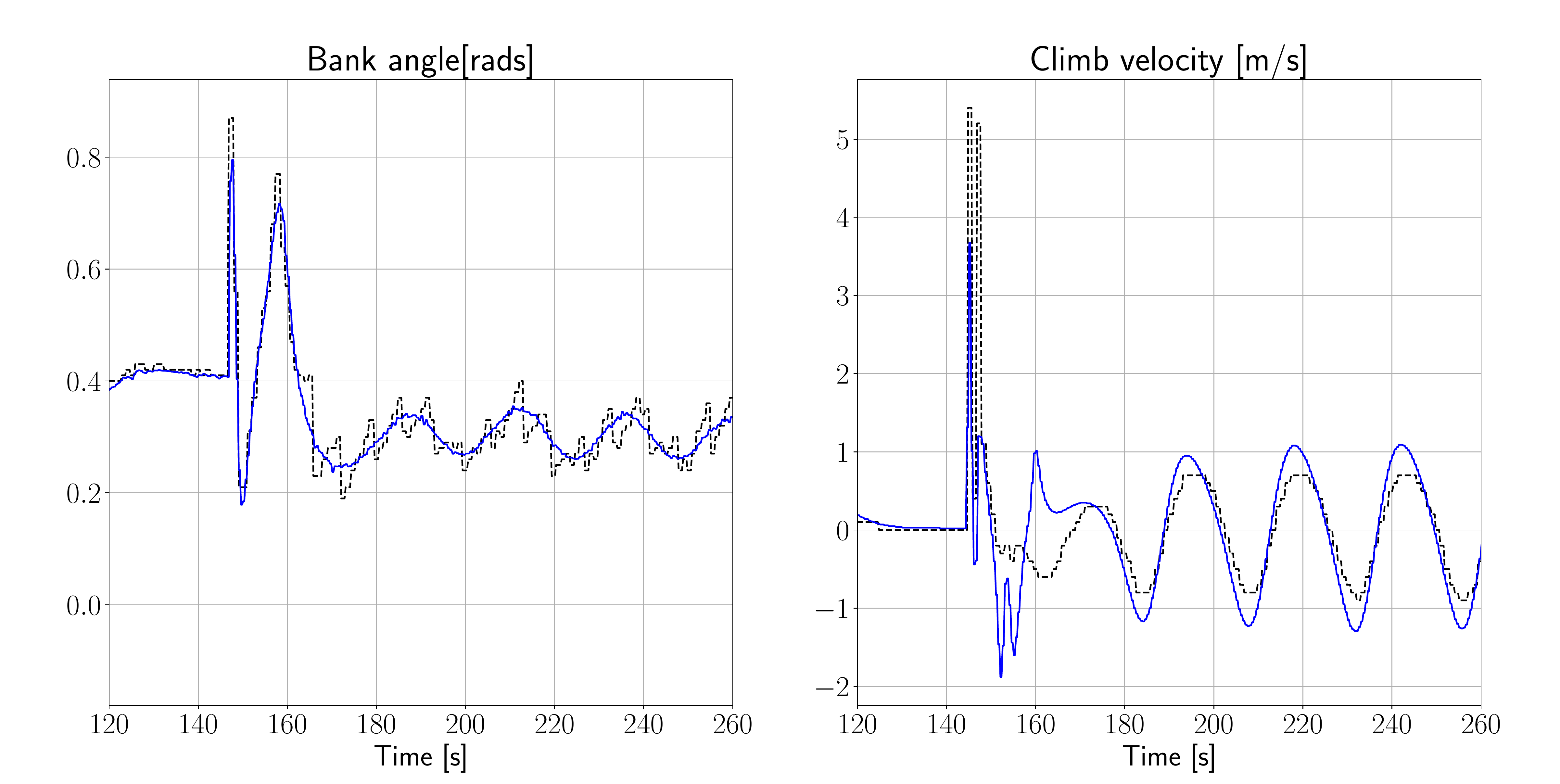}
	\includegraphics[width=\columnwidth]{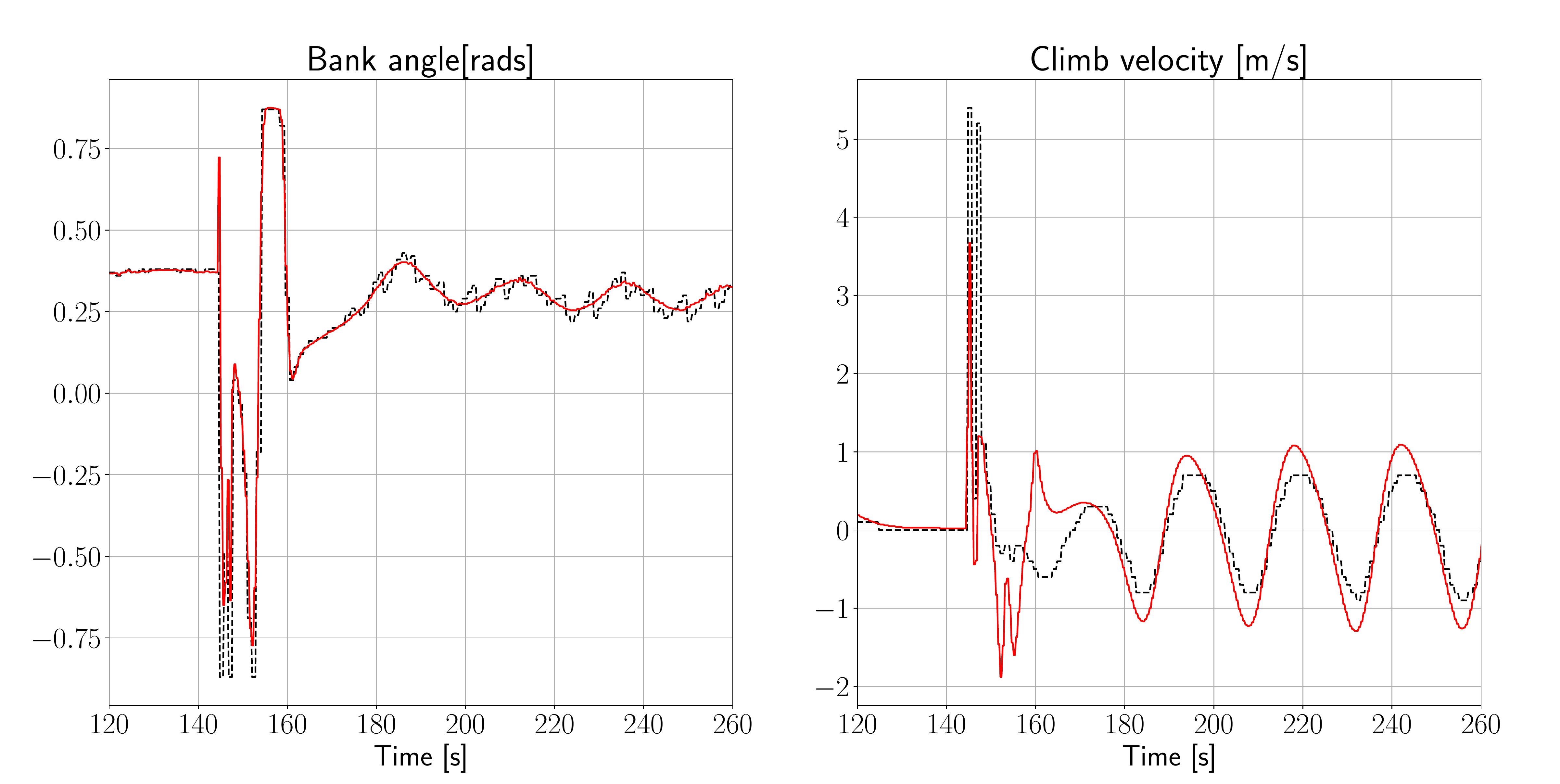}
	\caption{The black curves are the desired values for the bank angles and the vertical velocities produced by the guiding vector field algorithm for the flight in Figure \ref{fig: planes2phi}. The blue and red curves represent the actual states of the first and second aircraft, respectively. For more information on the employed low-level controllers, see \url{https://wiki.paparazziuav.org/wiki/Control_Loops}.}
	\label{fig: planesu}
	\vspace{-1em}
\end{figure}

The two experiments show that once an aircraft flies far ahead of its partner, the algorithm guides the airplane to deviate from the curve to travel more distance, hence, to ``wait'' for its partner. Nevertheless, these deviations from the desired path are within the order of one or two meters (see Fig. \ref{fig: planesphi} and \ref{fig: planes2phi}). The generated signals by the algorithm and the actual states of the aircraft can be seen in Fig. \ref{fig: planesu}.

Note that the employed aircraft do \emph{not} control their ground speeds. In fact, they have a reference signal in their throttle to keep a safe airspeed, and the aircraft increase/decrease such a reference to ascend/descend. %
A traditional trajectory tracking algorithm would \emph{force} the aircraft to track an \emph{open-loop} (i.e., state independent) point; hence, it requires controlling the airspeed/ground speed of the aircraft. Such a requirement is demanding if the aircraft are not equipped with the required sensors or actuators (e.g., spoilers/flaps), and the wind always affects the speed of the airplane. In contrast, our guidance algorithm is free from such a requirement since the parameter $w$ is in \emph{closed-loop} with the aircraft state, and thereby adapting automatically to the positions and velocities of the aircraft.

\section{Conclusion and future work} \label{sec:conclusion}
In this paper, we address the problem of multi-robot coordinated navigation using a novel guiding vector field. The proposed approach enables an arbitrary number of robots to follow or navigate possibly different desired paths or surfaces and achieve motion coordination in a distributed way. Specifically, we derive a time-invariant coordinating guiding vector field to rigorously guarantee the convergence and motion coordination on desired paths or surfaces. This is achieved by exploiting the higher-dimensional guiding vector field with path or surface parameters as virtual coordinates, and employing distributed consensus algorithms to render the virtual coordinate differences (i.e., parametric differences) converging to the predefined ones. Based on the coordinating guiding vector field, a control law is designed for a non-holonomic Dubins-car-like robot model, which considers the actuation saturation. Moreover, we also elaborate on how to effectively integrate the coordinating guiding vector field with a safety barrier certificate to realize collision avoidance among robots. Extensive simulation examples verify the effectiveness of our approach and showcase possible practical applications. Furthermore, we conduct outdoor experiments with multiple fixed-wing aircraft to demonstrate our algorithm's practical value including its robustness against wind perturbation, actuation saturation, etc. There are many other fascinating features of our approach as outlined at the end of Section \ref{subsec:contribution}, such as the scalability and the low computational cost. Note that the (communication) graph topology is assumed to be fixed in this work, but our algorithm is still effective under time-varying or switching graph topologies if additional assumptions/conditions are satisfied (e.g., \cite[Theorem 9]{olfati-saber2004ConsensusProblemsNetworks}, \cite[Theorem 2.31, Corollary 2.32]{ren2008DistributedConsensusMultivehicle}).

Our work opens multiple directions for future studies. i) Since the motion coordination is quantified in terms of path or surface parameters, it is reasonable that the robots might not keep equal or manually specified Euclidean distances if the parametric equations are highly nonlinear (e.g., trigonometric equations). Therefore, one may investigate a transformation method between the parameter space and the Euclidean space such that the motion coordination can be quantified in terms of Euclidean distances. ii) One may study time-varying coordinating guiding vector fields to track moving desired paths or time-varying desired parametric differences $\Delta^{[i,j]}(t)$. iii) The deadlock problem may arise in generating a new vector field incorporating collision avoidance in Section \ref{sec:collision}. This problem has been partially addressed by \cite[Section VII]{wang2017safety}, while the full solution is worth further investigation.

\appendices
\section{Proof of Theorem \ref{thm_vf_invariant}} \label{app_thm_vf_invariant}
\begin{proof}
	For simplicity, we first consider Robot $i$, and most of the function arguments are ignored henceforth unless ambiguity arises. We define 
	\begin{align*}
		K^{[i]} &\defeq \diag{\mk{i}{1}, \dots, \mk{i}{n}} \\
		\bm{{f^{[i]}}'}(w^{[i]}) &\defeq \transpose{(\mfdot{i}{1}(w^{[i]}), \dots, \mfdot{i}{n}(w^{[i]}) )}.
	\end{align*}
	Then one can calculate that
	\begin{align*}
		\transpose{\bm{\nabla \mphi{i}{j}} } \bm{\vfpf^{[i]}} &= \scalemath{0.9}{ \matr{0 \\ \vdots \\ 1 \\ \vdots \\ 0 \\ -\mfdot{i}{j}(w^{[i]})}^\top  \matr{ (-1)^n \mfdot{i}{1}(w^{[i]}) - \mk{i}{j} \mphi{i}{1} \\  
				\vdots \\ 
				(-1)^n \mfdot{i}{n}(w^{[i]}) - \mk{i}{n} \mphi{i}{n} \\ 
				(-1)^n + \sum_{l=1}^{n} \mk{i}{l} \mphi{i}{l} \mfdot{i}{l}(w^{[i]}) } } \\
		&= - \mk{i}{j} \mphi{i}{j} - \mfdot{i}{j}(w^{[i]}) \cdot [\transpose{\bm{{f^{[i]}}'}(w^{[i]})} K^{[i]} \bm{\Phi^{[i]}}] 
	\end{align*} 
	for $j \in \mathbb{Z}_1^n$, where $1$ is at the $j$-th entry of $\bm{\nabla \mphi{i}{j}}$. Therefore, 
	\begin{equation} \label{eq2_multi}
		\matr{\transpose{\bm{\nabla \mphi{i}{1}} }  \bm{\vfpf^{[i]}}  \\ \vdots \\ \transpose{\bm{\nabla \mphi{i}{n}} }  \bm{\vfpf^{[i]}} } = - K^{[i]} \bm{\Phi^{[i]}}  - \bm{{f^{[i]}}'}(w^{[i]}) \transpose{\bm{{f^{[i]}}'}(w^{[i]})} K^{[i]} \bm{\Phi^{[i]}}.   
	\end{equation}
	One can also calculate that
	\begin{multline} \label{eq3_multi}
		\matr{\transpose{\bm{\nabla \mphi{i}{1}} }   \bm{\vfcr^{[i]}}(\bm{w}) \\ \vdots \\ \transpose{\bm{\nabla \mphi{i}{n}} }  \bm{\vfcr^{[i]}}(\bm{w})} = -c^{[i]}(\bm{w}) \matr{\mfdot{i}{1}(w^{[i]})  \\ \vdots \\ \mfdot{i}{n}(w^{[i]}) } \\ =  -c^{[i]}(\bm{w}) \bm{{f^{[i]}}'}(w^{[i]}).   
	\end{multline}	
	Hence,
	\begin{multline} \label{eq_phi1dot_multi}
		\bm{\dot{\Phi}^{[i]}} =  \dt \matr{\mphi{i}{1}  \\ \vdots \\ \mphi{i}{n} } = \matr{\transpose{\bm{\nabla \mphi{i}{1}} }   \bm{\vfpf^{[i]}}  + k_c \transpose{\bm{\nabla \mphi{i}{1}} }  \bm{\vfcr^{[i]}}(\bm{w}) \\ \vdots \\ \transpose{\bm{\nabla \mphi{i}{n}} }  \bm{\vfpf^{[i]}}  + k_c \transpose{\bm{\nabla \mphi{i}{n}} }  \bm{\vfcr^{[i]}}(\bm{w})}   \\
		\stackrel{\eqref{eq2_multi},\eqref{eq3_multi}}{=} - K^{[i]} \bm{\Phi^{[i]}}  - \bm{{f^{[i]}}'}(w^{[i]}) \transpose{\bm{{f^{[i]}}'} (w^{[i]})} K^{[i]} \bm{\Phi^{[i]}} \\  -  k_c  c^{[i]}(\bm{w}) \bm{{f^{[i]}}'}(w^{[i]}).
	\end{multline}
	Now we consider all robots. Define 
	\begin{align*}
		\mathfrak{F} &\defeq \diag{\bm{{f^{[1]}}'}, \dots, \bm{{f^{[N]}}'}} \in \mbr[nN \times N] \\
		K &\defeq \diag{K^{[1]}, \dots, K^{[N]}} \in \mbr[nN \times nN].
	\end{align*}
	Taking the time derivative of $\bm{\Phi}$ in \eqref{eq_Phi}, then %
	we have
	\begin{align} \label{eq_phidot_multi}
		&\dot{\bm{\Phi}} = \notag \\ 
		&\scalemath{0.75}{\matr{- K^{[1]} \bm{\Phi^{[1]}}  - \bm{{f^{[1]}}'}(w^{[1]}) \transpose{\bm{{f^{[1]}}'} (w^{[1]})} K^{[1]} \bm{\Phi^{[1]}} - k_c  c^{[1]}(\bm{w}) \bm{{f^{[1]}}'}(w^{[1]}) \\ 
				\vdots \\
				- K^{[N]} \bm{\Phi^{[N]}}  - \bm{{f^{[N]}}'}(w^{[N]}) \transpose{\bm{{f^{[N]}}'} (w^{[N]})} K^{[N]} \bm{\Phi^{[N]}} - k_c  c^{[N]}(\bm{w}) \bm{{f^{[N]}}'}(w^{[N]})} }  \notag \\
		&=- K \bm{\Phi} - \mathfrak{F} \transpose{\mathfrak{F}} K \bm{\Phi} - k_c \mathfrak{F} \bm{c(\bm{w})}.
	\end{align}
	One can also calculate that
	\begin{align} \label{eq_dotwi}
		\begin{split}
			\dot{w}^{[i]} &= \matr{0 & \cdots & 0 & 1} (\bm{\vfpf^{[i]}}(\bm{\xi^{[i]}}) + k_c \bm{\vfcr^{[i]}}(\bm{\xi^{[i]}})) \\
			&= (-1)^n + \transpose{\bm{{f^{[i]}}'}(w^{[i]})} K^{[i]} \bm{\Phi^{[i]}}(\bm{\xi^{[i]}}) + k_c c^{[i]}(\bm{w}).
		\end{split}
	\end{align}
	Since $\bm{\tilde{w}} = \bm{w} - \bm{w}^*$, there holds that
	\begin{align} \label{eq_what_dot}
		\dot{\bm{\tilde{w}}} = \dot{\bm{w}} = (-1)^n \bm{1} + \transpose{\mathfrak{F}} K \bm{\Phi} + k_c \bm{c}(\bm{w}),
	\end{align}
	where $\bm{1} \in \mbr[N]$ is a vector consisting of all ones. 
	Therefore, from \eqref{eq_coordination_stack}, \eqref{eq_phidot_multi} and \eqref{eq_what_dot}, and noting that $D^\top\bm{1} = \bm{0}$, we have the following composite error dynamics:
	\begin{align} \label{eq_error_dyn1_multi}
		\dot{\bm{e}} = \matr{\dot{\bm{\Phi}} \\ \transpose{D} \dot{\bm{\tilde{w}}} } = \matr{ - K \bm{\Phi} - \mathfrak{F} \transpose{\mathfrak{F}} K \bm{\Phi} + k_c \mathfrak{F} L \bm{\tilde{w}} \\ 
			\transpose{D} \transpose{\mathfrak{F}} K\bm{\Phi} - k_c \transpose{D} L \bm{\tilde{w}}}.
	\end{align}
	One needs to prove that $\bm{e}(t) \to \bm{0}$ as $t \to \infty$.

	Consider the following Lyapunov function candidate
	\begin{align} \label{eq_lyapunov_multi}
		V(\bm{e}) = \frac{1}{2} \transpose{\bm{e}} \mathfrak{K} \bm{e}  =\frac{1}{2} \left( \bm{\Phi}^\top K \bm{\Phi} + k_c \transpose{\bm{\tilde{w}}} L \bm{\tilde{w}} \right),
	\end{align}
	where $\mathfrak{K}$ is the the composite gain matrix defined by
	$
	\mathfrak{K} = \diag{K, k_c I_{|\set{E}|}} \in \mbr[(nN+|\set{E}|) \times (nN+|\set{E}|)],
	$
	with $I_{|\set{E}|}$ being the $|\set{E}|$-by-$|\set{E}|$ identity matrix. The time derivative of $V(\bm{e})$ satisfies
	\begin{align}
		\dot{V}(\bm{e}) =& \transpose{\dot{\bm{e}}} \mathfrak{K} \bm{e} \notag \\
		\stackrel{\eqref{eq_error_dyn1_multi}}{=}& \matr{ - K \bm{\Phi} - \mathfrak{F} \transpose{\mathfrak{F}} K \bm{\Phi} + k_c \mathfrak{F} L \bm{\tilde{w}} \notag \\ 
			\transpose{D} \transpose{\mathfrak{F}} K\bm{\Phi} - k_c \transpose{D} L \bm{\tilde{w}}}^\top \matr{K &{} \\ {} & k_c I_{|\set{E}|} } \matr{\bm{\Phi} \\ D^\top \bm{\tilde{w}}} \\
		=& - \norm{K \bm{\Phi}}^2  - \norm{\mathfrak{F}^\top K \bm{\Phi}}^2 + k_c (\mathfrak{F} L \bm{\tilde{w}} )^\top K \bm{\Phi}  \notag \\
		&+k_c \transpose{(\mathfrak{F}^\top K \bm{\Phi})} L \bm{\tilde{w}} - k_c^2 \norm{L \bm{\tilde{w}}}^2  \\
		=&    - \norm{K \bm{\Phi}}^2  - \norm{\transpose{\mathfrak{F}} K \bm{\Phi}}^2 + 2 k_c \transpose{(\transpose{\mathfrak{F}} K \bm{\Phi})} (L \bm{\tilde{w}}) \notag \\
		&-k_c^2 \norm{L \bm{\tilde{w}}}^2    \label{eq35_multi} \\
		=& - \norm{K \bm{\Phi}}^2 - \norm{ \mathfrak{F}^\top K \bm{\Phi} - k_c L \bm{\tilde{w}} }^2  \\
		\le&  - \norm{K \bm{\Phi}}^2  \le 0.   \label{eq11_multi}
	\end{align}
	From \eqref{eq11_multi}, $\dot{V}$ is negative semi-definite. It follows from the LaSalle's invariance principle \cite[Theorem 4.4]{khalil2002nonlinear} that the trajectories of \eqref{eq_ode} will converge to the largest invariant set $\set{A}$ in 
	$
	\set{B} \defeq \{\bm{e} : \dot{V}(\bm{e})=0 \} \subseteq \{\bm{e} : \bm{\Phi}=\bm{0} \}.
	$
	One can observe from \eqref{eq_error_dyn1_multi} that to find the largest invariant set $\set{A}$ in $\set{B}$, one needs to let $\dot{\bm{e}}=\bm{0}$, which requires having $D^\top L\bm{\tilde w} = D^\top D D^\top \bm{\tilde w} = 0$. This further implies that $D^\top \bm{\tilde w} = 0$ as we will show now. Since $D^\top D D^\top \bm{\tilde w} = 0$, it follows that $D D^\top \bm{\tilde w}$ is in the null space of $D^\top$, which is orthogonal to the range space of $D$. However, $D D^\top \bm{\tilde w}$ is already in the range space of $D$. Namely, $D D^\top \bm{\tilde w}$ is both contained in and orthogonal to the range space of $D$. Therefore, $D D^\top \bm{\tilde w}=0$. Repeating the same argument for $D D^\top \bm{\tilde w}=0$ as before, it follows that $D D^\top \bm{\tilde w}=0$ implies $D^\top \bm{\tilde w}=0$. Under this condition and $\bm{\Phi}=0$, one has $\dot{\bm{e}}=\bm{0}$ as desired.
	Therefore, the largest invariant set $\set{A}$ in $\set{B}$ is 
	\[
	\set{A}=\{\bm{e} : \bm{\Phi}=\bm{0},\transpose{D} \bm{\tilde{w}} = \bm{0} \}.
	\]
	This implies that $\norm{\bm{\Phi^{[i]}}} \to 0$ for all $i\in \mathbb{Z}_{1}^{N}$ and  $(w^{[i]}(t) - w^{[j]}(t)) - \Delta^{[i,j]} \to 0$ for $(i,j)\in\mathcal{E}$ as $t \to \infty$. Namely, all robots' path following errors vanish asymptotically, and the differences of neighboring virtual coordinates $\tilde{\bm{w}}$ converge to the desired parametric differences $\bm{\Delta^{*}}$, and thus the coordinated motion is achieved. Note that \eqref{eq_lyapunov_multi} is positive definite and radially unbounded in $\bm{e}$ (i.e., $V(\bm{e}) \to \infty$ as $\norm{\bm{e}} \to \infty$), and its time derivative is negative semi-definite. Hence, the vanishing of the composite error $\bm{e}$ is global no matter how large the initial composite error $\norm{\bm{e}(t=0)}$ is. %
\end{proof}
	\begin{figure*}[!t]
	\small
	\begin{align} \label{eq_phidot_multi_surf}
		\dot{\bm{\Phi}} &= \scalemath{0.9}{\matr{
				- K^{[1]} \bm{\Phi^{[1]}}  - \bm{\mfdotwone{1}{\cdot}} \transpose{\bm{\mfdotwone{1}{\cdot}}} K^{[1]} \bm{\Phi^{[1]}} - \bm{\mfdotwtwo{1}{\cdot}} \bm{\mfdotwtwo{1}{\cdot}}^\top K^{[1]} \bm{\Phi^{[1]}} - k_{c1}  \mc{1}{1}(\bm{w}) \bm{\mfdotwone{1}{\cdot}}(\bm{w}) - k_{c2}  \mc{1}{2}(\bm{w}) \bm{\mfdotwtwo{1}{\cdot}}(\bm{w}) \\ 
				\vdots \\
				- K^{[N]} \bm{\Phi^{[N]}}  - \bm{\mfdotwone{N}{\cdot}} \transpose{\bm{\mfdotwone{N}{\cdot}}} K^{[N]} \bm{\Phi^{[N]}} - \bm{\mfdotwtwo{N}{\cdot}} \bm{\mfdotwtwo{N}{\cdot}}^\top K^{[N]} \bm{\Phi^{[N]}} - k_{c1}  \mc{N}{1}(\bm{w}) \bm{\mfdotwone{N}{\cdot}}(\bm{w}) - k_{c2}  \mc{N}{2}(\bm{w}) \bm{\mfdotwtwo{N}{\cdot}}(\bm{w}) } }   \notag \\
		&=- K \bm{\Phi} - \mathfrak{F}_{1} \transpose{\mathfrak{F}_{1}} K \bm{\Phi} - \mathfrak{F}_{2} \transpose{\mathfrak{F}_{2}} K \bm{\Phi} - k_{c1} \mathfrak{F}_{1} \bm{\mc{\cdot}{1}(\bm{w})} - k_{c2} \mathfrak{F}_{2} \bm{\mc{\cdot}{2}(\bm{w})}.
	\end{align}
	\hrulefill
	\vspace*{4pt}
	\normalsize
	\end{figure*}
\section{Proof of Theorem \ref{thm_vf_invariant_surf}} \label{app:thm_vf_invariant_surf}
\begin{proof}
	Let 
	$K^{[i]} \defeq \diag{\mk{i}{1}, \dots, \mk{i}{n}}$, $\bm{\mfdotwone{i}{\cdot}}(\cdot) \defeq (\mfdotwone{i}{1}, \dots, \mfdotwone{i}{n} )^{\top}$ and $\bm{\mfdotwtwo{i}{\cdot}}(\cdot) \defeq (\mfdotwtwo{i}{1}, \dots, \mfdotwtwo{i}{n} )^{\top}$.
	Then one can calculate that
	\begin{multline}
		\transpose{ \bm{\nabla \mphi{i}{j}} } \bm{\vfsf^{[i]}} \stackrel{\eqref{eq_vfpf_surf}}{=} 
			- \mk{i}{j} \mphi{i}{j} -\mfdotwone{i}{j} \cdot [\bm{\mfdotwone{i}{\cdot}}^\top K^{[i]} \bm{\Phi^{[i]}}]  \\ 
			 -\mfdotwtwo{i}{j} \cdot [\bm{\mfdotwtwo{i}{\cdot}}^\top K^{[i]} \bm{\Phi^{[i]}}] 
	\end{multline} 
	for $j \in \mathbb{Z}_{1}^{N}$, where $1$ is at the $j$-th entry of $\bm{\nabla \mphi{i}{j}}$. Therefore,
	\begin{multline} \label{eq2_multi_surf}
		\matr{\transpose{\bm{\nabla \mphi{i}{1}} }  \bm{\vfsf^{[i]}}  \\ \vdots \\ \transpose{\bm{\nabla \mphi{i}{n}} }  \bm{\vfsf^{[i]}} } = - K^{[i]} \bm{\Phi^{[i]}}  - \bm{\mfdotwone{i}{\cdot}} \cdot[ \transpose{\bm{\mfdotwone{i}{\cdot}}} K^{[i]} \bm{\Phi^{[i]}}] \\ - \bm{\mfdotwtwo{i}{\cdot}} \cdot [\bm{\mfdotwtwo{i}{\cdot}}^\top K^{[i]} \bm{\Phi^{[i]}}] .   
	\end{multline}
	One can also calculate that
	\begin{equation} \label{eq3_multi_surf}
		\begin{split}
			\matr{\transpose{\bm{\nabla \mphi{i}{1}} }   \bm{\vfcrone^{[i]}}(\bm{w}) \\ \vdots \\ \transpose{\bm{\nabla \mphi{i}{n}} }  \bm{\vfcrone^{[i]}}(\bm{w})} &=  -\mc{i}{1}(\bm{w}) \matr{\mfdotwone{i}{1}  \\ \vdots \\ \mfdotwone{i}{n} } \\
			&=  -\mc{i}{1}(\bm{w}) \bm{\mfdotwone{i}{\cdot}}(\bm{w}).   
		\end{split}
	\end{equation}	
	The same calculation applies for $\bm{\vfcrtwo^{[i]}}$. Therefore,
	\begin{align} \label{eq_phi1dot_multi_surf}
		&\bm{\dot{\Phi}^{[i]}} =  \dt \left( \mphi{i}{1}, \dots,  \mphi{i}{n} \right)^\top \notag \\ 
		&= \matr{\transpose{\bm{\nabla \mphi{i}{1}} }   \bm{\vfsf^{[i]}}  + k_{c1} \transpose{\bm{\nabla \mphi{i}{1}} }  \bm{\vfcrone^{[i]}} + k_{c2} \transpose{\bm{\nabla \mphi{i}{1}} }  \bm{\vfcrtwo^{[i]}} \\ \vdots \\ \transpose{\bm{\nabla \mphi{i}{n}} }  \bm{\vfsf^{[i]}}  + k_{c1} \transpose{\bm{\nabla \mphi{i}{n}} }  \bm{\vfcrone^{[i]}} + k_{c2} \transpose{\bm{\nabla \mphi{i}{n}} }  \bm{\vfcrtwo^{[i]}} }  \notag \\
		& 
		\begin{aligned} \stackrel{\eqref{eq2_multi_surf},\eqref{eq3_multi_surf}}{=} 
			&- K^{[i]} \bm{\Phi^{[i]}}  - \bm{\mfdotwone{i}{\cdot}} \transpose{\bm{\mfdotwone{i}{\cdot}}} K^{[i]} \bm{\Phi^{[i]}} \\
			& -\bm{\mfdotwtwo{i}{\cdot}} \bm{\mfdotwtwo{i}{\cdot}}^\top K^{[i]} \bm{\Phi^{[i]}} - k_{c1}  \mc{i}{1}(\bm{w}) \bm{\mfdotwone{i}{\cdot}}(\bm{w}) \\
			&  - k_{c2}  \mc{i}{2}(\bm{w}) \bm{\mfdotwtwo{i}{\cdot}}(\bm{w}).
		\end{aligned}
	\end{align}
	for $i \in \mathbb{Z}_{1}^{N}$. Now we consider all robots. Define 
	\begin{align*}
		\mathfrak{F}_{1} &\defeq \diag{\bm{\mfdotwone{1}{\cdot}}, \dots, \bm{\mfdotwone{N}{\cdot}}} \in \mbr[nN \times N] \\
		\mathfrak{F}_{2} &\defeq \diag{\bm{\mfdotwtwo{1}{\cdot}}, \dots, \bm{\mfdotwtwo{N}{\cdot}}} \in \mbr[nN \times N] \\
		K &\defeq \diag{K^{[1]}, \dots, K^{[N]}} \in \mbr[nN \times nN] \\
		\bm{\Phi} &\defeq \transpose{(\transpose{\bm{\Phi^{[1]}}},  \cdots, \transpose{\bm{\Phi^{[N]}}})} \in \mbr[nN].
	\end{align*}
	Then %
	we have \eqref{eq_phidot_multi_surf}. One can also calculate that
	\begin{align} \label{eq_dotwi_surf}
		\begin{split}
			\dot{w}^{[i]}_1 &= \matr{0 & \cdots & 0 & 1 & 0} (\bm{\vfsf^{[i]}} + k_{c1} \bm{\vfcrone^{[i]}} + k_{c2} \bm{\vfcrtwo^{[i]}} ) \\
			&= (-1)^n v_{n+2} + \transpose{\bm{\mfdotwone{i}{\cdot}}} K^{[i]} \bm{\Phi^{[i]}}(\bm{\xi^{[i]}}) + k_{c1} \mc{i}{1}, \\
			\dot{w}^{[i]}_2 &= \matr{0 & \cdots & 0 & 0 & 1} (\bm{\vfsf^{[i]}} + k_{c1} \bm{\vfcrone^{[i]}} + k_{c2} \bm{\vfcrtwo^{[i]}} ) \\
			&= (-1)^{n+1} v_{n+1} + \transpose{\bm{\mfdotwtwo{i}{\cdot}}} K^{[i]} \bm{\Phi^{[i]}}(\bm{\xi^{[i]}}) + k_{c2} \mc{i}{2}.
		\end{split}
	\end{align}
	By $\bm{\tilde{w}} \defeq \bm{w} - \bm{w}^*$, there holds
	\begin{align} \label{eq_what_dot_surf}
		\bm{\mdottildew{\cdot}{1}} &= \bm{\mdotw{\cdot}{1}} = (-1)^n v_{n+2} \bm{1} + \transpose{\mathfrak{F}_{1}} K \bm{\Phi} + k_{c1} \bm{\mc{\cdot}{1}}, \\
		\bm{\mdottildew{\cdot}{2}} &= \bm{\mdotw{\cdot}{2}} = (-1)^{n+1} v_{n+1} \bm{1} + \transpose{\mathfrak{F}_{2}} K \bm{\Phi} + k_{c2} \bm{\mc{\cdot}{2}}.
	\end{align}
	Therefore, from \eqref{eq_coordination_stack_surf}, \eqref{eq_error_surf}, \eqref{eq_phidot_multi_surf} and \eqref{eq_what_dot_surf}, and noting that $D^\top\bm{1} = \bm{0}$, we have the following composite error dynamics:
	\begin{equation} \label{eq_error_dyn1_multi_surf}
		\begin{split}
			\dot{\bm{e}} &= \matr{ \dot{\bm{\Phi}} \\ D^\top \bm{\mdottildew{\cdot}{1}} \\ D^\top \bm{\mdottildew{\cdot}{2}} } \\
			&=  \scalemath{0.8}{
				\matr{ - K \bm{\Phi} - \mathfrak{F}_{1} \transpose{\mathfrak{F}_{1}} K \bm{\Phi} - \mathfrak{F}_{2} \transpose{\mathfrak{F}_{2}} K \bm{\Phi} + k_{c1} \mathfrak{F}_{1} L \bm{\mtildew{\cdot}{1}} + k_{c2} \mathfrak{F}_{2} L \bm{\mtildew{\cdot}{2}} \\ 
					D^\top \transpose{\mathfrak{F}_{1}} K \bm{\Phi} - k_{c1} D^\top L \bm{\mtildew{\cdot}{1}} \\
					D^\top \transpose{\mathfrak{F}_{2}} K \bm{\Phi} - k_{c2} D^\top L \bm{\mtildew{\cdot}{2}}
				}.
			}
		\end{split}
	\end{equation}
	
	Define the composite gain matrix to be 
	$\mathfrak{K} = \diag{K, k_{c1} I_{|\set{E}|}, k_{c2} I_{|\set{E}|}} \in \mbr[(nN+2|\set{E}|) \times (nN+2|\set{E}|)]$, and consider the following Lyapunov function candidate
	\begin{multline} \label{eq_lyapunov_multi_surf}
		V(\bm{e}) = \frac{1}{2} \transpose{\bm{e}} \mathfrak{K} \bm{e} = \frac{1}{2} \left( \bm{\Phi}^\top K \bm{\Phi} + k_{c1} \transpose{\bm{\mtildew{\cdot}{1}}} L \bm{\mtildew{\cdot}{1}} \right. \\ \left. + k_{c2} \transpose{\bm{\mtildew{\cdot}{2}}} L \bm{\mtildew{\cdot}{2}} \right),
	\end{multline}
	of which the time derivative satisfies
	\begin{align}
		&\dot{V}(\bm{e}) = \transpose{\dot{\bm{e}}} \mathfrak{K} \bm{e} \notag \\
		\stackrel{\eqref{eq_error_dyn1_multi_surf}}{=}& \scalemath{0.9}{ \matr{ - K \bm{\Phi} - \mathfrak{F}_{1} \transpose{\mathfrak{F}_{1}} K \bm{\Phi} - \mathfrak{F}_{2} \transpose{\mathfrak{F}_{2}} K \bm{\Phi} + k_{c1} \mathfrak{F}_{1} L \bm{\mtildew{\cdot}{1}} + k_{c2} \mathfrak{F}_{2} L \bm{\mtildew{\cdot}{2}} \\ 
				D^\top \transpose{\mathfrak{F}_{1}} K \bm{\Phi} - k_{c1} D^\top L \bm{\mtildew{\cdot}{1}} \\
				D^\top \transpose{\mathfrak{F}_{2}} K \bm{\Phi} - k_{c2} D^\top L \bm{\mtildew{\cdot}{2}}
			}^\top} \notag \\&
		\matr{K &{} &{} \\ {} & k_{c1} I_{|\set{E}|} &{} \\ {} & {} & k_{c2} I_{|\set{E}|} } \matr{\bm{\Phi} \\ D^\top \bm{\mtildew{\cdot}{1}} \\ D^\top \bm{\mtildew{\cdot}{2}}  }   \\
		=& - \norm{K \bm{\Phi}}^2  - \norm{\mathfrak{F}_{1}^\top K \bm{\Phi}}^2  - \norm{\mathfrak{F}_{2}^\top K \bm{\Phi}}^2 \notag \\ & + k_{c1} (\mathfrak{F}_{1} L \bm{\mtildew{\cdot}{1}} )^\top K \bm{\Phi} + k_{c2} (\mathfrak{F}_{2} L \bm{\mtildew{\cdot}{2}} )^\top K \bm{\Phi}  \notag \\ & +k_{c1} \transpose{(\mathfrak{F}_{1}^\top K \bm{\Phi})} L \bm{\mtildew{\cdot}{1}} - k_{c1}^2 \norm{L \bm{\mtildew{\cdot}{1}}}^2  \notag \\ & +k_{c2} \transpose{(\mathfrak{F}_{2}^\top K \bm{\Phi})} L \bm{\mtildew{\cdot}{2}} - k_{c2}^2 \norm{L \bm{\mtildew{\cdot}{2}}}^2 \\
		=&    - \norm{K \bm{\Phi}}^2  - \norm{\transpose{\mathfrak{F}_{1}} K \bm{\Phi}}^2 - \norm{\mathfrak{F}_{2}^\top K \bm{\Phi}}^2 \notag \\ & + 2 k_{c1} \transpose{(\transpose{\mathfrak{F}_{1}} K \bm{\Phi})} (L \bm{\mtildew{\cdot}{1}}) + 2 k_{c2} \transpose{(\transpose{\mathfrak{F}_{2}} K \bm{\Phi})} (L \bm{\mtildew{\cdot}{2}})  \notag \\ & - k_{c1}^2 \norm{L \bm{\mtildew{\cdot}{1}}}^2  - k_{c2}^2 \norm{L \bm{\mtildew{\cdot}{2}}}^2   \label{eq35_multi_surf} \\
		=&  - \norm{K \bm{\Phi}}^2-  \norm{\transpose{\mathfrak{F}_{1}} K \bm{\Phi} - k_{c1} L \bm{\mtildew{\cdot}{1}} }^2  \notag \\ &   - \norm{\mathfrak{F}_{2}^\top K \bm{\Phi} - k_{c2} L \bm{\mtildew{\cdot}{2}} }^2     \notag \\
		\le&  - \norm{K \bm{\Phi}}^2  \le 0.   \label{eq11_multi_surf}
	\end{align}
	From \eqref{eq11_multi_surf}, $\dot{V}$ is negative semi-definite. 
	By \cite[Theorem 8.4]{khalil2002nonlinear}, $\norm{K \bm{\Phi}}^2 \to 0$ as $t \to \infty $, hence $\norm{\bm{\Phi^{[i]}}} \to 0$ as $t \to \infty$ for all $i\in \mathbb{Z}_{1}^{N}$; i.e., all robots' path following errors vanish asymptotically. Moreover, the quadratic form of the Lyapunov function $V$ implies that $V$ is radially unbounded with respect to $\norm{\bm{e}}$ (i.e., $V(\bm{e}) \to \infty$ as $\norm{\bm{e}} \to \infty$), and thus the convergence holds globally: the norm of the initial path-following error $\norm{\bm{e}(0)}$ can be arbitrarily large.
	
	To prove the convergence of the second and third term $\transpose{D} \bm{\mtildew{\cdot}{1}}$, $\transpose{D} \bm{\mtildew{\cdot}{2}}$ of the composite error vector, we use Barbalat's lemma \cite[Lemma 8.2]{khalil2002nonlinear}. Firstly, \eqref{eq11_multi_surf} shows that $\dot{V} \le 0$, hence $V(t) \le V(0)$ for $ t \ge 0$. This implies that the composite error $\bm{e}$ is bounded, and thus $\bm{\Phi}$ and $D^\top \bm{\mtildew{\cdot}{1}}$, $D^\top \bm{\mtildew{\cdot}{2}}$ are all bounded. Due to Assumption \ref{assump_bounded_f_deri_surf}, one can verify that $\dot{\bm{e}}$ in \eqref{eq_error_dyn1_multi_surf} is also bounded, and thus $\dot{\bm{\Phi}}$, $D^\top \bm{\mdottildew{\cdot}{1}}$, $D^\top \bm{\mdottildew{\cdot}{2}}$ are bounded as well. Next, we show that the second-order time derivative $\ddot{V}$ is bounded. One can calculate that $\ddot{V} =  \transpose{\ddot{\bm{e}}} \mathfrak{K} \bm{e} +  \transpose{\dot{\bm{e}}} \mathfrak{K} \dot{\bm{e}}$. It is obvious that the second term of $\ddot{V}$ is bounded, so we only need to show that $\ddot{\bm{e}}$ is bounded. We have 
	\begin{align*}
		\ddot{\bm{e}} = \matr{ \ddot{e}_1  \\ 
			\transpose{D} (\transpose{\dot{\mathfrak{F}}_1} K \bm{\Phi} + \transpose{\mathfrak{F}_1} K \dot{\bm{\Phi}}) - k_{c1} \transpose{D} L \bm{\mdottildew{\cdot}{1}} \\
			\transpose{D} (\transpose{\dot{\mathfrak{F}}_2} K \bm{\Phi} + \transpose{\mathfrak{F}_2} K \dot{\bm{\Phi}}) - k_{c2} \transpose{D} L \bm{\mdottildew{\cdot}{2}}
		}, 
	\end{align*}
	where $\ddot{e}_1=- K \dot{\bm{\Phi}} - \dot{\mathfrak{F}}_1 \transpose{\mathfrak{F}_1} K \bm{\Phi} - \mathfrak{F}_1 (\transpose{\dot{\mathfrak{F}}_1} K \bm{\Phi} + \transpose{\mathfrak{F}_1} K \dot{\bm{\Phi}})  + k_{c1} \dot{\mathfrak{F}}_1 L \bm{\mtildew{\cdot}{1}} + k_{c1} \mathfrak{F}_1 L \bm{\mdottildew{\cdot}{1}} 
	- \dot{\mathfrak{F}}_2 \transpose{\mathfrak{F}_2} K \bm{\Phi} - \mathfrak{F}_2 (\transpose{\dot{\mathfrak{F}}_2} K \bm{\Phi} + \transpose{\mathfrak{F}_2} K \dot{\bm{\Phi}})  + k_{c2} \dot{\mathfrak{F}}_2 L \bm{\mtildew{\cdot}{2}} + k_{c2} \mathfrak{F}_2 L \bm{\mdottildew{\cdot}{2}}$, and $\dot{\mathfrak{F}}_1 = \diag{\dt \bm{\mfdotwone{1}{\cdot}}, \dots, \dt \bm{\mfdotwone{N}{\cdot}} } $ with 
	\[
	\dt \bm{\mfdotwone{i}{\cdot}} = \matr{ \mfddotw{1}{1}{i}{1} & \mfddotw{2}{1}{i}{1} \\ \vdots & \vdots \\ \mfddotw{1}{1}{i}{n} & \mfddotw{2}{1}{i}{n} } \matr{\dot{w}^{[i]}_1 \\ \dot{w}^{[i]}_2}
	\]
	for $i\in \mathbb{Z}_{1}^{N}$.
	Since $\bm{\Phi}$, $D^\top \bm{\mtildew{\cdot}{1}}$, $D^\top \bm{\mtildew{\cdot}{2}}$, $\dot{\bm{\Phi}}$, $D^\top \bm{\mdottildew{\cdot}{1}}$, $D^\top \bm{\mdottildew{\cdot}{2}}$ are all bounded, and by \eqref{eq_dotwi_surf} and Assumption \ref{assump_bounded_f_deri_surf}, we have that $\dot{\mathfrak{F}}_1$, $\dot{\mathfrak{F}}_2$ are bounded, hence $\ddot{\bm{e}}$ is indeed bounded. Therefore, $\ddot{V}$ is bounded, and thus $\dot{V}$ is uniformly continuous in time $t$. By invoking Lemma 4.3 in \cite{slotine1991applied}, we have that $\dot{V}(\bm{e}) \to 0$ as $t \to \infty$. Equivalently, $\transpose{\dot{\bm{e}}} \mathfrak{K} \bm{e} \to 0$ as $t \to \infty$. Since the argument above shows that $\norm{\bm{\Phi}} \to 0$ as $t \to \infty$, from $\eqref{eq35_multi_surf}$ and the boundedness of $\mathfrak{F}_1$, $\mathfrak{F}_2$, $D^\top \bm{\mtildew{\cdot}{1}}$, $D^\top \bm{\mtildew{\cdot}{2}}$, we have $L \bm{\mtildew{\cdot}{1}}\to 0$ and $L \bm{\mtildew{\cdot}{2}}\to 0$ as $t \to \infty$.	
	Furthermore, by Assumption \ref{assump_graph}, $L \bm{\mtildew{\cdot}{1}}\to 0$ and $L \bm{\mtildew{\cdot}{2}}\to 0$ as $t \to \infty$ imply that $\tilde{w}^{[i]}_1 - \mtildew{j}{1} \to 0$ and $\tilde{w}^{[i]}_2 - \mtildew{j}{2} \to 0$ for all $i,j\in \mathbb{Z}_{1}^{N}$ \cite[Corollary 2.5]{ren2008distributed}. This further implies that $\mw{i}{1}(t) - \mw{j}{1}(t) \to \Delta^{[i,j]}_{1}(t)$ and $w^{[i]}_2(t) - \mw{j}{2}(t) \to \Delta^{[i,j]}_{2}(t)$. Therefore, the differences in the virtual coordinates $\mw{i}{1}$ and $\mw{i}{2}$ of all robots converge to the desired values, and thus the coordinated motion can be achieved. As in the proof of Theorem \ref{thm_vf_invariant}, the global convergence property is attained due to the radial unboundedness of the Lyapunov function $V$ with respect to $\norm{\bm{e}}$.
	
	Using the fact that $\lim_{t \to \infty} K \Phi = \bm{0}$, $\lim_{t \to \infty} L \bm{\mtildew{\cdot}{1}} = \bm{0}$ and $\lim_{t \to \infty} L \bm{\mtildew{\cdot}{1}}= \bm{0}$, we  have
	\begin{align*}
		&\lim\limits_{t \to \infty} \dot{w}^{[i]}_1 \\ 
		=& \lim\limits_{t \to \infty}  \matr{0 & \cdots & 1 & 0} \left( \bm{\vfsf^{[i]}}(\bm{\xi^{[i]}}) + k_{c1} \bm{\vfcrone^{[i]}}(t, \bm{\mw{\cdot}{1}}) \right. \notag \\ &  \left. + k_{c2} \bm{\vfcrtwo^{[i]}}(t, \bm{\mw{\cdot}{2}}) \right) \\
		=&  \lim\limits_{t \to \infty} \matr{0 & \cdots & 1 & 0} \left( \bm{\vfsf^{[i]}}(\bm{\xi^{[i]}}) \right) \\
		\stackrel{\eqref{eq_vfpf_surf}}{=}&  \lim\limits_{t \to \infty} \scalemath{0.9}{ \left( \matr{0 & \cdots & 1 & 0} \matr{  (-1)^n ( v_{n+2} \mfdotwone{i}{1} - v_{n+1} \mfdotwtwo{i}{1} ) \\
				\vdots \\
				(-1)^n ( v_{n+2} \mfdotwone{i}{n} - v_{n+1} \mfdotwtwo{i}{n} ) \\ 
				(-1)^n v_{n+2} \\
				(-1)^{n+1} v_{n+1} } \right) } \\
		=& (-1)^n (-1)^n \dot{w}_{1}^{*} = \dot{w}_{1}^{*}.
	\end{align*}
	Similarly, we have $\lim_{t \to \infty} \dot{w}^{[i]}_2 = \dot{w}_{2}^{*}$. Therefore, the desired motion on the surface described in Problem \ref{problem1_surf} is achieved.
\end{proof}

\section{Proof of Theorem \ref{thm_guidance}} \label{app:thm_guidance}
\begin{proof}
	Since $\mvfcb{i}{1}(\bm{\xi^{[i]}})^2 + \mvfcb{i}{2}(\bm{\xi^{[i]}})^2 > \gamma>0$ by assumption, the control inputs in \eqref{eq_u34} are continuously differentiable, and thus Lipschitz continuous. One can show that the yaw angular control input in \eqref{eq_u_theta} is also Lipschitz continuous. Therefore, the control inputs guarantee that the solution to the dynamical system in \eqref{model} exists and is unique \cite[Theorem 3.1]{khalil2002nonlinear}.  Define the orientation error by $\bm{e^{[i]}_{\mathrm{o}}} \defeq \normv{\bm{h^{[i]}}} - \normv{\bm{\vfcb^{[i]}_p}}$ and the Lyapunov-like function $V = \transpose{ {\bm{e^{[i]}_{\mathrm{o}}}} }  \bm{e^{[i]}_{\mathrm{o}}} / 2$. The time derivative of $V$ is 
	\begin{align} 
		\dot{V} &=  {\bm{\dot{e}^{[i]}_{\mathrm{o}}}} \transpose{{}} \bm{e^{[i]}_{\mathrm{o}}} = (\dot{\theta^{[i]}} - \dot{\theta}^{[i]}_d) \transpose{\normv{\bm{h^{[i]}}}} E \normv{\bm{\vfcb^{[i]}_p}} \label{eq_v1} \\
		&\overset{\eqref{eq_u_theta}}{=}  \big( \sat( \dot{\theta}^{[i]}_d - k_\theta \transpose{\normv{\bm{h^{[i]}}}} E \normv{\bm{\vfcb^{[i]}_p}} ) - \dot{\theta}^{[i]}_d \big) \transpose{\normv{\bm{h^{[i]}}}} E \normv{\bm{\vfcb^{[i]}_p}}, \label{eqvdot}
	\end{align}
	where $\dot{\theta}^{[i]}_d$ is shown in \eqref{eq_dotthetad}, and \eqref{eq_v1} utilizes the identities $\dot{\normv{\bm{h^{[i]}}}} = \dot{\theta^{[i]}} E \normv{\bm{h^{[i]}}}$ and $\dot{\normv{\bm{\vfcb^{[i]}_p}}} = \dot{\theta}^{[i]}_d E \normv{\bm{\vfcb^{[i]}_p}}$ \cite{yao2021singularity}. If the angular control input is not saturated, then \eqref{eqvdot} is simplified to $\dot{V}=-k_\theta (\transpose{\normv{\bm{h^{[i]}}}} E \normv{\bm{\vfcb^{[i]}_p}})^2 \le 0$, and $\dot{V}=0$ if and only if the angle difference between $\normv{\bm{h^{[i]}}}$ and $\normv{\bm{\vfcb^{[i]}_p}}$ is $\sigma^{[i]}=0$ or $\sigma^{[i]}=\pi$. Note that $\transpose{\normv{\bm{h^{[i]}}}} E \normv{\bm{\vfcb^{[i]}_p}} = \sin \sigma^{[i]}$. Therefore, during the upper saturation period when $\dot{\theta}^{[i]}_d - k_\theta \transpose{\normv{\bm{h^{[i]}}}} E \normv{\bm{\vfcb^{[i]}_p}} > b$, we have $\sigma^{[i]}(t) >0 \implies \transpose{\normv{\bm{h^{[i]}}}} E \normv{\bm{\vfcb^{[i]}_p}} >0 \implies \dot{V}= \big( b - \dot{\theta}^{[i]}_d \big) \transpose{\normv{\bm{h^{[i]}}}} E \normv{\bm{\vfcb^{[i]}_p}} < -k_\theta (\transpose{\normv{\bm{h^{[i]}}}} E \normv{\bm{\vfcb^{[i]}_p}})^2 \le 0$. Similarly, during the lower saturation period, we have $ \dot{V}= \big( a - \dot{\theta}^{[i]}_d \big) \transpose{\normv{\bm{h^{[i]}}}} E \normv{\bm{\vfcb^{[i]}_p}} < -k_\theta (\transpose{\normv{\bm{h^{[i]}}}} E \normv{\bm{\vfcb^{[i]}_p}})^2 \le 0$ (since $\transpose{\normv{\bm{h^{[i]}}}} E \normv{\bm{\vfcb^{[i]}_p}}< 0$). Therefore, $V$ is always decreasing in all three cases, and thereby the absolute value of the angle difference $|\sigma^{[i]}|$ is decreasing. Note that $V=0$ if and only if $\normv{\bm{h^{[i]}}}=\normv{\bm{\vfcb^{[i]}_p}}$, or equivalently $\sigma^{[i]}=0$. In addition, $\sigma^{[i]}(t=0) \ne \pi$, hence $\dot{V}(t=0) \ne 0$. Using the Lyapunov argument \cite[Theorem 4.1]{khalil2002nonlinear}, it follows that $V(t)$ converges to $0$ as $t \to \infty$; equivalently, $\sigma^{[i]}(t)$ converges to $0$ as $t \to \infty$.
\end{proof}

\section{Proof of Corollary \ref{coroll1}} \label{app:coroll1}
\begin{proof}
	Given that $k_\theta \in (0, \bar{k}_\theta)$, we have $\max\{\dot{\theta}^{[i]}_d - k_\theta \transpose{\normv{\bm{h^{[i]}}}} E \normv{\bm{\vfcb^{[i]}_p}}\}=d+k_\theta<b$ and $\min\{\dot{\theta}^{[i]}_d - k_\theta \transpose{\normv{\bm{h^{[i]}}}} E \normv{\bm{\vfcb^{[i]}_p}}\}=-d-k_\theta>a$. Therefore, saturation never happens and thus Condition \ref{cond2} in Theorem \ref{thm_guidance} can be neglected.
\end{proof}

\bibliographystyle{IEEEtran}
\bibliography{ref}

\end{document}